\documentclass[journal]{IEEEtran}

\usepackage{amsmath,amsfonts}
\usepackage{array}
\usepackage[caption=false,font=footnotesize,labelfont=sf,textfont=sf]{subfig}
\usepackage{textcomp}
\usepackage{stfloats}
\usepackage{url}
\usepackage{verbatim}
\usepackage{graphicx}
\def\BibTeX{{\rm B\kern-.05em{\sc i\kern-.025em b}\kern-.08em
    T\kern-.1667em\lower.7ex\hbox{E}\kern-.125emX}}
\usepackage{balance}

\newif\ifrebuttal
\rebuttalfalse   %

\usepackage{hyperref}
\usepackage{xcolor}
\usepackage{amsmath,amsfonts,bm,bbm,mathtools,amssymb}
\usepackage{physics}
\usepackage{cleveref}
\crefname{algocf}{algorithm}{algorithms}
\Crefname{algocf}{Algorithm}{Algorithms}
\usepackage{csquotes}
\usepackage{tikz}

\usepackage{booktabs}

\usepackage{wrapfig}

\usepackage{etoolbox}
\AtBeginEnvironment{appendices}{\crefalias{section}{appendix}}

\usepackage{enumitem}

\usepackage[
    style=ieee,
    backend=biber,
    doi=false,
    url=false,
    isbn=false,
    eprint=false,
    maxnames=2,
    minnames=2,
]{biblatex}

\AtBeginBibliography{\footnotesize} %
\ifrebuttal
    \def\colorcomments{true}
\else
    \def\colorcomments{false}
\fi

\newcommand{\defreviewermacro}[2]{%
  \ifthenelse{\equal{\colorcomments}{true}}%
    {\expandafter\newcommand\csname #1\endcsname[1]{\textcolor{#2}{##1}}}%
    {\expandafter\newcommand\csname #1\endcsname[1]{##1}}%
}

\defreviewermacro{reviewereditor}{cyan}
\defreviewermacro{reviewerone}{purple}
\defreviewermacro{reviewerthree}{olive}
\defreviewermacro{reviewersix}{orange}
\defreviewermacro{reviewerseven}{teal}
\defreviewermacro{reviewerten}{purple}

\NewDocumentCommand{\reviewercomment}{o m}{%
  \IfValueTF{#1}{}{\vspace{0.5cm}}%
  \textbf{\textit{#2}}%
  \vspace{0.25cm}%
}

\newcommand{\citet}{\cite}

\newcommand{\subsubsectionheading}[1]{
    \medskip
    \noindent \textbf{#1}
}

\def\secref#1{section~\ref{#1}}

\def\eqref#1{equation~\ref{#1}}

\usepackage[ruled,vlined,linesnumbered]{algorithm2e}

\usepackage{setspace} 

\SetCommentSty{mycommfont}
\SetKwComment{Comment}{$\triangleright$\ }{}
\let\oldnl\nl%

\newcommand{\nonl}{\renewcommand{\nl}{\let\nl\oldnl}}%
\SetAlFnt{\small}           %
\SetAlCapNameFnt{\small}    %
\SetAlCapFnt{\small}        %

\def\1{\bm{1}}

\DeclareMathOperator*{\argmax}{arg\,max}
\DeclareMathOperator*{\argmin}{arg\,min}

\newcommand{\st}{\text{\normalfont{s.t. }}}

\def\vmu{{\bm{\mu}}}

\def\vtheta{{\bm{\theta}}}
\def\vepsilon{{\bm{\epsilon}}}

\def\vtau{{\bm{\tau}}}

\def\va{{\bm{a}}}
\def\vb{{\bm{b}}}
\def\vc{{\bm{c}}}

\def\vg{{\bm{g}}}

\def\vp{{\bm{p}}}
\def\vq{{\bm{q}}}

\def\vu{{\bm{u}}}

\def\vw{{\bm{w}}}
\def\vx{{\bm{x}}}

\def\vz{{\bm{z}}}

\def\mB{{\bm{B}}}

\def\mH{{\bm{H}}}
\def\mI{{\bm{I}}}
\def\mJ{{\bm{J}}}

\def\mQ{{\bm{Q}}}
\def\mR{{\bm{R}}}

\def\mT{{\bm{T}}}

\def\mX{{\bm{X}}}

\def\mSigma{{\bm{\Sigma}}}

\DeclareMathAlphabet{\mathsfit}{\encodingdefault}{\sfdefault}{m}{sl}
\SetMathAlphabet{\mathsfit}{bold}{\encodingdefault}{\sfdefault}{bx}{n}

\def\sR{{\mathbb{R}}}

\def\sZ{{\mathbb{Z}}}

\newcommand{\pdata}{p_{\mathrm{data}}}

\newcommand{\E}[2]{\mathbb{E}_{#1}\left[#2\right]}

\newcommand{\R}{\mathbb{R}}

\newcommand{\tran}{^\top}

\newcommand{\SEthree}{\textsc{SE(3)}}
\newcommand{\sethree}{\mathfrak{se}(3)}
\newcommand{\SOthree}{\textsc{SO(3)}}
\newcommand{\sothree}{\mathfrak{so}(3)}

\newcommand{\qvel}{\dot{\vq}}
\newcommand{\ddq}{\ddot{\vq}}

\newcommand{\trajectory}{\bm{\tau}}

\newcommand{\HBinA}[2]{{}^{#1}\mH^{#2}}

\newcommand{\HEEinWorld}{\HBinA{W}{EE}}

\newcommand{\sdf}{\textsc{sdf}}

\newcommand{\Gaussian}[1]{\mathcal{N}\left(#1\right)}

\newcommand{\alphacumprod}{\bar{\alpha}}
\newcommand{\betaposterior}{\tilde{\beta}}

\newcommand{\relu}{\text{ReLU}}

\newcommand{\objective}{\mathcal{O}}

\graphicspath{
{./content/introduction/figures/}
{./content/experiments/figures/}
{./content/method/figures/}
{./content/rebuttal/figures/}
{./content/biography/figures/}
}

\begin{document}

\bstctlcite{IEEEexample:BSTcontrol}  %

\title{
Motion Planning Diffusion:
Learning and Adapting Robot Motion Planning
with Diffusion Models
}

\author{
Jo\~{a}o Carvalho$^{1}$, 
An T. Le$^{1}$,
Piotr Kicki$^{2,3}$,
Dorothea Koert$^{1,4}$,
and
Jan Peters$^{1,5,6}$%
\\
\vspace{0.5cm}
{\color{orange}
This work has been submitted to the IEEE for possible publication.\\
Copyright may be transferred without notice, after which this version may no longer be accessible.
}
\thanks{
Corresponding author: Jo\~{a}o Carvalho, \href{joao@robot-learning.de}{joao@robot-learning.de}
}%
\thanks{
$^{1}$Intelligent Autonomous Systems Lab, Computer Science Department, Technical University of Darmstadt, Germany;
$^{2}$Poznan University of Technology, Poland;
$^{3}$IDEAS, Warsaw, Poland;
$^{4}$Centre for Cognitive Science, Technical University of Darmstadt, Germany;
$^{5}$German Research Center for AI (DFKI), Research Department: SAIROL, Darmstadt, Germany; 
$^{6}$Hessian.AI, Darmstadt, Germany
}%
}

\markboth{SUBMITTED TO IEEE TRANSACTIONS ON ROBOTICS}%
{Shell \MakeLowercase{\textit{et al.}}: A Sample Article Using IEEEtran.cls for IEEE Journals}

\maketitle

\begin{abstract}
The performance of optimization-based robot motion planning algorithms is highly dependent on the initial solutions, commonly obtained by running a sampling-based planner to obtain a collision-free path.
However, these methods can be slow in high-dimensional and complex scenes and produce non-smooth solutions.
Given previously solved path-planning problems, it is highly desirable to learn their distribution and use it as a prior for new similar problems.
Several works propose utilizing this prior to bootstrap the motion planning problem, either by sampling initial solutions from it, or using its distribution in a maximum-a-posterior formulation for trajectory optimization.
In this work, we introduce Motion Planning Diffusion (MPD), an algorithm that learns trajectory distribution priors with diffusion models.
These generative models have shown increasing success in encoding multimodal data and have desirable properties for gradient-based motion planning, such as cost guidance. 
Given a motion planning problem, we construct a cost function and sample from the posterior distribution using the learned prior combined with the cost function gradients during the denoising process.
Instead of learning the prior on all trajectory waypoints, we propose learning a lower-dimensional representation of a trajectory using linear motion primitives, particularly B-spline curves.
This parametrization guarantees that the generated trajectory is smooth, can be interpolated at higher frequencies, and needs fewer parameters than a dense waypoint representation.
We demonstrate the results of our method ranging from simple $2$D to more complex tasks using a $7$-dof robot arm manipulator.
In addition to learning from simulated data, we also use human demonstrations on a real-world pick-and-place task.
\reviewerten{
The experiment results show that diffusion models are strong priors for encoding multimodal trajectory distributions for optimization-based motion planning.
}
\noindent
\textcolor{blue}{\href{https://sites.google.com/view/motionplanningdiffusion}{https://sites.google.com/view/motionplanningdiffusion}
}

\end{abstract}


\section{Introduction}
\label{sec:introduction-mpd}

Autonomous robots are becoming a ubiquitous technology, and motion planning is an important core component.
Among several methods~\cite{Lav2006planningalgorithms}, optimization-based motion planning is a popular approach for solving robot motion planning problems~\cite{ratliff2009chomp,Mukadam2018-gpmp-ijrr,Schulman2013trajopt}.
Methods in this category formulate planning as an optimization problem, where the goal is to find a trajectory that minimizes a cost function, e.g., collision avoidance, while satisfying constraints, such as joint limits.
Popular methods, such as CHOMP~\cite{ratliff2009chomp},
TrajOpt~\cite{Schulman2013trajopt} and GPMP/GPMP2~\cite{Mukadam2018-gpmp-ijrr}, are heavily dependent on initialization since a bad initial trajectory can lead to getting stuck in local minima and failing to find a collision-free path~\cite{Schulman2013trajopt}.
Without prior information, the initial trajectory is commonly assumed to be a straight line in the configuration space.
However, \citet{Mukadam2018-gpmp-ijrr} mentions that in practice, a straight-line initialization might fail for some tasks, e.g., those with narrow passages.
Thus, for complex motion planning tasks, it is common to first run a sampling-based planner, such as RRT-Connect~\cite{kuffner2000rrtconnect}, followed by an optimization-based planner for trajectory smoothing~\cite{DBLP:conf/amcc/LeuZST21}.

\begin{figure*}[!t]
  \centering
  \subfloat[CHOMP (task space)]{%
    \includegraphics[height=0.145\textheight]{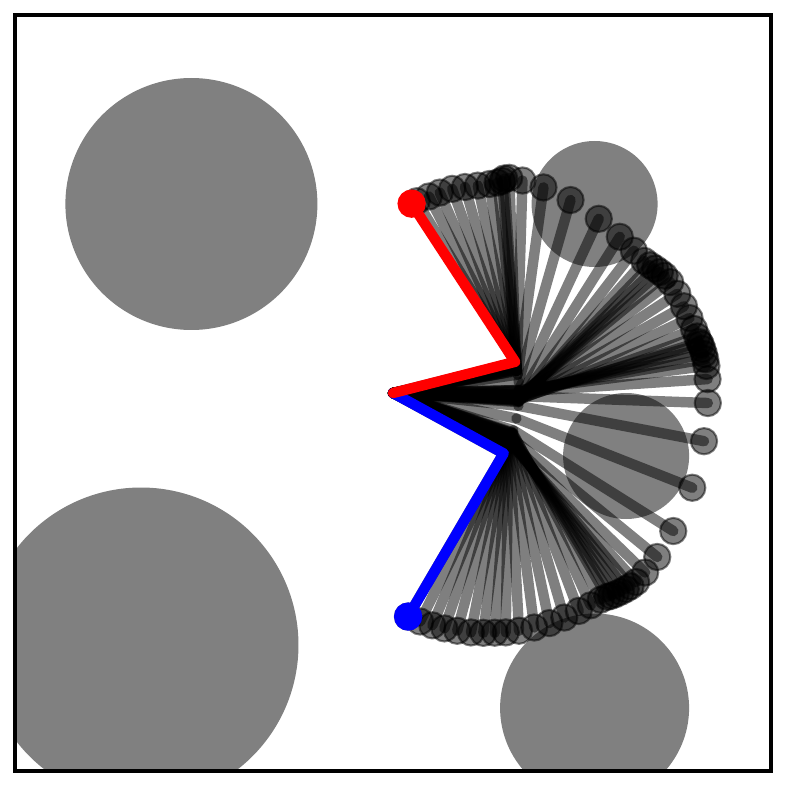}
    \label{fig:planar_2_link_start_goal_task_space_CHOMP}
    }
    \hfill
  \subfloat[CHOMP (joint space)]{%
    \includegraphics[height=0.145\textheight]{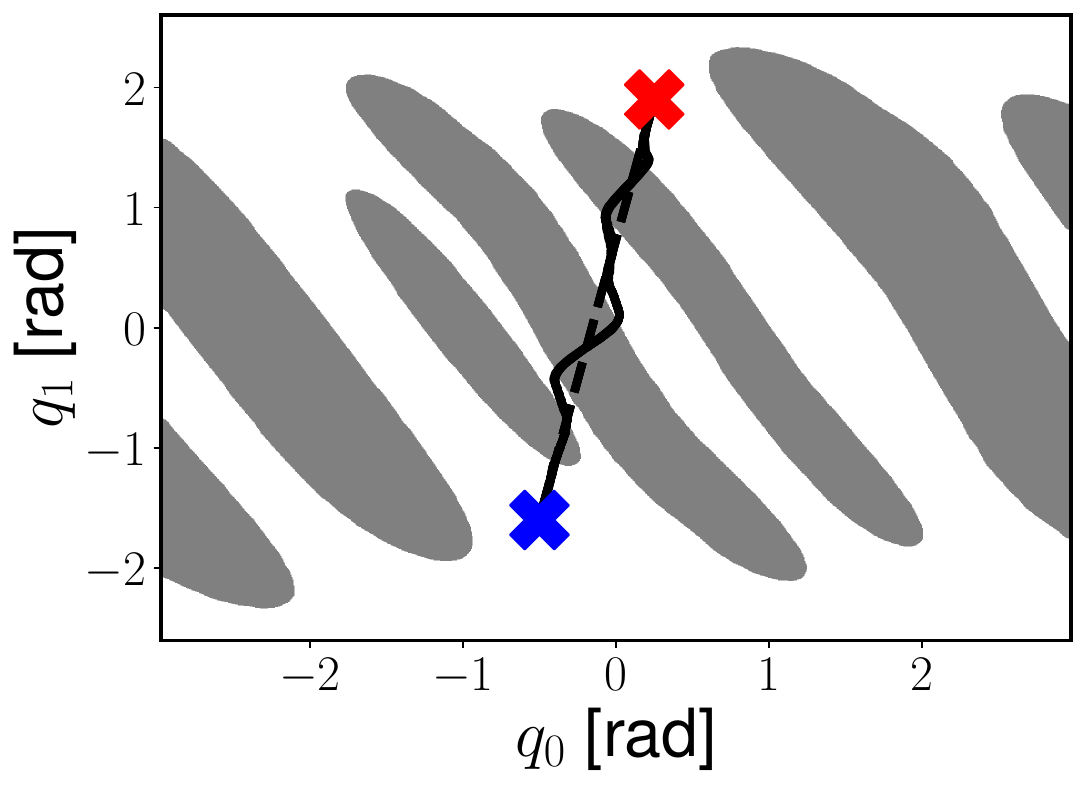}
    \label{fig:planar_2_link_config_space_CHOMP}
    }
    \hfill
  \subfloat[RRT-Connect + CHOMP (task space)]{%
    \includegraphics[height=0.145\textheight]{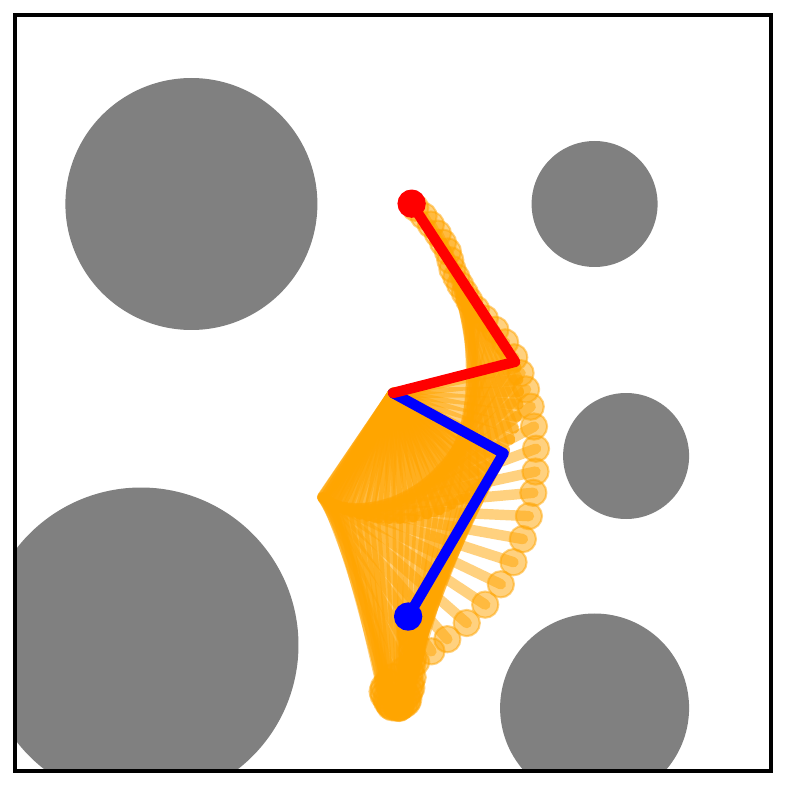}
    \label{fig:planar_2_link_start_goal_task_space_RRTConnect_CHOMP}
    }
    \hfill
  \subfloat[RRT-Connect + CHOMP (joint space)]{%
    \includegraphics[height=0.145\textheight]{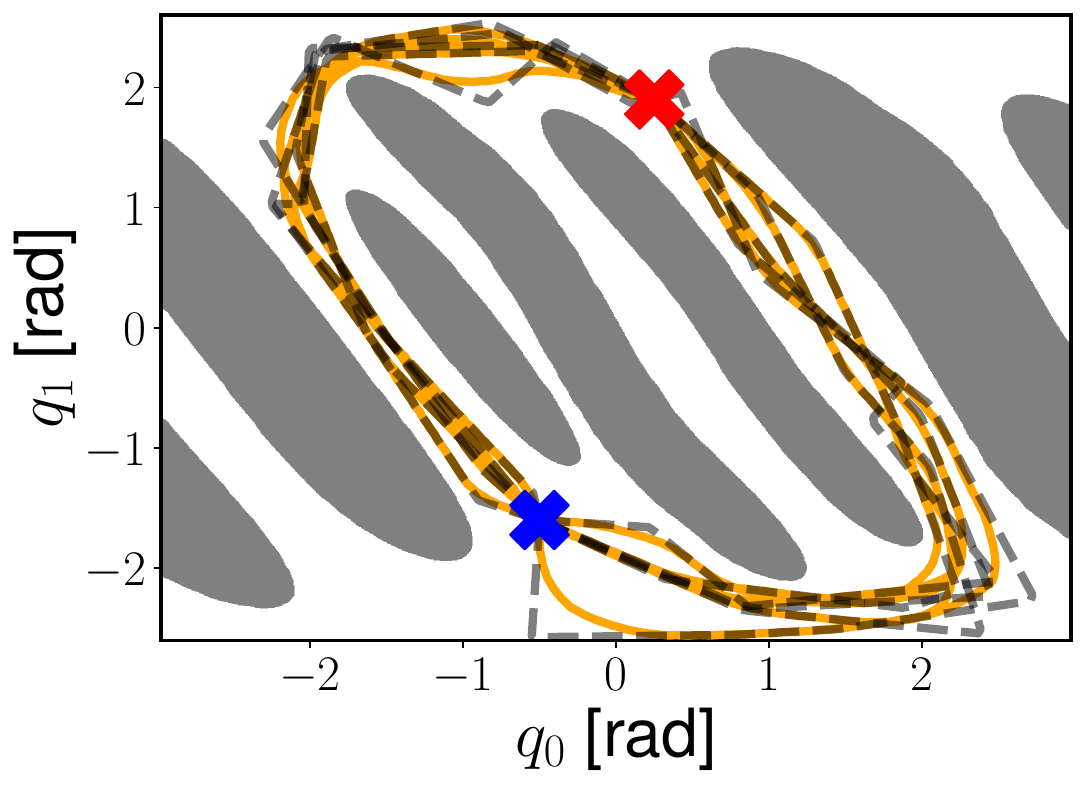}
    \label{fig:planar_2_link_config_space_RRTConnect_CHOMP}
    }
  \caption[Illustration of using CHOMP with and without initialization from a sampled-based planner.]{
    These figures illustrate the need for using \textit{good} initializations for optimization-based motion planning methods when there are narrow passages in the robot's configuration space.
    In \protect\subref{fig:planar_2_link_start_goal_task_space_CHOMP} and \protect\subref{fig:planar_2_link_start_goal_task_space_RRTConnect_CHOMP} the task space of a 2-link robot arm is depicted.
    The grey circles are obstacles that the robot needs to avoid while moving from the \textcolor{blue}{start} configuration to the \textcolor{red}{goal} configuration.
    One robot trajectory corresponding to \protect\subref{fig:planar_2_link_config_space_CHOMP} and \protect\subref{fig:planar_2_link_config_space_RRTConnect_CHOMP} is also shown.
    In \protect\subref{fig:planar_2_link_config_space_CHOMP} and \protect\subref{fig:planar_2_link_config_space_RRTConnect_CHOMP}, the configuration-free space is shown, where the grey areas indicate collision in the task space.
    In \protect\subref{fig:planar_2_link_start_goal_task_space_CHOMP}/\protect\subref{fig:planar_2_link_config_space_CHOMP}, we run CHOMP using a straight-line initialization in the configuration space (black dashed lines).
    The resulting optimized trajectories (solid lines) show that the robot gets stuck in a local minimum and cannot find a collision-free path.
    In \protect\subref{fig:planar_2_link_start_goal_task_space_RRTConnect_CHOMP}/\protect\subref{fig:planar_2_link_config_space_RRTConnect_CHOMP} we first run RRT-Connect (dashed black lines) and initialize CHOMP with these trajectories.
    The resulting optimized trajectories (solid orange lines) are collision-free.
    The initial paths are already collision-free, and during optimization, they are shortened and smoothened using CHOMP.
    It is also possible to observe the two modes found by RRT-Connect.
  }
  \label{fig:motivation_planar_2_link}
  \vspace{-0.5cm}
\end{figure*}

To better illustrate this issue, in \cref{fig:motivation_planar_2_link} we show a motion planning task for a planar 2-link robot.
The task is to obtain a smooth joint trajectory from a start to a goal configuration without colliding with the environment.
\Cref{fig:planar_2_link_config_space_CHOMP} shows how narrow passages appear in the configuration space.
When using a straight-line trajectory in the configuration space, as in \cref{fig:planar_2_link_config_space_CHOMP}, CHOMP cannot escape local minima to find a collision-free path.
As shown in \cref{fig:planar_2_link_config_space_RRTConnect_CHOMP}, if first a global sampling-based planner is used, particularly RRT-Connect, which produces a collision-free path, then the initial trajectory used in CHOMP is already collision-free, albeit being non-smooth.
Using CHOMP, this trajectory is further optimized for smoothness while avoiding collisions.
Additionally, in \cref{fig:planar_2_link_config_space_RRTConnect_CHOMP}, we can see two modes to traverse the configuration space found by multiple runs of RRT-Connect.
These insights also transfer to higher dimensions.
These examples depict characteristics that priors should have: being (almost) collision-free, representing complex trajectories, and encoding multimodality.

Instead of computing an initial path with a sampling-based planner, another approach is to learn from previous motion planning solutions a manifold of trajectories that are good initializations~\cite{Dragan2017,Lembono:295415}.
Several choices must be made when encoding a prior distribution, depending on the nature of the data and the sampling process.
The simplest prior is building a memory of past solutions and using them as initializations~\cite{DBLP:journals/ral/LembonoPPC20}.
However, this approach is not scalable,
and it is not straightforward to generalize to new start and goal configurations other than interpolating between the closest solutions using, e.g., the $k$-nearest neighbors.
A parametrized model removes these issues but must consider whether the data is unimodal or multimodal and how fast we can sample from this model.
If the prior trajectories are unimodal, Gaussian distributions representing the parameters of a Probabilistic Movement Primitive (ProMP)~\cite{koert2016debato} or a Gaussian Process (GP) of the waypoints of a discretized trajectory~\cite{Lembono:295415,rana2017towardsrobustskill} can be used.
These, however, fail to encode multimodal data.
For multimodality, GMMs~\cite{pml1Book} can be used, but they are not easy to train for high-dimensions and large datasets.
Additionally, choosing the number of modes is not trivial~\cite{DBLP:conf/icml/ArenzZN18}.
In contrast to the simple parametric models, deep generative models, such as Variational Auto Encoders (VAEs)~\cite{kingma2013vae}, Generative Adversarial Networks (GANs)~\cite{goodfellow2014gan}, Energy-Based Models (EBMs)~\cite{lecun2006ebm}, and diffusion models~\cite{ho2020denoisingdiffusion,song2021scorebased}, can express multimodal distributions due to the modeling power of deep neural networks.
Methods such as VAEs and GANs are easy to sample from but show problems such as mode-collapse or training instability~\cite{DBLP:conf/icml/ArjovskyCB17}.
EBMs can be difficult to train and sample from~\cite{florence2021implicit}.
Therefore, we chose to model prior trajectory distributions using diffusion models, which present several important qualities when learning from demonstrations.
First, they encode well multimodal data~\cite{DBLP:conf/iclr/Bayat23}, as there might be several collision-free paths~\cite{DBLP:journals/ijrr/Osa20}.
Second, diffusion models are empirically easier to train than others (e.g., GANs and EBMs)~\cite{DBLP:conf/iclr/Bayat23}.
Finally, after learning, given an external likelihood function to optimize, sampling from a posterior distribution is done by smoothly biasing samples from the prior toward the high-likelihood regions during the denoising process~\cite{Dhariwal2021diffusionbeatsgans}.

\reviewerseven{
The key contributions of our work are:
\begin{itemize}
    \item Motion Planning Diffusion (MPD) - a novel method that combines learning and adaptation of robot trajectories using diffusion models; implemented as a diffusion-based generative framework that learns trajectory priors in a compact B-spline parameterization.
    \item A cost-guided posterior sampling, following the planning-as-inference framework, that blends learned trajectory priors with task-specific objectives during the reverse diffusion process, enabling the generation of diverse, feasible, and collision-free trajectories.
    \item An empirical evaluation demonstrating MPD’s effectiveness across a variety of planning problems, ranging from simple 2D setups to environments with a 7-dof robot manipulator, and a real-world pick-and-place task learned from human demonstrations via kinesthetic teaching.
\end{itemize}
}

The rest of this article is structured as follows.
\Cref{sec:related_work} presents related works on learning priors for motion planning and how diffusion models are used in robotics and motion planning.
In \cref{sec:method} we provide the necessary background and present our method.
To highlight the benefits of diffusion models as priors, in \cref{sec:experimental_evaluation} we evaluate our method against several baselines in simulated and real-world tasks.
In \cref{sec:limitations} we discuss the limitations of our work and propose further research ideas.
Lastly, \cref{sec:conclusion} summarizes the key points of this article.

\textit{This work is an extended version} of our previous work~\cite{DBLP:conf/iros/Carvalho0BK023}, in which we added several improvements and more clarification and details.
Instead of learning a model on dense waypoints, we propose using parametric trajectories with lower-dimensional representations that ensure smoothness, namely B-splines.
Previously, we only considered contexts as joint positions at the start and goal.
In this work, we use the desired end-effector goal pose as a conditioning variable, which is more natural for some planning tasks.
We expanded the related work section, described MPD in more detail, and performed more experiments in simulated environments.
To show the method's applicability to real-world tasks, we learn a prior distribution from human demonstrations via kinesthetic teaching for a pick-and-place task, and adapt it by using obstacles not present in the training setup.

\section{Related Work}
\label{sec:related_work}

A large body of literature exists on learning to plan for robotics.
In this section, we discuss classical works in path and motion planning~(\cref{sec:path_and_motion_planning}), and methods that combine learning methods with optimization-based motion planning approaches~(\secref{sec:rw:optimization_based}).
Additionally, we present an overview of how diffusion models are used in robotics and, more specifically, in motion planning~(\secref{sec:rw:diffusion_robotics}).

\subsection{Path and motion planning}
\label{sec:path_and_motion_planning}

Path and motion planning are fundamental components of any robotics system.
Path planning aims to find a collision-free path, and motion planning aims to find a trajectory (a time-dependent path) that prevents the robot from colliding with its environment and respects its joint limits.
Importantly, for robot applications, the solutions should be smooth to avoid jerky movements.
There are two main branches of path and motion planning: sampling-based and optimization-based.

Sampling-based planning algorithms ensure probabilistic completeness by conducting an extensive search over the whole configuration space. 
They sample points, evaluate whether they are in collision, and connect them to form a path.
These include classical algorithms such as PRM~\cite{Kavraki1996PRM}, RRT~\cite{Lavalle98rapidly-exploringrandom}, RRTConnect~\cite{kuffner2000rrtconnect}, or their improved versions, PRM/RRT\uppercase{*}~\cite{karaman2011sampling}, Informed-RRT\uppercase{*}~\cite{gammell2014informedrrt}, BIT\uppercase{*}~\cite{gammell2020bitstar} and AIT\uppercase{*}~\cite{strub2020a}.
One drawback of these planners is that they simply connect configurations in such a way that the resultant paths are typically non-smooth.
Hence, after finding a path, it must be post-processed by a smoother guaranteeing the motion is collision-free~\cite{DBLP:journals/sensors/RavankarRKHP18}.

On the other hand, optimization-based motion planners search for a collision-free movement by locally optimizing a trajectory either via gradient descent or stochastic optimization, aiming for smoothness while satisfying other objective constraints.
These methods can be seen as performing sampling-based planners' search and smoothening steps in one step.
Gradient-based methods include CHOMP~\cite{ratliff2009chomp}, TrajOpt~\cite{Schulman2013trajopt} and GPMP/GPMP2~\cite{Mukadam2018-gpmp-ijrr}.
Stochastic optimization methods based on path integral formulations~\cite{Theodorou2010pathintegral} include STOMP~\cite{Kalakrishnan_RAIIC_2011_stomp}, Stochastic GPMP~\cite{urain_2022_learning_implicit_priors} and MGPTO~\cite{petrovic2022mixturegp}.
Other works solve motion planning by optimal transport~\cite{DBLP:conf/nips/0001CBP23}.

\subsection{Learning priors for optimization-based motion planning}
\label{sec:rw:optimization_based}

Optimization-based planning methods often use an uninformed initial/prior distribution, whose mean is a constant-velocity straight line between the start and goal configurations.
As a deterministic method, CHOMP has no initial distribution.
STOMP uses a distribution with high entropy in the middle of the trajectory and decreasing entropy towards the start and goal points.
GPMP/GPMP2 uses a Gaussian Process prior and likelihoods of the exponential family to perform maximum-a-posteriori (MAP) trajectory optimization.

To find good initializations, we might have access to a database/library containing a set of motion-planning tasks and their respective solutions.
When presented with a new task, some methods take the \mbox{$k$-nearest} neighbor ($k$-NN) solution in the database~\cite{DBLP:conf/icra/StolleA06,DBLP:conf/iros/LiuA09,DBLP:journals/ral/LembonoPPC20}.
$k$-NN approaches may achieve good results in low-dimensional tasks but have two issues:
they need to keep a growing database, which can hurt memory, and suffer from the curse of dimensionality when computing distances in higher dimensions.
Alternatively, other methods, commonly named memory of motion, build a function approximator that maps tasks to solutions~\cite{DBLP:conf/icra/MansardDGTS18, DBLP:journals/ral/LembonoPPC20}. 
Then, they query the learned function at inference time to obtain an initial solution to warm start a trajectory optimizer.
Care is needed when approximating the function mapping from tasks to trajectories.
If a deterministic function approximator is learned using mean squared error~\cite{DBLP:journals/ras/ForteGMU12,DBLP:conf/icra/LamparielloNCHP11}, the learned model averages the solutions, resulting in a poor fit.
A better approach is to learn a multimodal distribution of trajectories~\cite{DBLP:journals/ral/LembonoPPC20,power2022variationalmpcnf}.
With an expressive enough model, we can recover all the distribution modes effectively.
Several of the mentioned methods first obtain a sample from the prior distribution and then optimize it,
which can lead to the final solution being far away from the prior distribution.
This can be problematic if the prior is constructed from human demonstrations, which we'd like to be close to when performing optimization.
In turn, in our work, we use a diffusion model to build a memory of motions, which stay close to the prior while minimizing a cost function.

Other methods propose using learning from human demonstrations~\cite{Takayuki2018imitationlearning} to build trajectory distribution priors.
Then, given a likelihood function, they use sampling or optimization methods to obtain the maximum-a-posterior solution, a sample from the posterior distribution, or even recover the posterior distribution in closed form.
\citet{koert2016debato} introduced DEBATO, which learns a ProMP
of trajectories containing the position of a human hand during demonstrations.
The ProMP weights are Gaussian distributed, so they only encode unimodal distributions.
During inference, obstacles are included in the scene, 
and a return function that includes a negative collision cost and a KL divergence penalty between the current trajectory distribution and the demonstrations is used.
The Gaussian parametrization allows for computing the posterior distribution in closed form by solving the optimization problem using Relative Entropy Policy Search (REPS)~\cite{peters2010reps}.
This work only considers collisions between the end-effector and the environment.
In CLAMP~\cite{rana2017towardsrobustskill}, the authors encoded a trajectory prior distribution of the robot's end-effector position and velocity, which was enough for their tasks, using a Gaussian Process prior.
At inference, they use GPMP to solve the motion planning problem.
The likelihood factors/costs used are to start close to a given start joint state and to avoid collisions of the robot with the environment.
With maximum-a-posterior inference, one solution is found to optimize the likelihood function and penalize deviations from the prior GP distribution.
Like~\cite{koert2016debato}, this method works for unimodal distributions and produces only one solution and not a posterior distribution.
Contrary to these works, we use diffusion models to capture the multimodality of human demonstrations and learn trajectory distributions directly on the robot's joint space.
To handle trajectory multimodality,~\citet{urain_2022_learning_implicit_priors} learns an Energy-Based Model (EBM), whose inputs are the current state and a phase variable, and the output represents the density of the state distribution.
To generate a trajectory from this model, a cost function is built to minimize the energy of the single states across increasing phase values while minimizing the norm between adjacent states.
At inference, new costs are formulated and added as energy functions.
The posterior distribution is obtained in the form of particles by sampling from the EBM using stochastic optimization.
In contrast to incorporating the prior as a cost, we model the full trajectory using diffusion, and optimization is done with gradient-based approaches.
With our formulation, we can directly sample trajectories from the posterior by following the reverse diffusion sampling process, contrasting to sampling from the optimized proposal distribution,
which commonly fails because sampling trajectories is hard for EBMs~\cite{florence2021implicit}.

\subsection{Diffusion models in robotics and motion planning}
\label{sec:rw:diffusion_robotics}

In recent years, several works have explored score-based and diffusion models in robotics. 
They are commonly used as expressive multimodal generative model priors for downstream tasks, as similarly done in image generation~\cite{DBLP:conf/nips/GraikosMJS22}.
At task planning levels,~\citet{Kapelyukh2022-dall-e-bot} proposed \textsc{DALL-E-Bot}, which, given a set of objects, generates a text description from a scene image and then prompts the text-to-image generator DALL-E~\cite{ramesh2022dalle2} to generate a ``proper goal scene''.
\citet{liu2022structdiffusion}~proposes \textit{StructDiffusion} for arranging objects based on language commands by predicting the object arrangements using a language-conditioned diffusion model.
For grasping,~\citet{urain2022se3diffusion} used denoising score matching to learn a generative grasp pose model for a parallel gripper in $\SEthree$ to optimize motion and grasping poses jointly.
Several works have also used diffusion models for generating grasps using dexterous grippers~\cite{weng2024dexdiffuser,DBLP:journals/corr/abs-2407-09899}.
In trajectory planning, \citet{janner2022diffuser}~introduced Diffuser, a trajectory generative model used for planning in Offline RL and long-horizon tasks.
This work introduced a U-Net deep learning architecture to encode trajectories, which is the basis for many follow-up works that model trajectory distributions with diffusion models.
\reviewersix{
MPD builds on the ideas of Diffuser but with a focus on robot motion planning.
Namely, we use a B-spline trajectory representation instead of waypoints, leading to faster inference and smooth trajectories; and at inference time, we add new obstacles not seen during training and generate collision-free trajectories using the environment's signed distance function as a cost function.
}
In behavior cloning and visuomotor imitation learning from human demonstrations, \citet{DBLP:conf/rss/ChiFDXCBS23}~introduced diffusion policy, a method that learns policies that take as input a history of visual observations and outputs a sequence of actions.
The action distribution is modeled with a diffusion model.
Concurrently, \citet{DBLP:conf/rss/ReussLJL23}~introduced BESO, which has a similar structure to diffusion policy, except they use a different formulation of score-based models.
\reviewersix{
These works focused on closed-loop reactive policies that generate short-horizon trajectories, while we focus on generating open-loop point-to-point collision-free trajectories.
}
Recent methods have been built on top of these two works to include better perception, language commands, collision avoidance (for dynamic environments), equivariance and define subgoals for an MPC controller~\cite{Ze2024DP3,ke20253d,feng2025language,DBLP:journals/corr/abs-2406-09767,DBLP:journals/corr/abs-2406-05309,wang2025equivariant,yang2024equibot,DBLP:conf/icra/HuangLYB24}.
In these works, trajectories are typically encoded using waypoints, but in~\cite{DBLP:journals/ral/ScheiklSHFNLM24} the authors introduced Movement Primitive Diffusion, which instead generates the parameters of a Probabilistic Dynamic Movement Primitive (ProDMP)~\cite{DBLP:journals/ral/LiJVOLN23}, guaranteeing smoothness in the predict trajectories.
Similarly, we predict the control points of a B-spline.

In motion planning for robot manipulators, \citet{DBLP:conf/iros/Carvalho0BK023}~introduced the first version of Motion Planning Diffusion.
Given motion planning trajectories obtained from an expert planner, this method learns a diffusion model as a prior distribution over trajectories.
Then, at inference, when new obstacles are added to the scene, MPD samples trajectories from the posterior distribution that is formed with the diffusion prior and a likelihood function that includes collision costs and trajectory smoothening, among others, using classifier guided diffusion~\cite{Dhariwal2021diffusionbeatsgans}.
Since the trajectories are modeled using dense waypoints, an additional cost function (a Gaussian Process trajectory prior) is needed to ensure smooth solutions.
In this work, we encode the trajectory using B-spline coefficients, which have a smaller dimension than the dense waypoints and ensure smoothness by construction.
EDMP~\cite{DBLP:conf/icra/SahaMRSASSK24} follows up on the work of~\cite{DBLP:conf/iros/Carvalho0BK023} by using an ensemble of collision cost functions instead of a single one.
\citet{power2023sampling} used diffusion trajectory models as priors to sample good initializations for solving downstream constrained trajectory optimization problems.
In our work, we show that first sampling from the prior and then only optimizing the cost is less performant than the proposed approach of sampling from the posterior distribution.
SafeDiffuser~\cite{xiao2023safediffuser} uses control barrier functions to extend Diffuser to safety-critical applications.
In APEX~\cite{dastider2024apex}, the idea of integrating sampling and collision avoidance cost guidance using diffusion models is used for bimanual tasks.
Some works have also used environment conditioning to build conditional diffusion models that generate multimodal trajectories given an encoding of the environment in the form of object positions and dimensions, point clouds or RGB images~\cite{carvalho2022conditionedsbm,DBLP:conf/icml/Luo0TD24,huang2023scenediffuser,huang2024diffusionseeder}.
For proper generalization, these approaches require obtaining multiple scenes and several motion plans, which might take too much effort to generate~\cite{fishman2022mpinets}.
Several planning-with-diffusion works can also be found in autonomous driving, quadruped path planning, and UAVs, whose techniques are closely related to robotic manipulators~\cite{teng2023motion,DBLP:conf/icra/LuW0SU24,DBLP:journals/corr/abs-2405-01758,Yang_2024_CVPR,DBLP:conf/icra/LiuSK24,stamatopoulou2024dippest}. 
For a recent review of generative models for robotics consult~\cite{urain2024deepgenerativemodelsrobotics}.

\section{Robot Motion Planning with Diffusion Models}
\label{sec:method}

\begin{figure*}[t]
    \centering
    \includegraphics[width=0.99\linewidth]{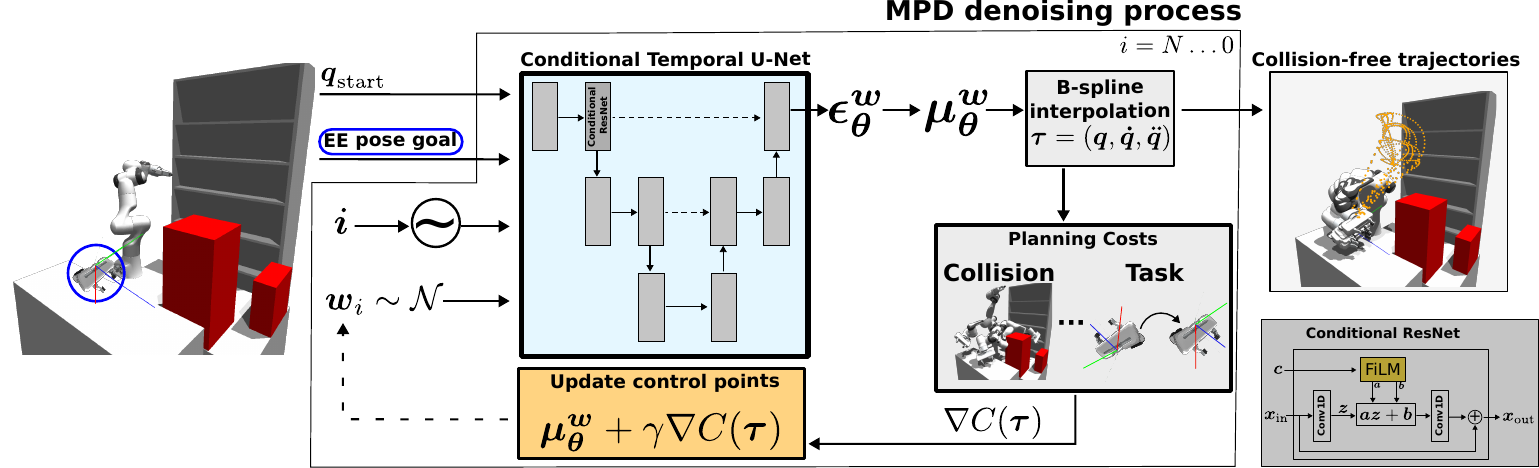}
    \caption[Overview of inference using Motion Planning Diffusion.]{
        Overview of inference using Motion Planning Diffusion.
        An initial joint position $\vq_{\text{start}}$ and an end-effector goal pose $\HEEinWorld_{\text{goal}}$ are context variables to a conditional diffusion model.
        The diffusion model consists of a conditional U-Net architecture (made of conditional ResNet blocks) that takes as input the context, the current time index $i$ and the B-spline control points $\vw_i$, and outputs the denoising vector $\vepsilon_{\vtheta}$.
        The control points mean $\vmu_i$ are obtained with \cref{eq:ddpm_posterior_mean} and used to compute the gradient of a motion planning cost function, which is used to bias the samples towards higher (negative) cost likelihood regions.
        These particles are injected as input to the U-Net at the next denoising step.
        In the conditional ResNet block, the time and context embeddings are stacked and used as input for computing the parameters $\va$ and $\vb$ of a FiLM encoder, which are used to transform the embedded control points ($\vx_{\text{in}}$), computed with a one-dimensional convolution.
        MPD's output are joint space trajectories that balance the prior distribution and the cost likelihood.
    }
    \label{fig:mpd_method_overview}
    \vspace{-0.5cm}
\end{figure*}

This section details our method and the necessary background.
\Cref{fig:mpd_method_overview} shows an overview of the inference process.

\subsection{Optimization-based Motion Planning}
\label{sec:motion_planning}

Starting from an initial joint position $\vq_{\text{start}} \in \R^d$ with zero velocity and acceleration, we consider the task of generating a smooth joint trajectory that either reaches a goal joint position $\vq_{\text{goal}} \in \R^d$ or a goal end-effector pose $\HEEinWorld_{\text{goal}} \in \SEthree$, with zero velocity and acceleration.
The obtained trajectory must avoid collisions, be smooth, short, and satisfy joint limits.
Let $\vq(0)$ and $\vq(T) \in \sR^{d}$ be the start and final joint positions of a robot with $d$ degrees-of-freedom, respectively, $T$ a fixed trajectory duration, $\dot{\vq}$ the velocities, $\ddot{\vq}$ the accelerations and ${\trajectory(t) = (\vq(t), \dot{\vq}(t), \ddot{\vq}(t))}$ a trajectory.
This problem is formulated as trajectory optimization:
\begin{align}
    \label{eq:trajectory_optimization}
    \min_{\trajectory} & \quad C_{\text{vel}}(\trajectory) + C_{\text{acc}}(\trajectory) \\
    \text{s.t.} & \quad \vq(0) = \vq_{\text{start}} \nonumber \\
                & \quad \text{FK}(\vq(T)) = \HEEinWorld_{\text{goal}} \quad \left(\text{or} \; \vq(T) = \vq_{\text{goal}} \right) \nonumber \\
                & \quad \qvel(0) = \ddq(0) = \qvel(T) = \ddq(T) = \mathbf{0} \nonumber \\
                & \quad \text{within joint limits} \nonumber \\
                & \quad \text{no collisions}, \nonumber
\end{align}
where the costs $C_{\text{vel}}$ and $C_{\text{acc}}$ promote short and smooth paths, and are detailed in \cref{sec:motion_planning_costs}.
\reviewersix{
In this work we consider a fixed trajectory duration $T$, similarly to other motion planning algorithms~\cite{ratliff2009chomp,Kalakrishnan_RAIIC_2011_stomp,Mukadam2018-gpmp-ijrr}.
The optimization of minimum-time trajectories is left for future work.
}

\reviewerseven{As the optimization can become}
unfeasible while trying to satisfy the constraints, it is common practice to relax and include them in the cost function as soft constraints~\cite{ratliff2009chomp,zucker2013chomp,Mukadam2018-gpmp-ijrr}.
As we do batch optimization, we can filter the solutions for the ones respecting the constraints.
Hence, we solve
\begin{align}
    \argmin_{\trajectory} \sum_j \lambda_j C_j(\trajectory),
    \label{eq:trajectory_optimization_unconstrained}
\end{align}
where $\lambda_j$ are positive weights to balance the costs.
These costs are detailed in \cref{sec:motion_planning_costs}.

\subsection{Motion Planning as Inference}
\label{sec:motion_planning_as_inference}

Solving~\cref{eq:trajectory_optimization} with gradient-based methods relies on an initial trajectory.
In some cases, we would like the solution to remain close to the initialization to keep properties such as smoothness or if the initial trajectory is derived from human demonstrations.
Then, instead of a single particle, it is useful to describe a prior distribution over trajectories.
Hence, the motion planning problem can be formulated as inference~\cite{pmlr-vR4-attias03a, toussaint2009robusttrajopt}.
The goal is to either sample from or maximize the posterior distribution of 
trajectories given the task objective
\begin{align}
    p(\trajectory | \objective) \propto p(\objective | \trajectory) p(\trajectory)^{\lambda_{\text{prior}}},
    \label{eq:planning_as_inference}
\end{align}
where $p(\trajectory)$ is a prior over trajectories with temperature $\lambda_{\text{prior}}$, and $p(\objective | \trajectory)$ is the likelihood of achieving the task objectives.
A common assumption is that the likelihood factorizes as~\cite{urain_2022_learning_implicit_priors}
\begin{align}
    p(\objective | \trajectory) \propto \prod_{j} p_j(\objective_j | \trajectory)^{\lambda_j}
    \label{eq:objective_likelihood}
\end{align}
with $\lambda_j > 0$.
The costs relate to distributions by
$p_j(\objective_j |  \trajectory) \propto \exp(-C_j(\trajectory))$.
Then, performing Maximum-a-Posteriori (MAP) on the trajectory posterior
\begin{align}
  \label{eq:mp_objective}
  & \argmax_{\trajectory} \log  p(\objective | \trajectory) p(\trajectory)^{\lambda_{\text{prior}}}  \nonumber \\
  = & \argmax_{\trajectory} \sum_{j} \log \exp(-C_j(\trajectory))^{\lambda_j} + \log  p(\trajectory)^{\lambda_{\text{prior}}} \nonumber \\
  = & \argmin_{\trajectory}  \sum_{j} \lambda_j C_j(\trajectory) - \lambda_{\text{prior}} \log p(\trajectory) \nonumber
\end{align}
is equivalent to \cref{eq:trajectory_optimization_unconstrained} regularized with the prior.
Contrary to classical optimization-based motion planning, planning-as-inference has several advantages.
Notably, it provides a 
\reviewerten{principled}
way to introduce informative priors to planning problems, e.g., GPMP2~\cite{Mukadam2018-gpmp-ijrr} utilizes a GP to encode dynamic feasibility and smoothness.
Additionally, specialized methods can be employed to sample from the posterior. 
We use diffusion models as a prior over trajectories, and instead of computing the MAP solution, sample from the posterior to obtain a trajectory distribution.
Sampling from the posterior was done previously for GPMP with Gaussian approximations using variational inference, which are inherently unimodal~\cite{DBLP:journals/ral/YuC23}.
Instead, the diffusion framework provides an approach to obtain a multimodal distribution, which allows for a more expressive posterior and a natural way to treat stochasticity.

\subsection{Diffusion Models for Trajectory Priors}
\label{sec:diffusion_model_trajectory_priors}

Generative modeling methods, such as Generative Adversarial Networks (GAN)~\cite{goodfellow2014gan} and Variational Auto Encoders (VAE)~\cite{kingma2013vae}
are trained to maximize the data log-likelihood, and sampling is done by applying deterministic transformations to a random variate of an easy-to-sample distribution.
Instead, diffusion models~\cite{song2019generativemodeling,ho2020denoisingdiffusion}
perturb the original data distribution $\pdata(\vx)$ via a diffusion process to obtain noise, and learn to reconstruct it by denoising using the score-function $\grad_{\vx} \log \pdata (\vx)$.
In this work, we use Denoising Diffusion Probabilistic Models (DDPM)~\cite{ho2020denoisingdiffusion}.

We first consider an unconditional diffusion model.
Let $\trajectory_0$ be a sample from the data distribution $\trajectory_0 \sim q(\trajectory_0)$, which is transformed into Gaussian noise by a Markovian forward diffusion process
$q(\trajectory_{0:N}) = q(\trajectory_0) \prod_{i=1}^N q(\trajectory_i | \trajectory_{i-1}, i)$ with
\begin{align*}
    & q(\trajectory_{i} | \trajectory_{i-1}, i) = \Gaussian{\trajectory_{i}; \sqrt{1-\beta_i} \trajectory_{i-1}, \beta_i \mI} \\
    & q(\trajectory_N) \approx \Gaussian{\trajectory_N; \bm{0}, \mI},
\end{align*}
where ${i=1, \ldots, N}$ is the diffusion time step, $N$ is the number of diffusion steps, and $\beta_i$ is the noise scale at time step $i$.
Common schedules for $\beta$ are linear and cosine~\cite{ho2020denoisingdiffusion,nichol2021improvedddpm}.

The denoising process transforms noise back to the data distribution such that ${p(\trajectory_0) \approx q(\trajectory_0)}$
\begin{align*}
    & p(\trajectory_{0:N}) = p(\trajectory_N) \prod_{i=1}^{N} p(\trajectory_{i-1} | \trajectory_i, i), 
    \quad 
    p(\trajectory_N) = \Gaussian{\trajectory_N; \bm{0}, \mI}
\end{align*}
Sampling from $p(\trajectory_0)$ is done by first sampling from an isotropic Gaussian and sequentially sampling from the posterior distribution $p(\trajectory_{i-1} | \trajectory_i, i)$.
For $N \to \infty$ and $\beta_i \to 0$, this posterior converges to a Gaussian distribution~\cite{Sohl-Dickstein2015diffusion}.
Hence, during training, the goal is to learn to approximate a Gaussian posterior with parameters $\vtheta$ such that
\begin{equation}
    p_{\vtheta}(\trajectory_{i-1} | \trajectory_i, i) = \Gaussian{\trajectory_{i-1}; \vmu_i = \vmu_{\vtheta}(\trajectory_{i}, i), \mSigma_i} \approx p(\trajectory_{i-1} | \trajectory_i, i)
    \label{eq:denoising_process}
\end{equation}
For simplicity, only the mean is learned, and the covariance is set to
${\mSigma_i = \sigma_i^2 \mI = \betaposterior_i \mI}$,
with ${\betaposterior_i = (1-\alphacumprod_{i-1})/(1-\alphacumprod_i) \beta_i}$,
${\alpha_i=1-\beta_i}$ and 
${\alphacumprod_i = \prod_{k=0}^i \alpha_k}$~\cite{ho2020denoisingdiffusion}.
Instead of learning the posterior mean, we learn the noise vector $\vepsilon$, since
\begin{equation}
    \label{eq:ddpm_posterior_mean}
    \vmu_{\vtheta}(\trajectory_i, i) = \frac{1}{\sqrt{\alpha_i}} \left( \trajectory_i - \frac{1 - \alpha_i}{\sqrt{1-\alphacumprod_i}} \vepsilon_{\vtheta}(\trajectory_i, i) \right).
\end{equation}
$\vepsilon_{\vtheta}$ is typically implemented using a neural network, whose architecture we detail in~\cref{sec:implementation}.

The network parameters are learned by maximizing a lower bound on the expected data log-likelihood
$\E{\trajectory_0 \sim \pdata}{\log p_{\vtheta}(\trajectory_0)}$.
After simplifying,
we minimize
\begin{align}
    \label{eq:diffusion_loss}
    & \mathcal{L}(\vtheta) = \E{i, \vepsilon, \trajectory_0 }{ \| \vepsilon - \vepsilon_{\vtheta}(\trajectory_i, i) \|_2^2 } \\
    & i \sim \mathcal{U}(1, N), \; \vepsilon \sim \Gaussian{\bm{0}, \mI},
    \; \trajectory_0 \sim q(\trajectory_0), \; \trajectory_i \sim q(\trajectory_{i}|\trajectory_0, i) \nonumber.
\end{align}
For a complete derivation, we refer the readers to~\cite{luo2022understandingdiffusionmodels}.
Due to the Markov property, and given the data is in Euclidean space, the distribution of the noisy trajectory at time step $i$ given the original one is Gaussian and can be written in closed form as
$q(\trajectory_{i}|\trajectory_0, i) = \Gaussian{\trajectory_i; \sqrt{\alphacumprod_i}\trajectory_0, (1-\alphacumprod_i) \mI}$.
This allows for efficient training by sampling from $q(\trajectory_{i}|\trajectory_0, i)$ without running the forward diffusion process.
Intuitively, \cref{eq:diffusion_loss} states that the goal of training is to learn a denoising network that removes the noise added to the data sample at step $i$.

At inference, we obtain a sample $\Tilde{\trajectory}_0 \sim p_{\vtheta}(\trajectory_0)$ through a series of denoising steps from \cref{eq:denoising_process}.
In practice, to control stochasticity, we also pre-multiply the posterior variance, resulting in ${\alpha \mSigma_i}$~\cite{DBLP:conf/iclr/AjayDGTJA23}, with $\alpha \in [0, 1]$.
$\alpha=0$~means no noise and $\alpha=1$ is the DDPM sampling algorithm.
This is particularly important in our work because it lowers the chance that trajectories that are already collision-free come into collision due to large noise values.

To generate a trajectory distribution based on a context $\vc$, e.g., a start joint position and a desired end-effector pose, we extend to a conditional diffusion model $p_{0}(\trajectory | \vc)$, and thus
${p(\trajectory_{0:N} | \vc) = p(\trajectory_N) \prod_{i=1}^{N} p(\trajectory_{i-1} | \trajectory_i, \vc)}$.
The only change to the above is that now we learn a noise model with an additional $\vc$ input
$\vepsilon_{\vtheta}(\trajectory_i, i, \vc)$.
The expectation in \cref{eq:diffusion_loss} is modified to include the context distribution ${\vc \sim p(\vc)}$ and ${\trajectory_0 \sim q(\trajectory_0 | \vc)}$.

\subsection{Blending Sampling and Optimization}
\label{subsec:blending-sampling-optimization}

With a prior trajectory diffusion model and an optimality variable $\objective$ (e.g., representing collision avoidance), our goal is to sample from the posterior distribution
\begin{align}
    \label{eq:ddpm_posterior}
    p(\trajectory_0 | \objective) \propto p(\objective | \trajectory_0) p(\trajectory_0).
\end{align}
To show that this is equivalent to sampling from the prior while biasing the trajectories towards the high-likelihood regions, for completeness, we replicate here the proof from~\cite{Dhariwal2021diffusionbeatsgans}.
By definition of the Markovian reverse diffusion
\begin{equation}
    p(\trajectory_0 | \objective) = \int p(\trajectory_N | \objective) \prod_{i=1}^{N} p(\trajectory_{i-1} | \trajectory_i, i, \objective) \dd\trajectory_{1:N},
\end{equation}
where $p(\trajectory_N | \objective)$ is standard Gaussian noise by definition.
Hence, to sample from  $p(\trajectory_0 | \objective)$, we iteratively sample from the objective-conditioned posterior using Bayes' law
\begin{align}
    \label{eq:ddpm_posterior_step}
    p(\trajectory_{i-1} | \trajectory_i, i, \objective) \propto p(\objective | \trajectory_{i-1}) p(\trajectory_{i-1} | \trajectory_i, i),
\end{align}
where $p(\trajectory_{i-1} | \trajectory_i, i) \approx p_{\vtheta}(\trajectory_{i-1} | \trajectory_i, i)$ is the approximate learned prior, and $p(\objective | \trajectory_{i-1}) = p(\objective | \trajectory_{i-1}, \trajectory_i, i)$, since due to the Markov property in diffusion, $\objective$ and $\trajectory_{i}$ are conditionally independent given $\trajectory_{i-1}$.
The objective-conditioned posterior cannot be sampled in closed-form,
but given the learned denoising prior model over trajectories is Gaussian, its logarithm equates to
\begin{align}
    \label{eq:logprior}
    \log p_{\vtheta}(\trajectory_{i-1} | \trajectory_i, i) & = \log \Gaussian{\trajectory_i; \vmu_i = \vmu_{\vtheta}(\trajectory_i, i), \mSigma_i} \\
    & \propto -\frac{1}{2} (\trajectory_{i-1} - \vmu_i)\tran \mSigma_i^{-1} (\trajectory_{i-1} - \vmu_i) \nonumber.
\end{align}
By definition of the noise schedule $\beta$, as the denoising step approaches zero, so does the noise covariance ${\lim_{i \to 0} \| \mSigma_i \| = 0}$.
Therefore, $p_{\vtheta}(\trajectory_{i-1} | \trajectory_i, i)$ concentrates its mass close to the mean $\vmu_i$, and the task log-likelihood is approximated with a first-order Taylor expansion around $\vmu_i$
\begin{align}
    \label{eq:logtaskliklelihood}
    & \log p(\objective | \trajectory_{i-1}) \approx \log p(\objective | \trajectory_{i-1} = \vmu_i ) + (\trajectory_{i-1} - \vmu_i)\tran \vg \\
    & \text{with} \quad \vg = \grad_{\trajectory_{i-1}} \log p(\objective | \trajectory_{i-1} = \vmu_i). \nonumber
\end{align}
Combining~\cref{eq:logprior} and~\cref{eq:logtaskliklelihood} we obtain
\begin{align}
    & \log p(\trajectory_{i-1} | \trajectory_i, i, \objective)
    \\
    & \propto -\frac{1}{2} (\trajectory_{i-1} - \vmu_i)\tran \mSigma_i^{-1} (\trajectory_{i-1} - \vmu_i) + (\trajectory_{i-1} - \vmu_i) \vg \nonumber \\
    & \propto -\frac{1}{2} (\trajectory_{i-1} - \vmu_i - \mSigma_i \vg)\tran \mSigma_i^{-1} (\trajectory_{i-1} - \vmu_i - \mSigma_i \vg) \nonumber \\
    & = \log p(\vz = \trajectory_{i-1}),
    \quad p(\vz) = \mathcal{N}(\vz; \underbrace{\vmu_i + \mSigma_i \vg}_{\vmu_z}, \mSigma_i). \label{eq:ddpm_guided_diffusion}
\end{align}
In motion planning-as-inference we have from \cref{eq:objective_likelihood}
\begin{align*}
    \vg & = \grad_{\trajectory_{i-1}} \log p(\objective | \trajectory_{i-1}) = - \sum_{j} \lambda_j \grad_{\trajectory_{i-1}} C_j(\trajectory_{i-1} = \vmu_i),
\end{align*}
where the costs are assumed to be differentiable w.r.t. the trajectory.
Hence, sampling from the task-conditioned posterior is equivalent to sampling from a Gaussian distribution with mean and covariance as $p(\vz)$.
At every denoising step, we sample from the prior, move the particle toward low-cost regions, and repeat this process until $i=0$.
This formulation benefits us by allowing us to use gradient-based techniques to guide the sampling process while running the denoising process for the prior.
The updated mean of $\vz$ in \cref{eq:ddpm_guided_diffusion} can be interpreted as taking a single gradient step starting from $\vmu_i$, but several aspects are crucial for making guided sampling work in practice~\cite{DBLP:conf/iclr/MaHWS24}.

First, the entries in $\mSigma_i$ approach $0$ with $i \to 0$.
Hence, the guidance function vanishes towards the end of the denoising process.
This is desirable if the prior covers a very large space of the data manifold where the objective likelihood is high.
However, if there is no (or little) data in those regions, the generative model might be unable to move samples towards high objective likelihood regions.
To counteract this and keep the influence of the task likelihood, we drop the covariance $\mSigma_i$.
This is equivalent to changing the cost weights per denoising step $\lambda_j \coloneqq \mSigma_i^{-1} \lambda_j$, which keeps the likelihood relevancy.

Second, instead of performing one gradient step, we take $M$ gradient steps and choose the step size such that the optimized mean of $\vz$ is within a small deviation $\delta$ of the prior mean as motivated 
in~\cite{DBLP:conf/icra/ZhongRXCVCRP23}.
This is equivalent to solving
\begin{align}
    \min_{\vmu_z} & \quad \sum_{j} \lambda_j C_j (\vmu_z), \quad \st \;  | \vmu_z - \vmu_i | \leq \delta  \label{eq:guidance_inner_gradient},
\end{align}
using first-order gradients starting from $\vmu_i$, and $\delta$ prevents the optimization from moving \textit{too} far from the prior, which could lead to a distribution shift.

Third, recall from \cref{eq:planning_as_inference} that we can control the influence of the prior by sampling from
${
p(\trajectory_0 | \objective) \propto p(\objective | \trajectory_0) p(\trajectory_0)^{\lambda_{\text{prior}}}
}$.
In image generation, typically $\lambda_{\text{prior}}=1$, and the classifier weight $\lambda_j$ is increased to generate images belonging to a desired class.
This balance works well because image models are trained with large amounts of data, and the desired class is covered in the prior distribution.
For instance, when generating images of dogs, we steer particles towards regions of the prior that include dogs.
In motion planning, we want to obtain collision-free trajectories in \textit{new environments}.
This means our prior might not have samples in the new collision-free region.
Hence, we treat $\lambda_{\text{prior}}$ as a hyperparameter to control the influence of the prior.
From the relation between the score function and the predicted noise we have~\cite{song2021scorebased}
\begin{align}
    & \grad_{\trajectory_i} \log p(\trajectory_i) = -\frac{1}{\sqrt{1-\alphacumprod_i}} \vepsilon(\trajectory_i, i) \propto \vepsilon(\trajectory_i, i) \label{eq:score_function_noise_vector} \\
    & \grad_{\trajectory_i} \log p(\trajectory_i)^{\lambda_{\text{prior}}} = \lambda_{\text{prior}} \grad_{\trajectory_i} \log p(\trajectory_i) \propto \lambda_{\text{prior}} \vepsilon(\trajectory_i, i) \nonumber.
\end{align}
Therefore, we replace $\vepsilon_{\vtheta}(\trajectory_i, i)$ in \cref{eq:ddpm_posterior_mean} with $\lambda_{\text{prior}} \vepsilon_{\vtheta}(\trajectory_i, i)$.

\subsubsectionheading{Accelerated sampling.}
A standard sample method uses the same number of time steps as the ones from training, which leads to a slow sampling process.
However, the initial steps of the denoising process include regions with large noise levels, and it was shown that several steps can be discarded by making the diffusion model non-Markovian, using Denoising Diffusion Implicit Models (DDIM)~\cite{DBLP:conf/iclr/SongME21}.
Instead of sampling with $N$ steps as in DDPM, we select a subset of time indices $i \in M = \{1, \ldots, N\}$, with typically $|M| \ll N$.
To adapt DDIM to cost guidance, to sample from the posterior in \cref{eq:ddpm_posterior_step}, we make use of \cref{eq:score_function_noise_vector} and write
\begin{align}
    & \grad_{\trajectory_{i-1}} \log p(\trajectory_{i-1} | \trajectory_i, i, \objective) \\
    & \propto \grad_{\trajectory_{i-1}} \log p(\objective | \trajectory_{i-1}) + \grad_{\trajectory_{i-1}} \log p(\trajectory_{i-1} | \trajectory_i, i) \label{eq:ddim_posterior_guidance}  \\
    & = -\frac{1}{\sqrt{1-\alphacumprod_i}} \vepsilon(\trajectory_i, i) + \grad_{\trajectory_{i-1}} \log p(\objective | \trajectory_{i-1}) \nonumber \\
    & = -\frac{1}{\sqrt{1-\alphacumprod_i}} \underbrace{ \left( \vepsilon(\trajectory_i, i) - \sqrt{1-\alphacumprod_i}  \grad_{\trajectory_{i-1}} \log p(\objective | \trajectory_{i-1}) \right)}_{ = \vepsilon}, \nonumber
\end{align}
and replace $\vepsilon$ in the original DDIM equations~\cite{Dhariwal2021diffusionbeatsgans}.
In practice, we note that, similarly to DDPM, ${\sqrt{1-\alphacumprod_i} \to 0}$ with ${i \to 0}$, and therefore remove this term to maintain the influence of the cost function by adjusting the weights ${\lambda_j \coloneqq (\sqrt{1-\alphacumprod_i} )^{-1}} \lambda_j$.

\subsubsectionheading{A note on classifier-free guidance.}
An orthogonal approach to introduce guidance in diffusion models is to use classifier-free guidance~\cite{DBLP:conf/iclr/AjayDGTJA23}.
This method works by embedding information from the objective function (or constraints) into a context variable.
This is not as flexible since the constraints need to be known during training.
On the contrary, cost guidance can easily integrate new constraints via differentiable cost functions, which makes it easier to adapt to new problems.

\subsection{Trajectory Parametrization}
\label{sec:trajectory_parametrization}

Until this point we did not specify how the trajectory $\trajectory$ is parametrized.
In previous work~\cite{DBLP:conf/iros/Carvalho0BK023}, the diffusion prior over trajectories was represented as a time-discretized vector of joint positions
${\trajectory = [\vq_0,\ldots,\vq_{H-1}] \in \sR^{H \times d}}$, where $H$ is the number of waypoints.
Given a planning frequency $1/\Delta t$, the trajectory duration is $T = H\Delta t$.
On the positive side, the waypoint representation ensures the trajectory passes on the waypoints, which is useful when precise path-following is required, especially for tasks that require strict position control.
Waypoints can be placed freely, making them suitable for representing paths with sharp turns or discontinuities.
However, velocities and accelerations were computed with finite differences, which did not provide smoothness guarantees.
A cost function/regularizer (the GP prior cost) was needed to enforce smoothness, which needed extra hyperparameter tuning.
Moreover, if part of a trajectory is in collision and the corresponding waypoints are moved, then an already smooth trajectory might become non-smooth, so a balance between the two costs needs to be accounted for.

In turn, in this work, we propose using a B-spline parametrization instead~\cite{DBLP:journals/trob/KickiLTBWSP24}.
There are several properties of B-splines that are interesting for our application.
As B-splines are smooth and have continuous derivatives, if part of the trajectory that is in collision is moved, then the overall trajectory remains smooth.
This representation has a locality property, which means that modifying control points only affects neighbor trajectory segments.
Additionally, as denoising is computationally expensive due to the iterative denoising process, a lower-dimensional parametrization with fewer parameters than $H \times d$ is desirable.
In B-splines, the parametrization is $n_b \times d$, where $n_b$ is the number of control points, and $n_b \ll H$.
For these reasons, we choose B-splines as a trajectory representation.

\begin{figure}[!t]
  \centering
  \subfloat[Control points and curve]{%
    \includegraphics[height=0.115\textheight,keepaspectratio]{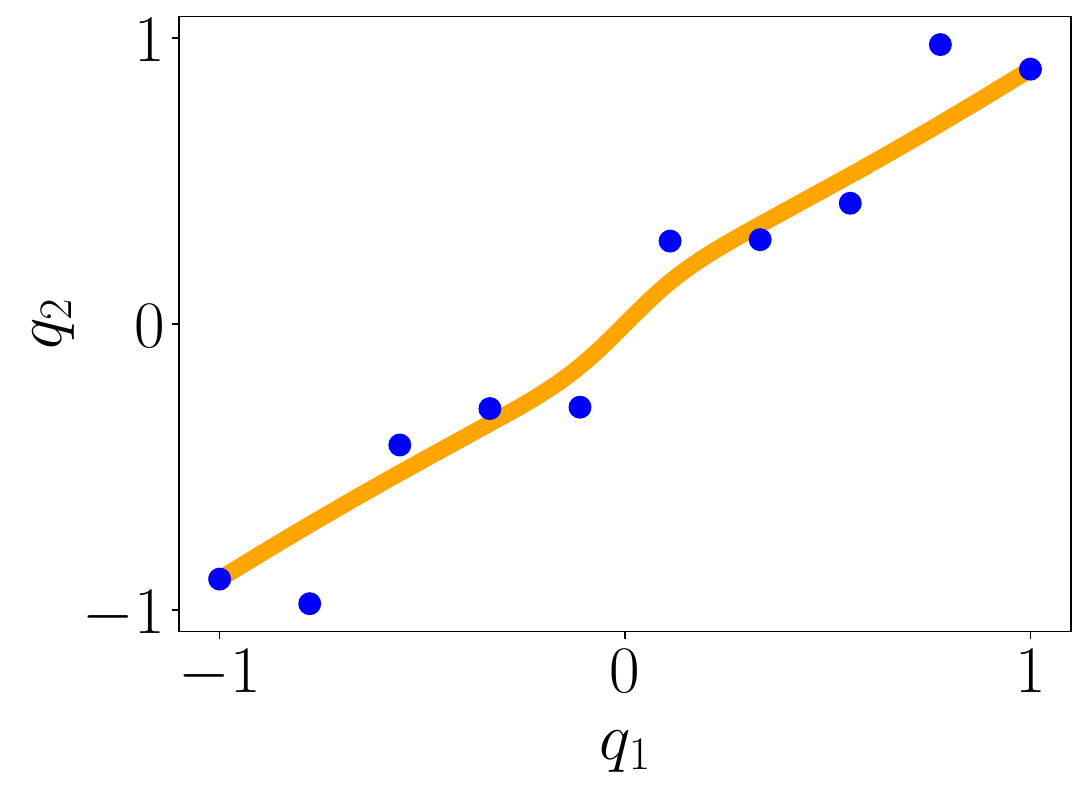}
    \label{fig:phase_time_control_points}
    }
    \hfill
  \subfloat[Trajectories in time]{%
    \includegraphics[height=0.115\textheight,keepaspectratio]{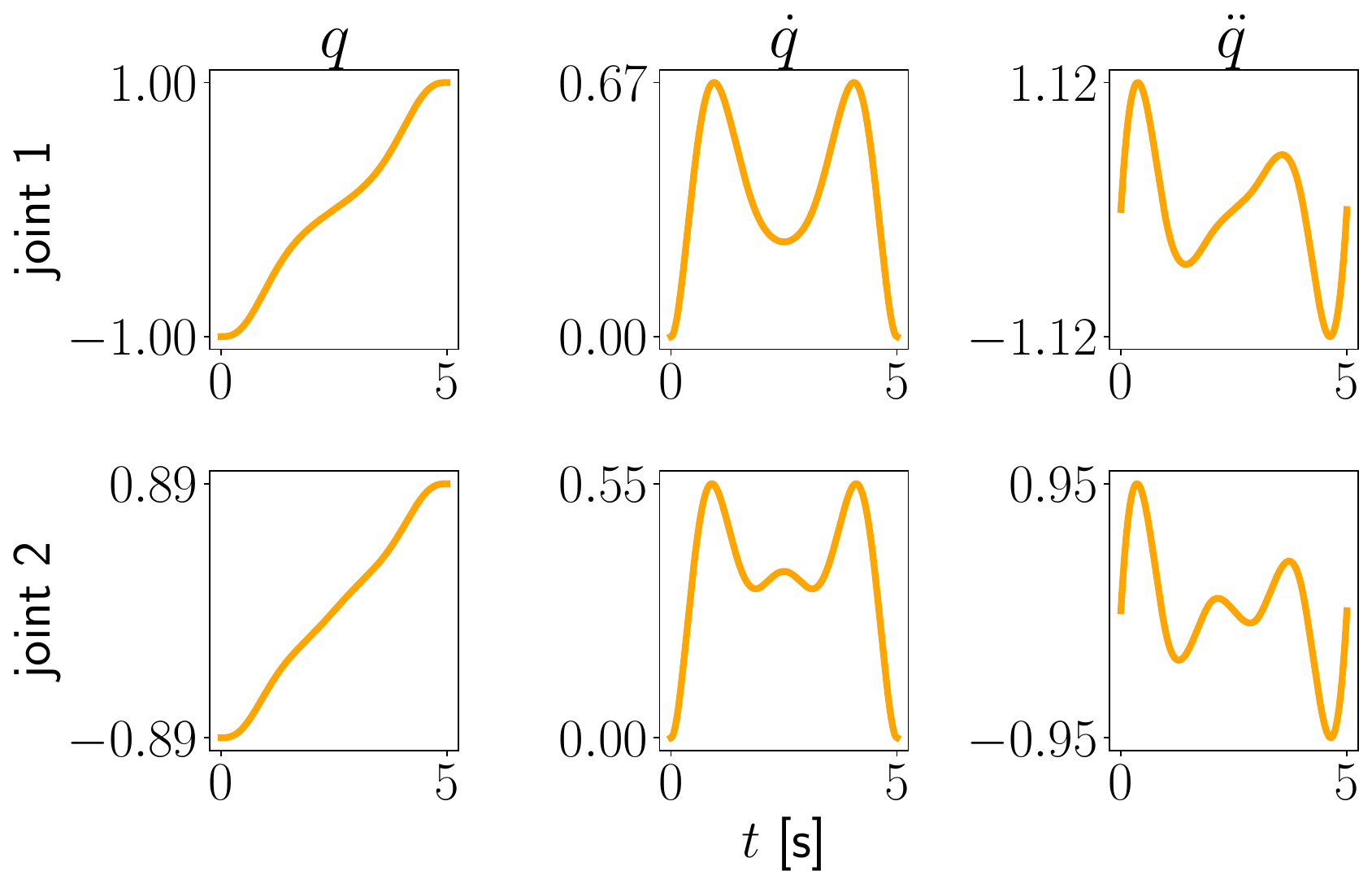}
    \label{fig:phase_time_q_dq_ddq}
    }
    \hfill
  \\
  \hfill
  \caption[Illustration of linear phase-time scaling for B-splines trajectories.]{
    The figures show the resulting trajectories when using linear phase-time scaling.
    In \protect\subref{fig:phase_time_control_points}, the control points (blue dots) and path of a clamped $5$th order B-spline are displayed.
    \protect\subref{fig:phase_time_q_dq_ddq} shows the trajectory in time of positions, velocities, and accelerations for the two degrees of freedom.
  }
  \label{fig:phase_time}
  \vspace{-0.6cm}
\end{figure}

A B-spline joint position trajectory as a function of a phase variable $s \in [0, 1]$ is defined as a linear combination of $n_b$ basis-splines~\cite{deBoorBsplines}
\begin{equation}
    \label{eq:bspline}
    \vq(s) = \sum_{i=0}^{n_b-1} B^{\vu}_{i, p}(s) \vw_i = \mB(s) \vw,
\end{equation}
where
$B_{i,p}^{\vu} \in \R$ is a basis-spline of degree $p \in \sZ^+$ with support between knots $u_i$ and $u_{i+p+1}$ defined in a knot vector $\vu=[u_0, \ldots, u_m]$, with $m \geq n_b-p-1$ and non-decreasing entries,
${\mB(s) = [B^{\vu}_{0,p}(s) \ldots B^{\vu}_{n_b-1,p}(s)] \in \R^{1 \times n_b}}$,
${\vw_i \in \R^{1 \times d}}$ and ${\vw = [\vw_0 \ldots \vw_{n_b-1}]\tran \in \R^{n_b \times d}}$ are the B-spline control points (or coefficients).
Note that with a fixed basis $\mB_p$, the control points are the degrees of freedom and completely define the B-spline.
We learn a diffusion model over $\vw$.
The basis functions are defined recursively with De Boor's algorithm~\cite{deBoorBsplines}.
With $s \in [0,1]$, we limit the knots vector to be in the same range, and to distribute the basis functions evenly, we set subsequent knots to be equidistant~\cite{DBLP:journals/trob/KickiLTBWSP24}
\begin{equation}
\label{eq:knots}
    \vu = \left[ \underbrace{0 \ldots 0 \vphantom{\frac{1}{n_b-p}}}_{p+1 \, \text{times}} \underbrace{\frac{1}{n_b-p} \ldots \frac{n_b-p-1}{n_b-p}}_{n_b-p-1 \, \text{elements}} \underbrace{1 \ldots 1 \vphantom{\frac{1}{n_b-p}}}_{p+1 \, \text{times}} \right],
\end{equation}
such that $|\vu| = n_b+p+1$.
To ensure the boundary constraints on start and final positions, zero velocities, and accelerations as in \cref{eq:trajectory_optimization}, we repeat boundary knots (see~\cref{eq:knots} and set the first and last control points as~\cite{mtuDerivativesBspline}
\begin{align}
    & \vw_0 = \vw_1 = \vw_2 = \vq_{\text{start}} \label{eq:control_points_first} \\
    & \vw_{n_b-1} = \vw_{n_b-2} = \vw_{n_b-3} = \vq_{\text{goal}} \label{eq:control_points_last}.
\end{align}
\reviewersix{
Note that if $\vq_\text{goal}$ is not specified, but rather a desired end-effector pose $\HEEinWorld_{\text{goal}}$, then the last control point $\vw_{n_b-1}$ is generated by the diffusion model.
}
To accelerate computations, similarly 
\reviewerseven{to}~\cite{DBLP:journals/trob/KickiLTBWSP24}, we discretize the phase variable into $n_s$ steps 
and pre-compute a matrix ${\mB \in \R^{n_s \times n_b}}$ of B-splines basis,
where $n_s$ is equivalent to $H$ in the dense waypoint representation.
A joint position trajectory in phase-space is given by $\mQ = \mB \vw \in \R^{n_s \times d}$.

A trajectory in time is obtained by transforming the phase variable.
Let $f$ be a monotonically increasing function such that $t = f(s)$, with $f(0) = 0$ and $f(1) = T$, where $T$ is the trajectory duration.
Hence, we have $\vq(t) = \vq(f(s))$.
Therefore, the position trajectory is completely defined by the B-spline in phase space, and with slight notation abuse, we write
$\vq(t) \triangleq \vq(s)$.
Let the derivative of the phase w.r.t.~time be a function of the phase variable~\cite{DBLP:journals/trob/KickiLTBWSP24,kicki2024bridging}
\begin{equation}
    \label{eq:bspline_rs}
    r(s) = \frac{\dd s}{\dd t} (s) = \left( \frac{\dd t}{\dd s} \right)^{-1} (s).
\end{equation}
The first and second derivatives of $\vq$ w.r.t.~time, velocity and acceleration, respectively, can be computed as
\begin{align}
    \dot{\vq}(t) & = \dv{\vq(t)}{t} = \pdv{\vq(s)}{s} \dv{s}{t} = \pdv{\vq(s)}{s} r(s) \label{eq:bspline_derivative_time_1} \\
    \ddot{\vq}(t) & = \dv[2]{\vq(t)}{t} = \dv{}{t}\left( \pdv{\vq(s)}{s} \right) r(s)  + \pdv{\vq(s)}{s} \dv{}{t} r(s) \nonumber \\
    & = \pdv[2]{\vq(s)}{s}\, (r(s))^2 + \pdv{\vq(s)}{s} \pdv{r(s)}{s} r(s) \nonumber
\end{align}
These expressions compute the trajectory in time using the phase variable and their derivative relation.
For completeness, we must compute the derivatives of B-spline curves w.r.t.~the phase variable and define the phase-time function.

The $k$-th order derivative $\partial \vq(s)^{(k)} / \partial s^{(k)}$ is also a \mbox{B-spline}~\cite{DBLP:books/daglib/0079701,mtuDerivativesBspline}, which can be written as
\begin{align}
    & \pdv{\vq}{s}\, (s) = \sum_{i=0}^{n_b-2} {B'}^{\vu}_{i, p}(s) \vw_i \label{eq:bspline_dq_ds_A}
    \\
    & {B'}^{\vu}_{i,p}(s) \triangleq p \left( \frac{B_{i,p-1}(s)}{u_{i+p} - u_i} - \frac{B_{i+1,p-1}(s)}{u_{i+p+1} - u_{i+1}} \right).
\end{align}
Note that ${B'}^{\vu}_{i,p}$ is not a proper basis.
Higher order derivatives can be computed similarly ${\partial^2 \vq / \partial s^2 = \sum_{i=0}^{n_b-3} {B''}^{\vu}_{i, p}(s) \vw_i}$,
and the new basis matrices can be pre-computed for a fixed number of steps $n_s$.

One of the benefits of phase-time decoupling is being able to speed up and slow down movements.
As common in the literature~\cite{DBLP:journals/arobots/ParaschosDPN18}, we assume a linear relation, but nonlinear functions can also be used~\cite{DBLP:journals/trob/KickiLTBWSP24}
\begin{align*}
    & s = f^{-1}(t) = t/T, \quad r(s) = 1/T, \quad \pdv{r(s)}{s} = 0 \nonumber \\
    & \dot{\vq}(t) = \pdv{\vq(s)}{s} \frac{1}{T}, \quad  \ddot{\vq}(t) = \pdv[2]{\vq(s)}{s} \frac{1}{T^2}. \nonumber
\end{align*}

\Cref{fig:phase_time} illustrates the properties of the B-spline trajectory representation in $2$-dimensional example.
This parametrization allows for a low-dimensional representation of smooth trajectories, which are useful for motion planning.

\subsubsectionheading{Other trajectory representations}
The definition in \cref{eq:bspline} encompasses other types of movement primitives by modifying the basis functions, such as
B\'{e}zier curves~\cite{DBLP:journals/cad/Tiller04}, Probabilistic Movement Primitives (ProMP)~\cite{DBLP:journals/arobots/ParaschosDPN18} or Probabilistic Dynamic Movement Primitives (ProDMP)~\cite{DBLP:journals/ral/LiJVOLN23}.
However, some properties of B-splines are suited to our tasks.
Unlike other spline methods such as B\'{e}zier curves, the B-spline has a \textit{local property} instead of the global property~\cite{DBLP:journals/cad/Tiller04}.
This property can be important because the collision cost is local in gradient-based optimization.
By moving control points locally, we ensure that only the part of the trajectory that is in collision is affected.
Compared to ProMPs and ProDMPs, B-spline trajectories lie within the convex hull defined by the control points, which allows for the definition of constraints on the coefficients instead of constraints on the whole trajectory.
Moreover, commonly, these movement primitives are defined by individual basis functions with support in $\R$, such as an exponential kernel, making it difficult to set boundary conditions for position, velocity, and acceleration~\cite{kicki2024bridging}.
Establishing these constraints is easier with \mbox{B-splines} due to the basis being defined over closed intervals determined by the knots vector.

\subsection{Implementation Details}
\label{sec:implementation}

For learning, we assume having a dataset of pairs of $C$~contexts and $N_C$~(possibly) multimodal trajectories per context ${\mathcal{D} = \{\{(\vc_j, \trajectory_{jk})\}_{k=1}^{N_C}\}_{j=1}^{C}}$.
Following \cref{sec:diffusion_model_trajectory_priors}, we learn a prior over trajectories by learning the conditioned denoising model
$\vepsilon_{\vtheta}(\vw_i, i, \vc)$,
where $\vw_i \in \R^{n_b \times d}$ are the control points of a B-spline.
To ensure zero velocities and accelerations at the boundaries, the first and last control points are fixed according to \cref{eq:control_points_first,eq:control_points_last}.
Hence, the diffusion model uses the inner control points of size $n_b - 6$, or $n_b - 5$ if the end-effector goal pose is used, meaning the last joint configuration is not fixed.
The conditioning variable $\vc$ consists of the current (start) joint position, and if a goal joint position is defined, ${\vc = [\vq_{\text{start}}, \vq_{\text{goal}}]}$,
otherwise, if a desired end-effector goal is provided, then $\vc = [\vq_{\text{start}}, \HEEinWorld_{\text{goal}}]$.
The end-effector pose input to the network consists of the position concatenated with a flattened rotation matrix.
We separate these two conditioning cases because the latter option is more natural for humans to specify.
For instance, in a pick-and-place task, we want the manipulated object to be placed in a certain pose and might not care too much about the final joint positions.
Moreover, by specifying a task space pose, we might get multiple robot configurations that achieve the same end-effector pose, which gives the user the freedom to decide which configuration to choose.
However, if the user has access to an inverse kinematics solver, for instance, to bias the solutions to a neutral configuration, then she can use conditioning with goal joint positions.
Nevertheless, with our framework, biasing the solution could also be achieved by adding another cost function.

\citet{janner2022diffuser} proposed a temporal U-Net architecture with \mbox{$1$-dimensional} convolutions to encode local relations between states and actions, acting as a low-pass filter that prevents jumps between neighboring states.
Due to its success in modeling trajectories, we adapt it and replace states with \mbox{B-spline} control points and extend this network architecture to include a conditioning variable, along with the diffusion step index, using Feature-wise Linear Modulation (FiLM)~\cite{DBLP:conf/aaai/PerezSVDC18} (see \cref{fig:mpd_method_overview}).
Additionally, all data is normalized to $[-1, 1]$.

For sampling, we tested both DDPM and DDIM algorithms.
In DDIM, the number of steps $|M|$ is commonly selected linearly from $\{1, \ldots, N\}$.
With this schedule, several steps closer to $N$ add too much noise to samples during the denoising process.
We can skip those first steps by choosing quadratically distributed steps, which means the denoising process jumps from larger indexes to ones closer to $0$, leading to fewer denoising steps.
Moreover, we use the deterministic version of DDIM with $\gamma=0$, which slightly prevents the risk of adding noise to a trajectory that is already collision-free.
During inference, cost guidance is computed w.r.t. the $\trajectory_i$ at denoising step $i$.
However, for higher noise levels, the trajectory is still quite noisy, and this gradient produces little effect.
Hence, we only apply the cost gradients on the last $i_{\text{cost}}$ steps of the denoising process.

We run our algorithm by sampling and optimizing a batch of trajectories in parallel.
Therefore, to sample many trajectories, evaluate their costs, and compute gradients in our framework, we maximize parallelization by implementing all components in PyTorch~\cite{pytorch2019} by leveraging GPU utilization.
Although a batch of trajectories is used for planning, in the real world, only one trajectory can be executed.
Choosing this trajectory is task-dependent and a user choice.
For tasks where the context includes reaching a desired end-effector pose, one could choose the trajectory corresponding to the lowest end-effector pose error, and for other tasks, one could use the minimum length trajectory or other combinations.

\subsection{Motion Planning Costs}
\label{sec:motion_planning_costs}

In this section, we describe the motion planning costs used.
The unconstrained cost function from \cref{eq:trajectory_optimization_unconstrained} is computed with the trajectory time-integral costs and can be converted into an integral in phase-space with a change of variables
{\small
\begin{align}
    C(\trajectory) & = \sum_{j} \lambda_{j} \int_0^{T} C_j(\trajectory(t)) \dd t
    = \sum_{j} \lambda_{j} \int_0^{1} C_j(\trajectory(s)) r(s)^{-1} \dd s  \label{eq:cost_trajectory}
\end{align}
}
where the integral is approximated by discretizing the phase variable into $n_s$ segments (\cref{sec:trajectory_parametrization}). 

The costs we consider are computed in joint and task space, and the total derivative w.r.t. the B-spline trajectory parametrization $\vw$ (the control points) is computed using the chain rule, e.g., for joint position costs
\begin{align}
    \dv{C_j(\vq)}{\vw} = \pdv{C_j(\vq)}{\vq} \dv{\vq(s)}{\vw} \label{eq:cost_trajectory_gradient_wrt_w},
\end{align}
where $C_j \in \R$, $\partial C(\vq) / \partial \vq \in \R^{1 \times d}$, and for \mbox{B-splines} $\dd \vq / \dd \vw = \mB \in \R^{d \times n_b \times d}$, where the basis matrix is repeated along the degrees-of-freedom to match dimensions.
If $C_j$ is a task space cost, let $\mX_m = \text{FK}_m(\vq) \in \SEthree$ be the pose of a link $m$ on the robot arm computed with forward kinematics, and consider a cost $C_j(\mX_m)$.
Then we have
\begin{align}
    \pdv{C_j(\vq)}{\vq} = \pdv{C_j(\mX_m)}{\mX_m} \pdv{\mX_m(\vq)}{\vq}, \nonumber
\end{align}
where $\partial C_j(\mX_m) / \partial \mX_m \in \R^{1 \times 6} \in \sethree$ is a vector in the Lie algebra of \SEthree, and
\begin{align}
   \partial \mX_m(\vq) / \partial \vq \triangleq \mJ_m(\vq) = 
   \begin{bmatrix}
       \mJ^p_m(\vq) \\
       \mJ^o_m(\vq)
   \end{bmatrix}
   \in \R^{6 \times d}
   \label{eq:fk_jacobian}
\end{align}
is the forward kinematics geometric Jacobian at link $m$, for position $p$ and orientation $o$.
This formulation is important because we do not need to use automatic differentiation for gradient computations, and make use of the library Theseus~\cite{DBLP:conf/nips/PinedaFMVSCODWA22} to compute geometric Jacobians and Lie algebra quantities without backpropagation through the computational graph.
This separation exploits the structure of the Jacobian of kinematic chains, allowing faster and more stable computation, as well as the separation of position and orientation costs.
Moreover, due to the B-spline linear parametrization, gradient computation can be easily parallelized in the GPU.

\subsubsectionheading{Velocity cost.}
We minimize the joint space velocity trajectory using a squared penalty, which promotes straight paths by using a squared penalty on the velocity trajectories
\begin{align*}
    C_{\text{vel}} (\trajectory(s)) = \frac{1}{2} \| \vq'(s) \|_2^2,
\end{align*}
where $\vq'(s)$ is the first-order derivative of $\vq$ w.r.t. $s$.

\subsubsectionheading{Acceleration cost.}
To prevent sharp turns in the robot trajectory, we maximize smoothness by using a squared penalty on the acceleration trajectories
\begin{align*}
    C_{\text{acc}} (\trajectory(s)) = \frac{1}{2} \| \vq''(s) \|_2^2,
\end{align*}
where $\vq''(s)$ is the second-order derivative of $\vq$ w.r.t. $s$.

\subsubsectionheading{Task cost.}
In several manipulation tasks, such as pick-and-place or pouring, it is much more intuitive to specify a desired end-effector goal pose instead of a goal joint position.
Hence, we minimize the end-effector pose error obtained at the last joint position ($s=1$) with
\begin{align*}
    C_{\text{task}}(\trajectory(s))\rvert_{s=1} = d_{\SEthree} \left( \HEEinWorld_{\text{goal}}, \, \text{FK}_{EE}(\vq(1)) \right).
    \label{eq:cost_task}
\end{align*}
$ d_{\SEthree}$ defines the distance between two elements of $\SEthree$.
Given two poses (homogeneous transformations) ${\mT_1 = [\mR_1, \vp_1]} \in \SEthree$ and ${\mT_2 =[\mR_2, \vp_2]} \in \SEthree$, consisting of a translational and rotational part, we choose
\begin{equation*}
    d_{\SEthree}(\mT_1, \mT_2) = \frac{1}{2} \|\vp_1 - \vp_2\|_2^2 + \frac{1}{2} \|\text{LogMap}(\mR_1\tran\mR_2)\|_2^2,
\end{equation*}
where $\text{LogMap}(\cdot)$ is the operator that maps an element of the Lie group $\SOthree$ to the vector representation of its tangent space at the identity element, the Lie algebra $\sothree$~\cite{sola2018micro}.
With the Jacobian decomposition from \cref{eq:fk_jacobian}, we can map the derivatives of $d_{\SEthree}$ w.r.t.~position and orientation task space errors to the joint space.

\subsubsectionheading{Collision costs.}
The task-space signed distance function of an environment $\sdf (\vx)$ is the 
\begin{wrapfigure}{r}{0.63\columnwidth}
  \centering
  \includegraphics[width=0.28\columnwidth]{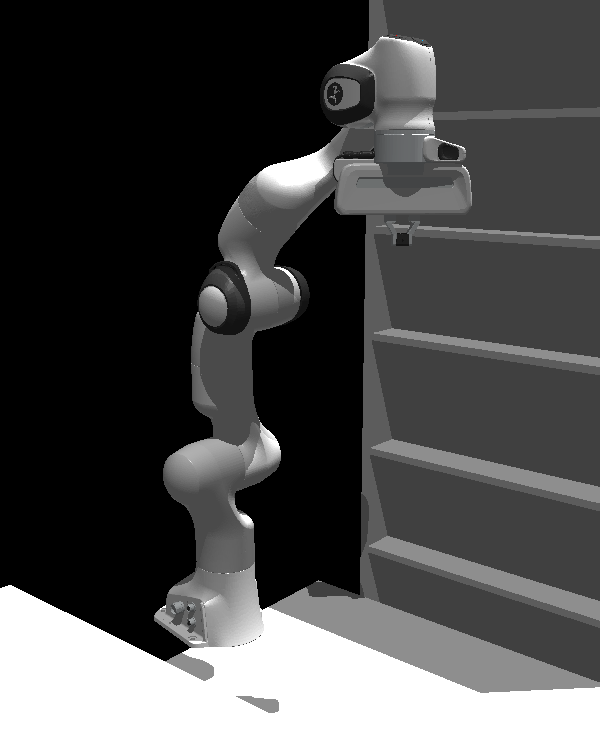}
  \includegraphics[width=0.28\columnwidth]{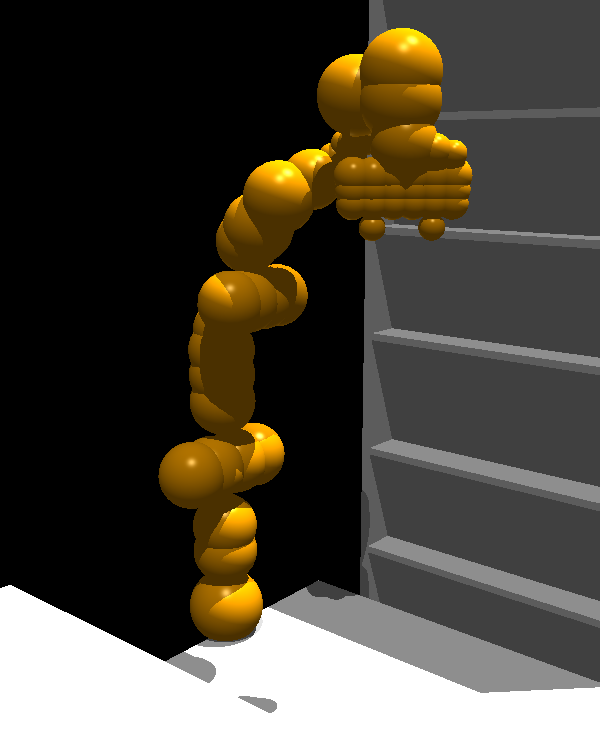}
  \hfill
  \caption[Visualization of the Franka Emika Panda finite collision spheres model.]{
    Visualization of the Franka Emika Panda finite collision spheres model (right) used for faster collision cost computations.
  }
  \label{fig:robot_collision_spheres}
  \vspace{-0.5cm}
\end{wrapfigure}
smallest signed Euclidean distance between a point in space $\vx \in \R^3$ and the closest surface (negative if inside an obstacle, and positive otherwise).
Given an environment with obstacles, we precompute and store the \sdf~using a fine voxel grid representation and project a continuous point $\vx$ to the nearest point in the grid.
The \sdf~representation is a common assumption in motion planning~\cite{ratliff2009chomp,Mukadam2018-gpmp-ijrr}, which can be obtained fast in the real world using newer software modules~\cite{DBLP:conf/icra/MillaneOWSRTS24}.
Also, if a new object is added to the scene, and we know its \sdf, the resulting \sdf~is the minimum between the environment and the new object \sdf.
Similarly, we can precompute the \sdf~gradient $\nabla_{\vx} \sdf(\vx)$, which has L$2$-norm equal to $1$.
This gradient indicates a direction to push the robot out of collision when the $\sdf(\vx) < 0$.
To compute robot collisions, we represent it with $S$ spheres along its body $\mathcal{B}$ (see \cref{fig:robot_collision_spheres}), allowing for faster computation of the \sdf~between the robot and the environment compared to using a full mesh.
We consider two types of collisions: with the environment and self collisions.
The environment collision cost is defined as
\begin{align}
    C_{\text{coll-env}}(\trajectory(s)) & = \int_{\mathcal{B}} C_{\text{env}}(\vx_{m}(\vq(s))) \dd m \approx \sum_{m=0}^{S-1} C_{\text{env}}(\vx_{m})  \nonumber
\end{align}
with 
$
C_{\text{env}}(\vx_m) = \relu(-\sdf(\vx_m) + r_m + \epsilon)
$
, where $\vx_m \in \R^3$ is the position of the $m$th sphere center computed with forward kinematics at configuration $\vq(s)$ and $r_m$ its radius.
To prevent (and control) the robot passing \textit{too} close to objects, we use a safety margin $\epsilon \geq 0$ and activate the collision cost only if the robot is inside the safety margin.

The self-collision is similarly implemented,
except the cost is computed between selected pairs of links
\begin{equation}
    C_{\text{self}}(\trajectory(s)) = \max_{i, j \in \mathcal{S}_c} \left( \text{ReLU} \left( -\| \vx_i - \vx_j \| + r_i + r_j + \epsilon \right) \right) \nonumber,
\end{equation}
where $S_c$ is the set containing all pairs of spheres of centers $\vx_i$ and $\vx_j$, and radius $r_i$ and $r_j$, considered for self-collision.

\subsubsectionheading{Joint limits.}
Joint limits in position, velocity, and acceleration are enforced by computing the L$2$-norm joint violations, with a small margin ${\epsilon \geq 0}$ on the $d$ degrees-of-freedom
\begin{align}
    & C_{\text{limits}}(\trajectory (s)) = \sum_{\text{lim} \in \{\text{pos, vel, acc}\}} \sum_{i=0}^{d-1} C_{\text{lim}} (\trajectory_d (s))\nonumber \\
    & C_{\text{lim}}(\trajectory_d(s)) = \nonumber \\
    & \qquad \begin{cases}
        \frac{1}{2} \| \trajectory_{d,\text{min}}(s) + \epsilon - \trajectory_d (s) \|_2^2 & \text{if } \trajectory_d (s) < \trajectory_{d,\text{min}}(s) + \epsilon\\
        \frac{1}{2} \| \trajectory_{d,\textrm{max}}(s) - \epsilon - \trajectory_d (s) \|_2^2 & \text{if } \trajectory_d (s) > \trajectory_{d,\text{max}}(s) - \epsilon \\
        0 & \text{else} \\
    \end{cases} \nonumber
\end{align}
where $\trajectory_d(s)$ can be one of the partial derivatives w.r.t.~$s$, ${(\vq_d(s), \vq_d'(s), \vq_d''(s))}$.

When the constraints are in phase space, the minimum, and maximum limits are phase-dependent due to the dependency on $r(s)$~(\cref{eq:bspline_rs}).
For positions, this plays no role since ${\vq_{\text{max}}(s) = \vq_{\text{max}} \, \forall s}$.
However, for velocities we have ${\vq'_{\text{max}}(s) = \dot{\vq}_{\text{max}}} r(s)^{-1}$, with $\dot{\vq}_{\text{max}}$ the maximum specified joint velocity.
The acceleration limit follows similarly.
With a linear phase-time relation we have ${\vq'_{\text{max}}(s) = \dot{\vq}_{\text{max}}} T$ and ${\vq''_{\text{max}}(s) = \ddot{\vq}_{\text{max}} T^2}$.

\Cref{alg:mpd_algorithm_learning,alg:mpd_algorithm_planning} summarizes the procedures for learning and planning with MPD, respectively.

\section{Experimental Evaluation}
\label{sec:experimental_evaluation}

To understand the benefits of MPD, we construct experiments to answer the following questions: 
\begin{enumerate}[label=Q\arabic*)]
    \item Can MPD learn highly-multimodal collision-free trajectory distributions?
    \item How does MPD compare to other generative models?
    \item Is cost guidance during the denoising process necessary?
    \item What are the impacts of the B-spline parametrization?
    \item Can we encode trajectory priors from human demonstrations and adapt them at test time?
\end{enumerate}

\begin{algorithm}[t]
\caption{Motion Planning Diffusion -- Learning}
\label{alg:mpd_algorithm_learning}
    \DontPrintSemicolon
    \KwIn{
    Collision-free trajectories dataset $\mathcal{D}$, B-spline\\
    parameters $b$ (knots, degree, number of basis), denoising\\
    function $\vepsilon_{\vtheta}$, noise schedule terms $\alphacumprod_i$, learning rate $\gamma$
    }
    \While{not converged}{
        $\vc, \trajectory_0 \sim \mathcal{D}$  \Comment*[l]{sample a batch of contexts and trajectories}
        $\vw_0 = \text{fit-B-spline}(b, \trajectory_0)$ \Comment*[l]{fit a B-spline to the trajectory and get control points}
        \Comment*[l]{compute the denoising loss function}
        $\vepsilon \sim \Gaussian{\bm{0}, \mI}, i \sim \mathcal{U}(1, N) $, $\vw_i = \sqrt{\alphacumprod_i} \vw_0 + \sqrt{1-\alphacumprod_i} \vepsilon$ \\
        $\mathcal{L}(\vtheta) = \| \vepsilon - \vepsilon_{\vtheta}(\vw_i, i, \vc) \|_2^2 $ \\
        $\vtheta = \vtheta - \gamma \grad_{\vtheta} \mathcal{L}(\vtheta) $  \Comment*[l]{gradient optimization step}
    }
    \KwOut{optimized $\vtheta$}
\end{algorithm}

\begin{algorithm}[t]
\caption{Motion Planning Diffusion -- Planning}
\label{alg:mpd_algorithm_planning}
    \DontPrintSemicolon
    (DDPM version. For DDIM we adapt the posterior mean update with \cref{eq:ddim_posterior_guidance})\\
    \KwIn{
    Pre-trained denoising model $\vepsilon_{\vtheta}$, noise schedule terms \\
    $(\alpha_i, \alphacumprod_i, \sigma_i)$, start joint position $\vq_{\text{start}}$ and goal end-effector pose $\HEEinWorld_{\text{goal}}$, B-spline basis matrix $\mB$, motion planning costs $C_j$ and temperatures $\lambda_j$, prior temperature $\lambda_{\text{prior}}$, cost gradient start index $i_{\text{cost}}$, inner gradient parameters (steps $M$, step size $\gamma$, maximum step $\delta$)
    }
    $\vc = [\vq_{\text{start}}, \HEEinWorld_{\text{goal}}] $    \Comment*[l]{build the conditioning variable}
    $\vw_{N} \sim \Gaussian{\bm{0}, \mI} $ \Comment*[l]{sample noisy control points}
    \For{$i=N,\ldots,1$}{
        \lIf{$i > i_{\text{cost}}$}{
            $\lambda_{\text{prior} = 1}$
        }
        \Comment*[l]{compute the diffusion prior mean  \cref{eq:ddpm_posterior_mean}}
        $\vmu_i(\vw_i, i, \vc) = \frac{1}{\sqrt{\alpha_i}} \left( \vw_i - \frac{1 - \alpha_i}{\sqrt{1-\alphacumprod_i}} \lambda_{\text{prior}} \vepsilon_{\vtheta}(\vw_i, i, \vc) \right)$ \\
        \If{$i < i_{\text{cost}}$}{
            \Comment*[l]{inner gradient steps \cref{eq:guidance_inner_gradient}}
            $\vmu_i^0 = \vmu_i$  \\
            \For{$k=1,\ldots,M$}{
                \Comment*[l]{compute B-spline trajectories \cref{eq:bspline,eq:bspline_dq_ds_A,eq:bspline_derivative_time_1} }
                $\trajectory = (\vq, \vq', \vq'') = (\mB \vmu_i^k, \mB' \vmu_i^k, \mB'' \vmu_i^k)$  \\
                \Comment*[l]{compute costs gradient \cref{eq:cost_trajectory,eq:cost_trajectory_gradient_wrt_w}}
                $\vg = -\sum_{j} \lambda_j \grad_{\vmu_i^k} C_j(\trajectory) $ \\
                \Comment*[l]{Clip and apply the gradient}
                $\vmu_i^k = \vmu_i^{k-1} + \gamma \vg$ \\
                $\Delta \vmu = \text{clip}(|\vmu_i^k - \vmu_i^0|, -\delta, \delta)$ \\
                $\vmu_{i}^k = \vmu_i^0 + \Delta \vmu$
            }
        }
        \Else{
            $\vmu_i^M = \vmu_i$
        }
        \Comment*[l]{sample from the posterior distribution}
        $\vw_{i-1} = \vmu_i^M + \mSigma_i \vz, \; \vz \sim \Gaussian{\bm{0}, \mI}$  
    }
    \Comment*[l]{compute B-spline trajectories}
    $\trajectory_0 = (\vq_0, \vq_0', \vq_0'') = (\mB \vw_0, \mB' \vw_0, \mB'' \vw_0)$  \\
    \KwOut{batch of trajectories $\trajectory_0$}
\end{algorithm}

\subsection{Environments, Tasks and Datasets}
\label{sec:tasks_datasets}

\begin{figure*}[!t]
    \centering

    \captionsetup[subfloat]{labelfont=tiny,textfont=tiny,justification=raggedright,position=top}

    \subfloat[EnvSimple2D-RobotPointMass2D]{%
        \includegraphics[width=0.1445\textwidth]{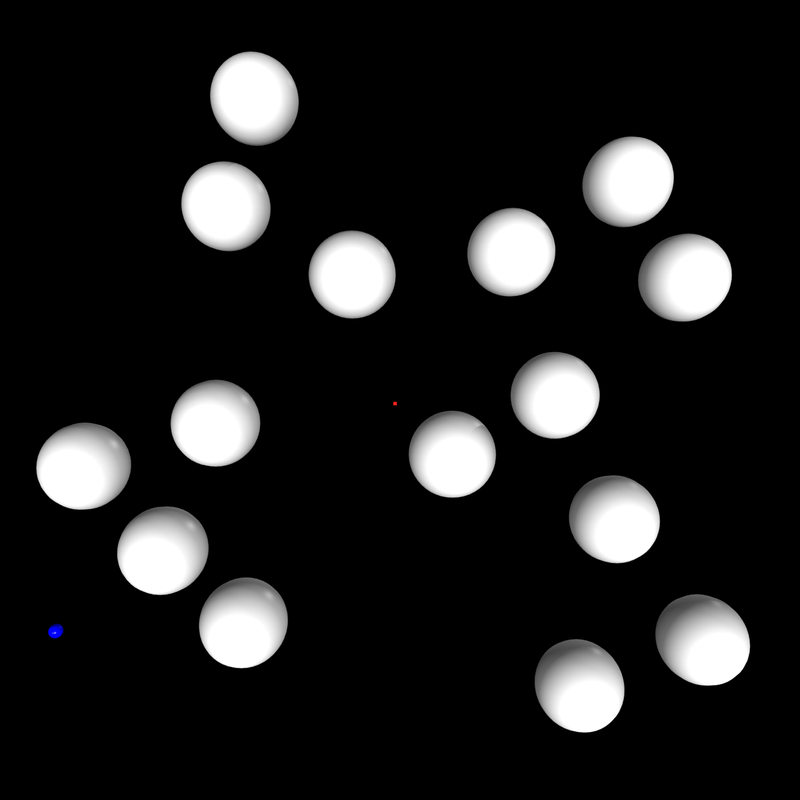}
        \label{fig:EnvSimple2D-RobotPointMass2D}
    }
    \hfill
    \subfloat[EnvNarrowPassageDense2D-RobotPointMass2D]{%
        \includegraphics[width=0.1445\textwidth]{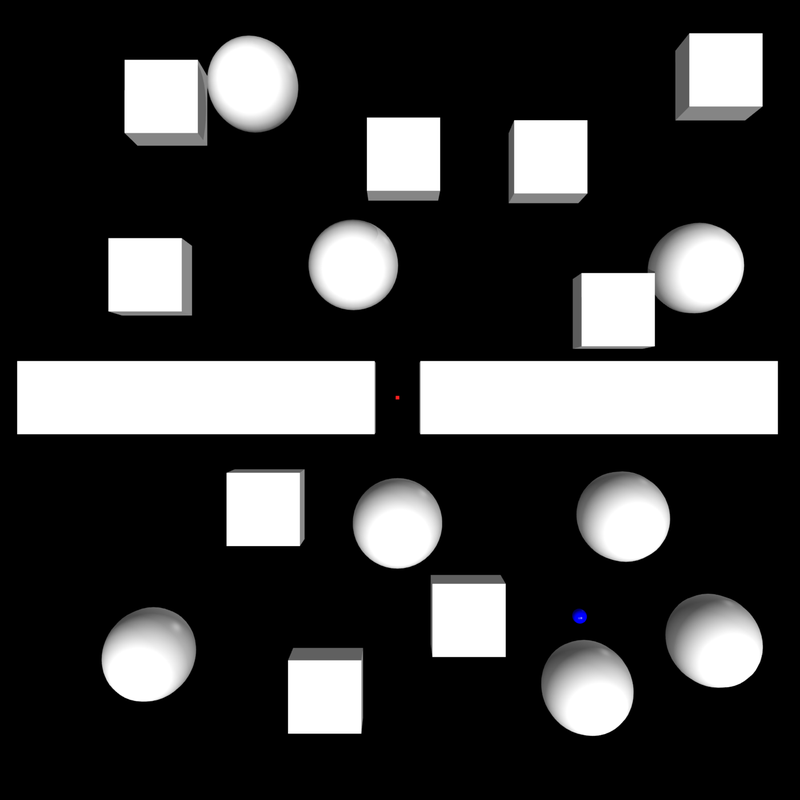}
        \label{fig:EnvNarrowPassageDense2D-RobotPointMass2D}
    }
    \hfill
    \subfloat[EnvPlanar2Link-RobotPlanar2Link]{%
        \includegraphics[width=0.1445\textwidth]{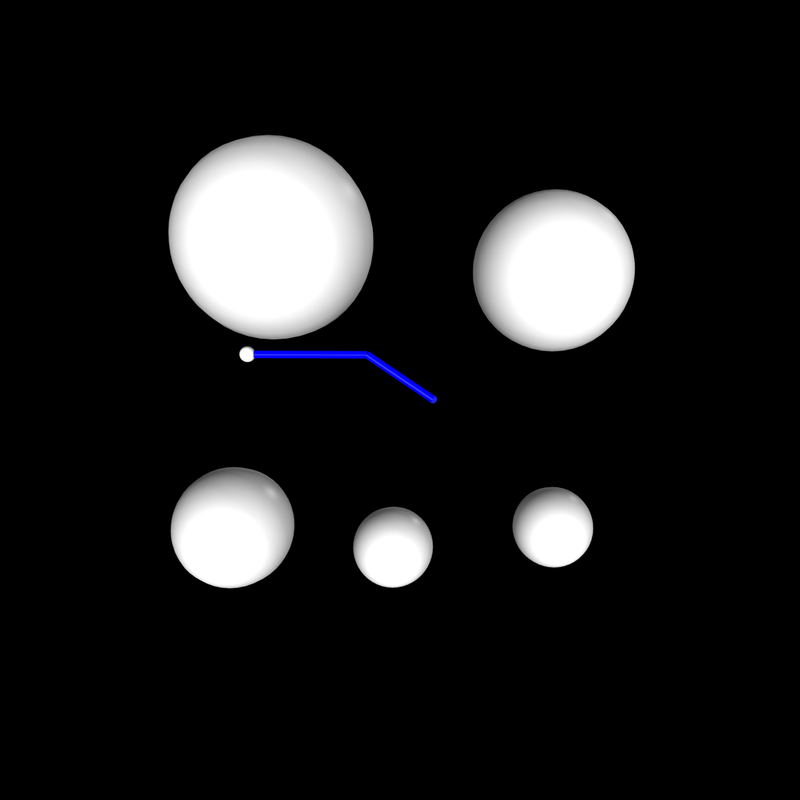}
        \label{fig:EnvPlanar2Link-RobotPlanar2Link}
    }
    \hfill
    \subfloat[EnvPlanar4Link-RobotPlanar4Link]{%
        \includegraphics[width=0.1445\textwidth]{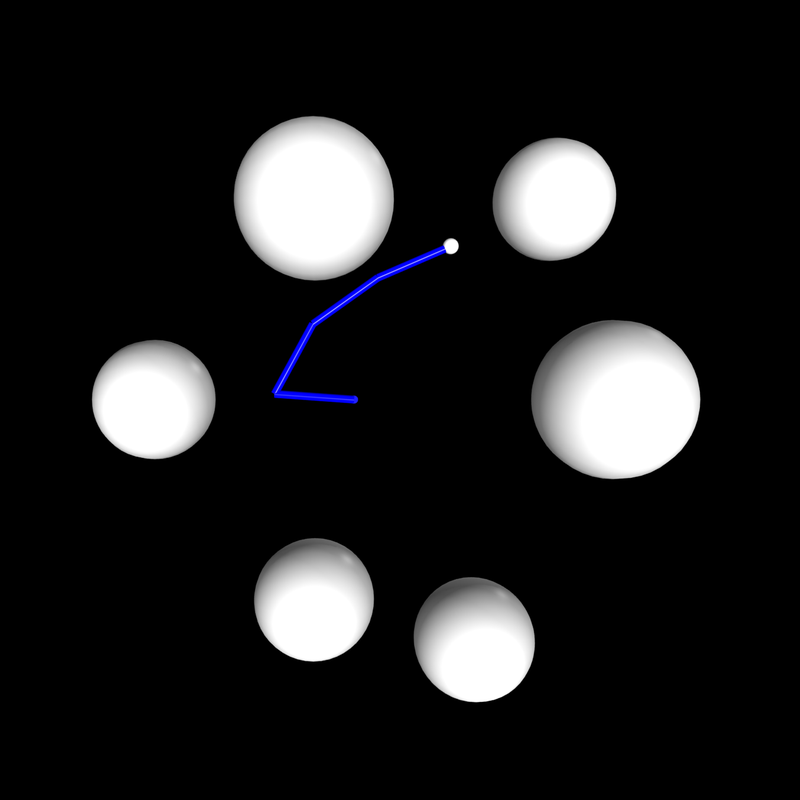}
        \label{fig:EnvPlanar4Link-RobotPlanar4Link}
    }
    \hfill
    \subfloat[EnvSpheres3D-RobotPanda]{%
        \includegraphics[width=0.1445\textwidth]{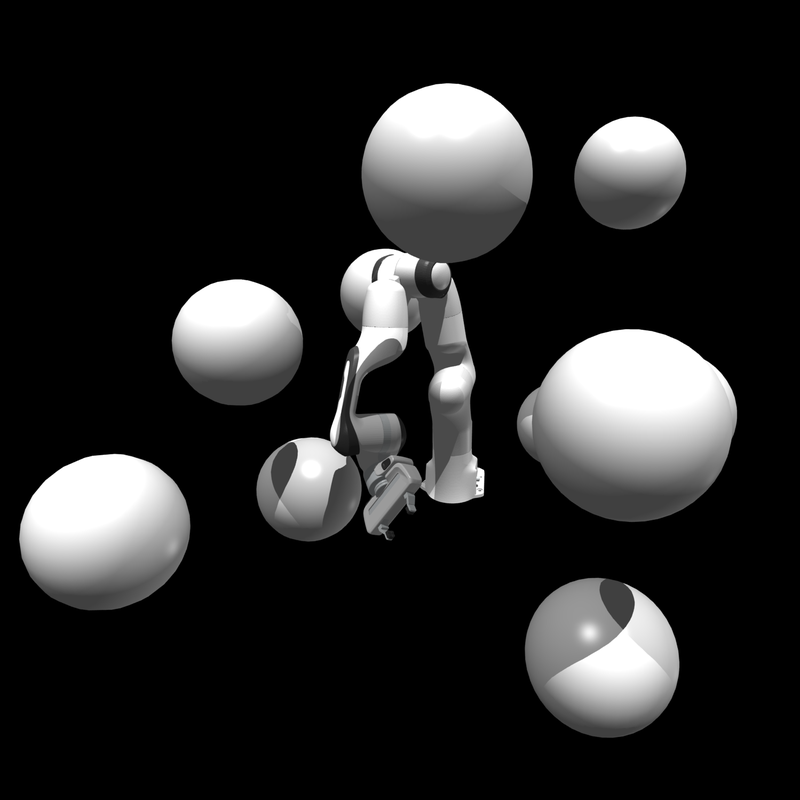}
        \label{fig:EnvSpheres3D-RobotPanda}
    }
    \hfill
    \subfloat[EnvWarehouse-RobotPanda]{%
        \includegraphics[width=0.1445\textwidth]{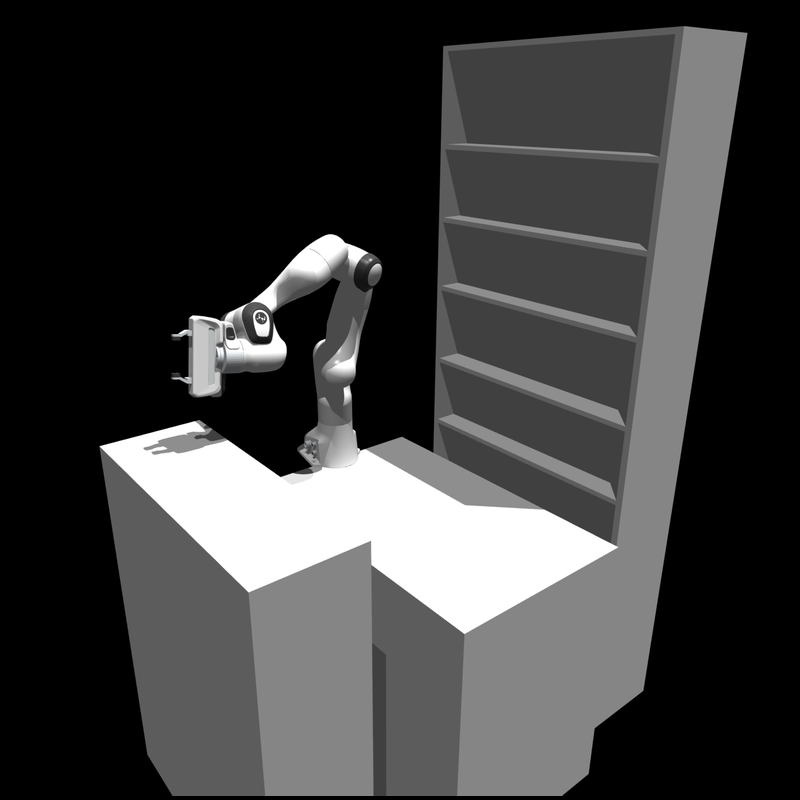}
        \label{fig:EnvWarehouse-RobotPanda}
    }
    \\
    \vspace{-0.25cm}

    \captionsetup[subfigure]{labelformat=empty}

    \subfloat[]{%
        \includegraphics[width=0.1445\textwidth]{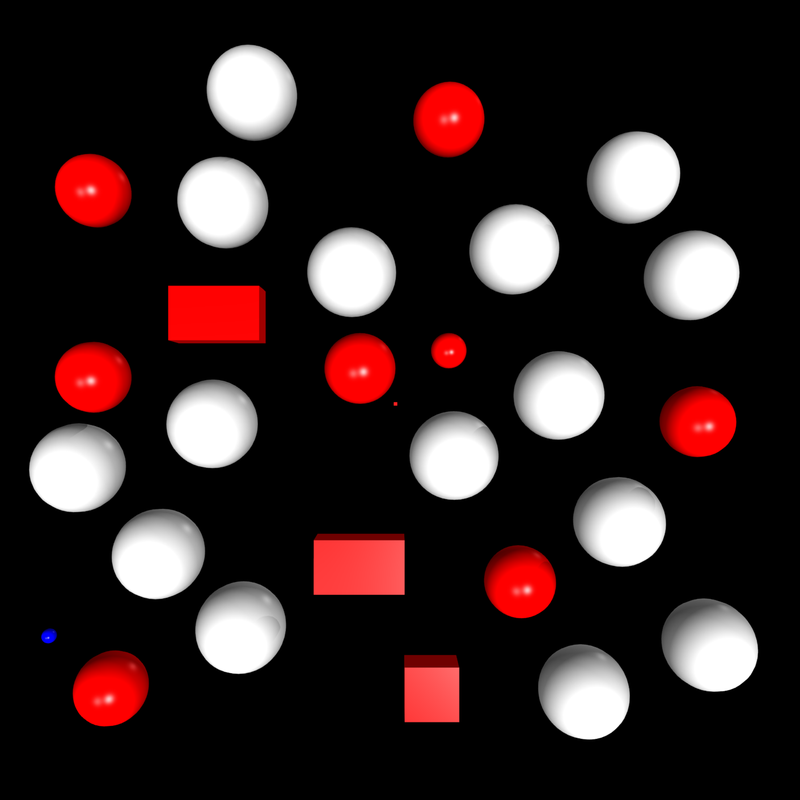}
        \label{fig:EnvSimple2D-RobotPointMass2D-extra}
    }
    \hfill
    \subfloat[]{%
        \includegraphics[width=0.1445\textwidth]{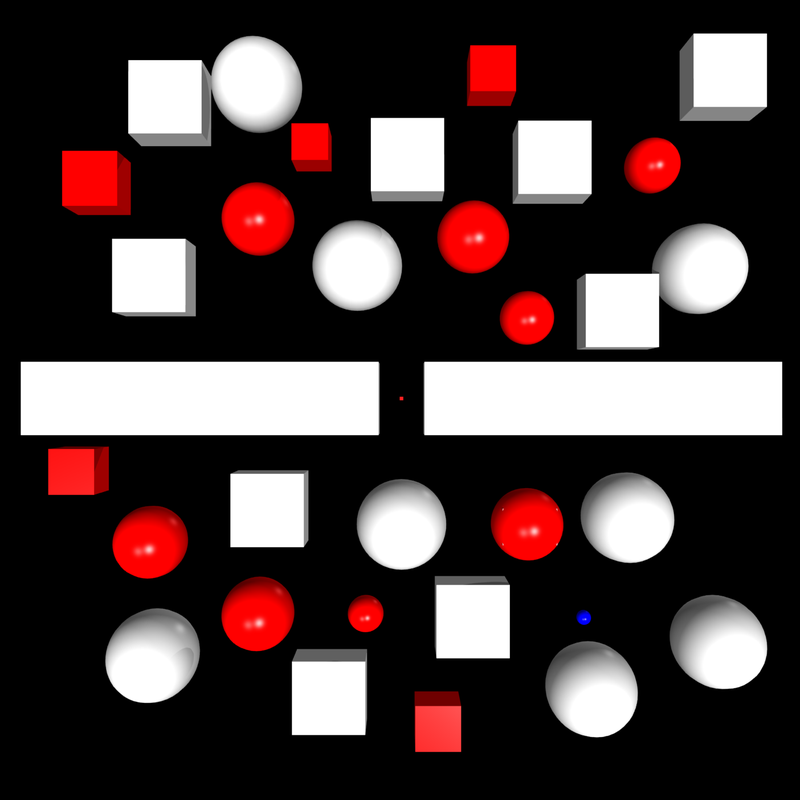}
        \label{fig:EnvNarrowPassageDense2D-RobotPointMass2D-extra}
    }
    \hfill
    \subfloat[]{%
        \includegraphics[width=0.1445\textwidth]{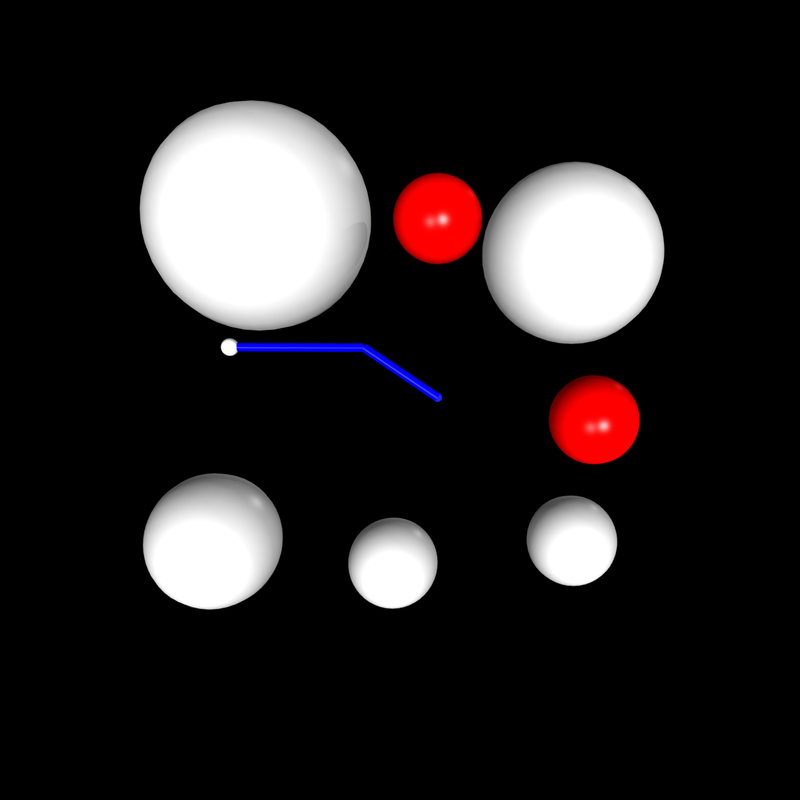}
        \label{fig:EnvPlanar2Link-RobotPlanar2Link-extra}
    }
    \hfill
    \subfloat[]{%
        \includegraphics[width=0.1445\textwidth]{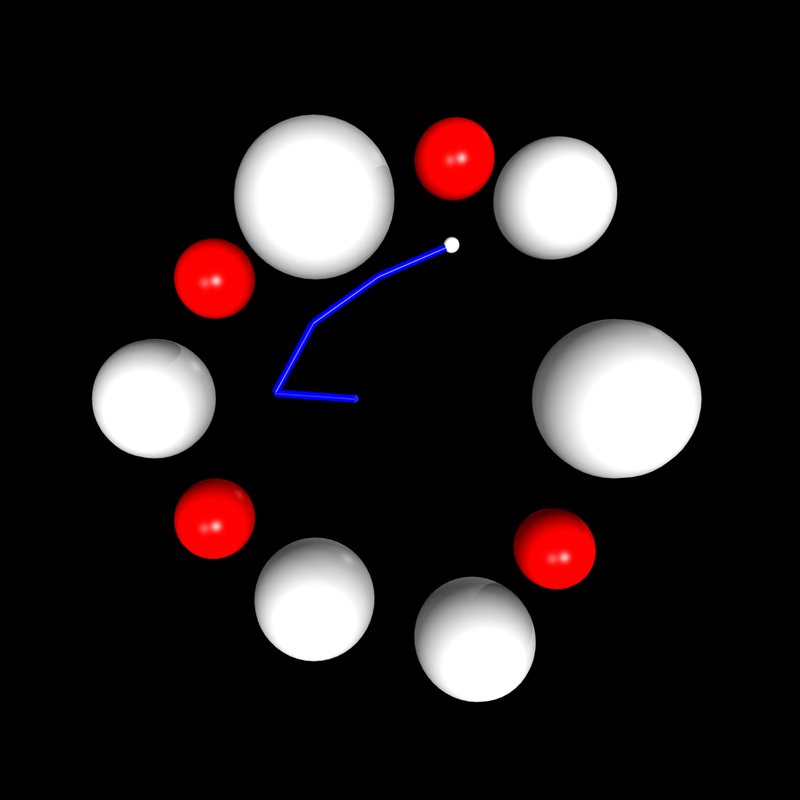}
        \label{fig:EnvPlanar4Link-RobotPlanar4Link-extra}
    }
    \hfill
    \subfloat[]{%
        \includegraphics[width=0.1445\textwidth]{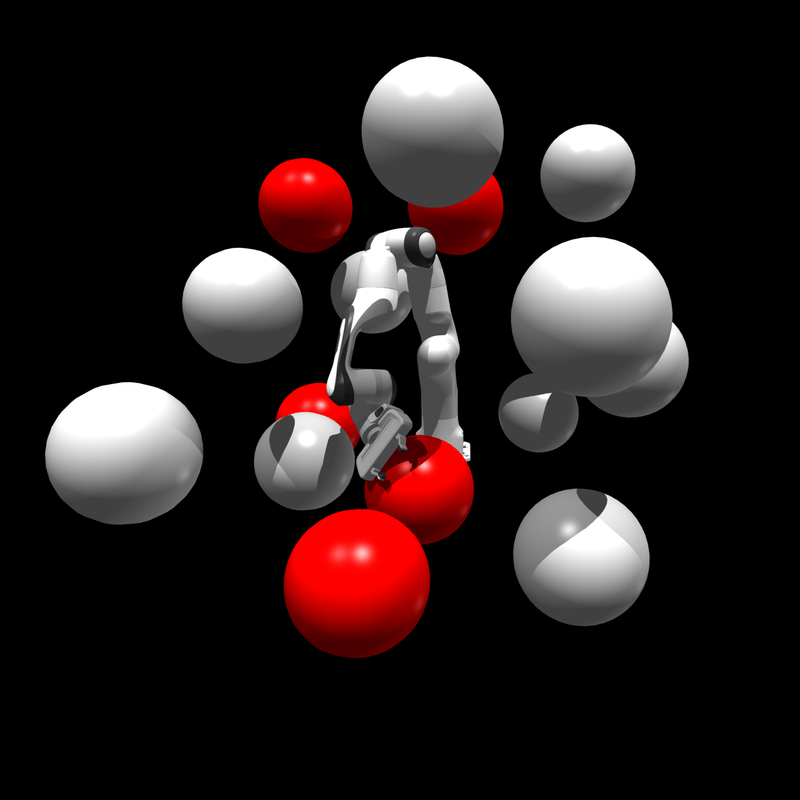}
        \label{fig:EnvSpheres3D-RobotPanda-extra}
    }
    \hfill
    \subfloat[]{%
        \includegraphics[width=0.1445\textwidth]{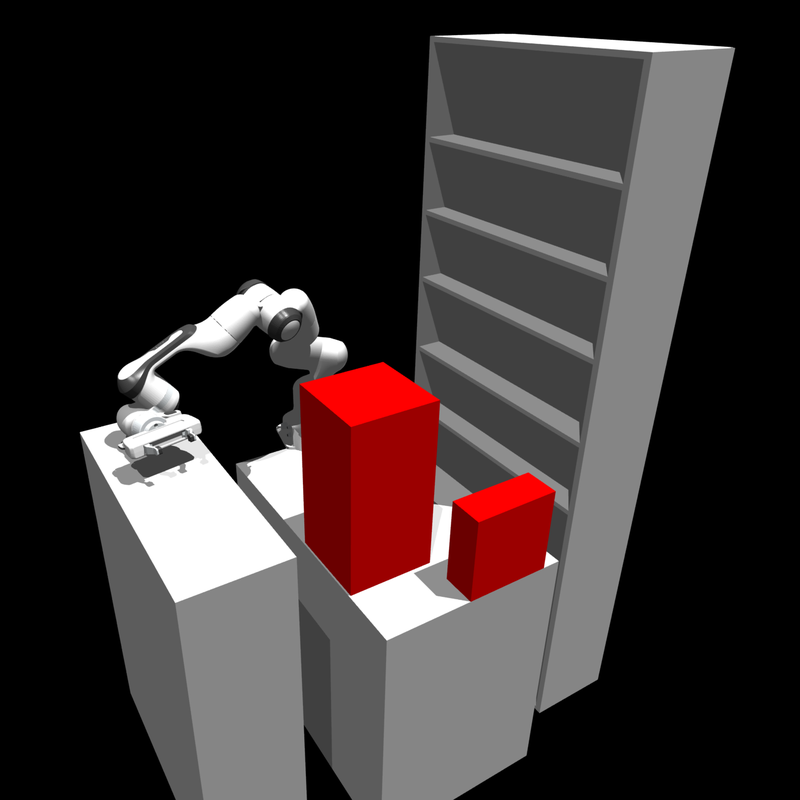}
        \label{fig:EnvWarehouse-RobotPanda-extra}
    }

    \caption[Environments and robots used for the motion planning experiments.]{
        The environments used for the motion planning experiments, with different robot models and increasing complexity.
        In $2$D tasks the robots are depicted in blue (as a dot for RobotPointMass2D).
        The bottom row shows the tasks with additional obstacles (in red), which are not present during training.
    }
    \label{fig:tasks}
    \vspace{-0.5cm}  
\end{figure*}

We consider a set of tasks that are representative of the challenges faced in robot motion planning, from simple to complex, including
a $2$D point mass navigating an environment with simple obstacles,
a $2$D point mass navigating a narrow passage,
a $2$-link planar robot,
a $4$-link planar robot,
and a $7$-dof robot arm manipulator in two settings:
one illustrative environment with collision spheres;
and a warehouse environment with a table and two shelves.
\reviewersix{
The latter aims to motivate the applicability of our method in a real-world task, such as shelf rearrangements in distribution facilities, where shelves are static, but at deployment new objects can be placed on the table or shelves.
}
The tasks are displayed in \cref{fig:tasks}, with naming convention [Environment]-[Robot].
The goal is to move the robot (without collisions) from start to goal configurations -- a desired joint position or an end-effector pose.

For each task, we generate a dataset of collision-free paths with the OMPL library Python bindings~\cite{DBLP:journals/ram/SucanMK12}, along with PyBullet~\cite{DBLP:conf/siggraph/Coumans15} to setup the environment and perform collision checking.
We randomly sample multiple contexts and use RRT~Connect~\cite{kuffner2000rrtconnect} to generate paths, which are afterward shortcutted and smoothed.
Moreover, we generate an additional path by reverse-connecting the goal to the start configuration.
Alternatively, if we know the task requires goal poses to be located in some regions of the workspace, we can specify the goal pose and use inverse kinematics to obtain the goal joint configuration.
For example, in the EnvWarehouse-RobotPanda task, the goal pose is located on top of the table or on the shelves, and we sample paths from any initial configuration to those goal poses.
There are two reasons for using RRT Connect to generate data.
First, because it is faster than optimal planners like RRT$^*$, and since we optimize these paths further using defined cost functions, we trade optimality for speed.
Second, it is probabilistically complete, and its vertices cover all the configuration-free space (Corollary $1$ of~\cite{kuffner2000rrtconnect}), thus finding multiple modes for reaching the goal from the start configuration.
The number of paths, which equals the number of contexts, generated per environment can be found in \cref{tab:hyperparameters_training}.
Since we are planning in joint space, the number of points needed to cover the whole space increases exponentially with the number of joints.
Thus, a sufficient amount of data is needed to counter the curse of dimensionality.
E.g., for the discretization of $100$ points per joint, a $4$-dof robot arm requires $100^4 = 100$M points to cover the discretized space.
When training the diffusion model prior, we sample a context and path from the dataset and fit a B-spline to the path using the \texttt{splprep} function from the SciPy library~\cite{virtanen2020scipy}.
The number of B-spline control points is a hyperparameter and is chosen so the path almost does not collide with the environment.
Note that we do not need to be 100\% collision-free, as collision-avoidance costs are used during optimization.

Choosing the number of control points for the B-spline and the denoising network size is a trade-off between model expressivity and computational cost.
The hyperparameters for training are detailed in \cref{sec:hyperparameters_training}.
The loss function in \cref{eq:diffusion_loss} is optimized using mini-batch gradient descent with the Adam optimizer~\cite{DBLP:journals/corr/KingmaB14} and a learning rate ${3 \times 10^{-4}}$.

Generalization is tested on contexts not seen during training and in the presence of new obstacles (see \cref{fig:tasks}), which assess the method's ability to exploit the prior knowledge to generate collision-free paths in unseen scenarios.
The hyperparameters for inference are detailed in \cref{sec:hyperparameters_inference}.

\subsection{Baselines}
\label{sec:baselines:mpd}

We compare MPD against the following baselines:
\begin{itemize}
    \item To compare the drawbacks of using an uninformed prior, we use a Gaussian Process (GP) prior that connects the start and goal configurations and optimizes the cost function using gradient descent.
        We name this baseline GPprior+Cost, and it is essentially CHOMP~\cite{ratliff2009chomp}.
        When the context includes the goal end-effector pose, we use a goal joint position computed with inverse kinematics.
    \item To assess the benefits of using a diffusion model, we use a baseline that learns a Conditional Variational Autoencoder (CVAE)~\cite{DBLP:conf/nips/SohnLY15} to model the trajectory distribution.
        It shares the same U-Net architecture as the denoising function $\vepsilon_{\vtheta}$, but at the lowest level of the U-Net the embedding is projected into a $32$-dimensional latent space that encodes the mean and standard deviation of a Gaussian distribution.
        The optimized loss function is a sum of L$2$-norm of the predicted and ground-truth trajectory and the KL-divergence between the latent space posterior and a standard Gaussian distribution weighted by $\beta=0.1$.
    \item We use a baseline named [Prior]+Cost to compare MPD against sampling first from the prior and then optimizing the cost function.
    The [Prior] is the diffusion (Dprior) or the CVAE model.
\end{itemize}
All baselines use the B-spline parametrization.
For a similar comparison, we use the same number of cost function optimization steps in all baselines.
To accelerate sampling from the diffusion model, we use DDIM with $15$ steps and a quadratic step schedule.
All methods are implemented with PyTorch, and the experiments were conducted on a machine with an AMD EPYC 7453 28-Core Processor and NVIDIA RTX 3090.

\subsection{Metrics}
\label{sec:metrics:mpd}

We report the following average metrics.
\textit{Success rate} is $1$ if at least one of the trajectories in the batch is not in collision and inside joint limits, and 0 otherwise.
\textit{Fraction of valid trajectories} is the percentage of generated trajectories that are collision-free and inside the joint limits.
\textit{Diversity} is measured by the Vendi score~\cite{DBLP:journals/tmlr/FriedmanD23} with the similarity kernel
${k(a, b) = \exp(-\|a - b\|_2^2)}$~\cite{li2024diffusolvediffusionbasedsolvernonconvex}, for the dense interpolation of two trajectories $a$ and $b$.
A higher Vendi score means more diverse trajectories.
\textit{Error pos.} and \textit{Error ori.}
are the position and orientation errors, respectively, between the end-effector goal pose and the one at the end of the generated trajectory.
The last four metrics are reported only for valid trajectories.

\subsection{General Results in Simulation}
\label{sec:general_results_simulation}

\begin{figure*}[t]
    \centering

    \includegraphics[width=0.99\textwidth]{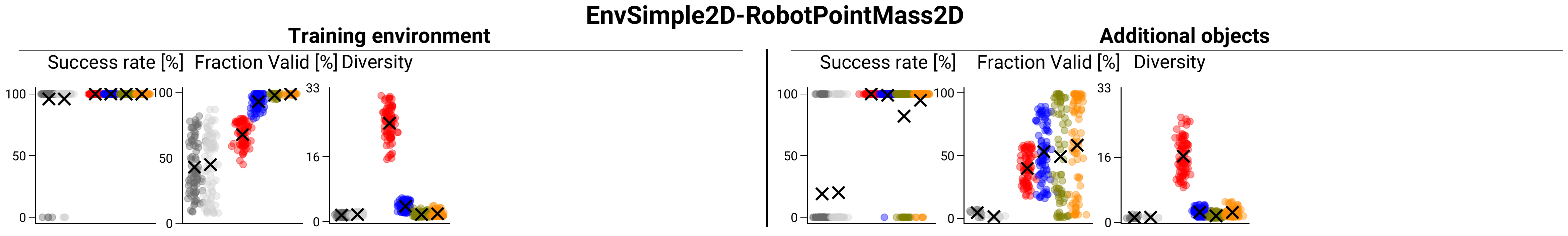}
    \includegraphics[width=0.99\textwidth]{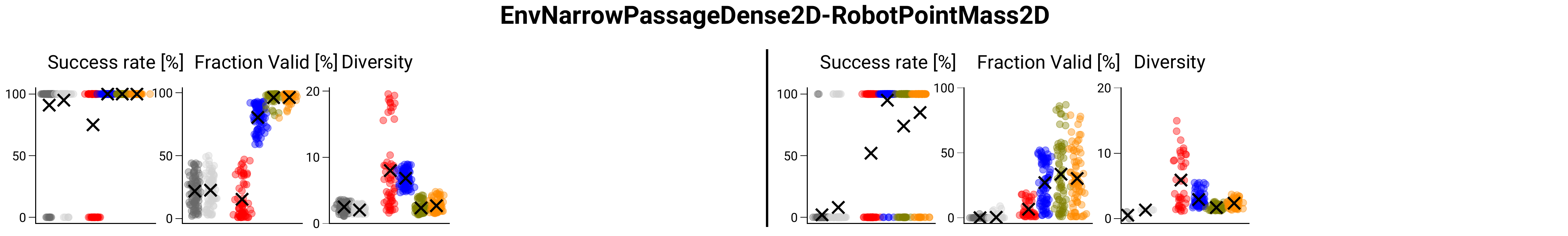}
    \includegraphics[width=0.99\textwidth]{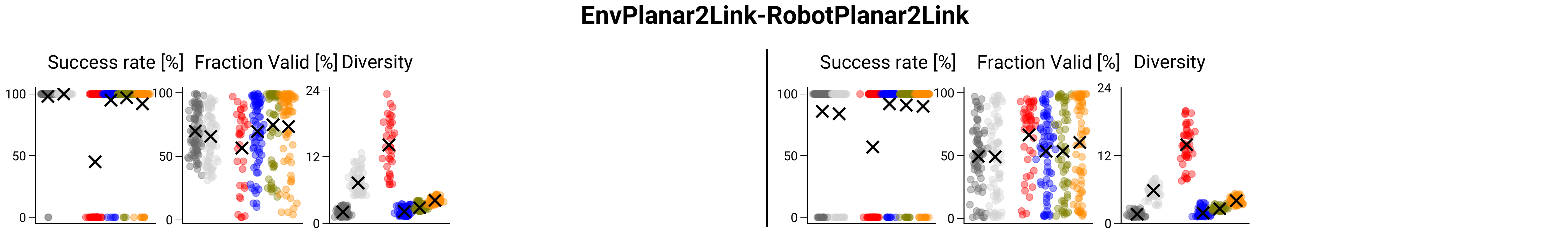}
    \includegraphics[width=0.99\textwidth]{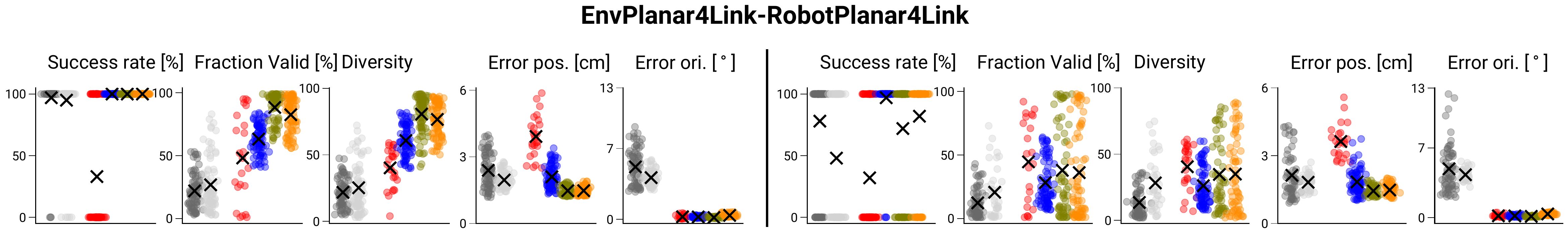}
    \includegraphics[width=0.99\textwidth]{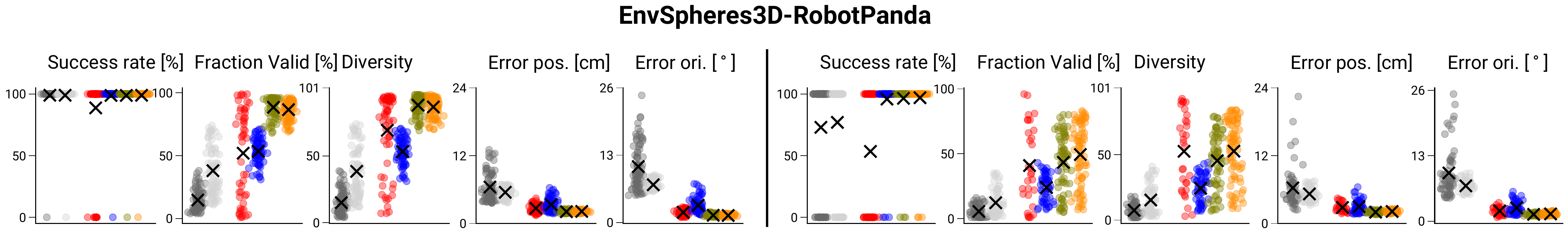}
    \includegraphics[width=0.99\textwidth]{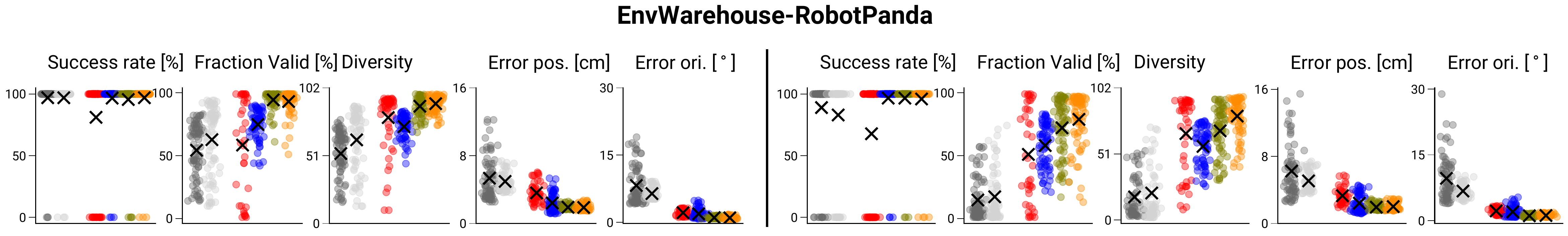}
    \\
    \includegraphics[width=0.65\textwidth]{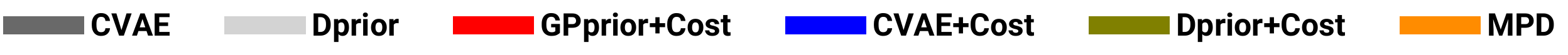}

    \caption[Results for motion planning algorithms.]{
        Performance metrics for different algorithms on the tasks from \cref{fig:tasks}.
        The results report the swarm plot of sampling $100$ contexts from the test set and optimizing $100$ trajectories per context.
        The cross represents the mean value.
        The columns under \textit{training environment} show the results without additional objects.
        In $2D$ environments, the errors in end-effector position and orientation are always $0$ since a joint goal is provided.
    }
    \label{fig:results_simulation_swarmplots}
    \vspace{-0.5cm}  
\end{figure*}

In this experiment, we report the results using all motion planning costs.
Here, we want to evaluate how the methods perform in terms of collisions and reaching the desired end-effector goal poses.
Therefore, we use a large time duration of $10$ seconds to prevent optimization over velocities and accelerations.
We sample $100$ contexts
and optimize $100$ trajectories per context.
The results are displayed in
\cref{fig:results_simulation_swarmplots}.

To answer Q1, we observe the results under \textit{training environment} since they show how well the generative priors model the data distribution.
They show that the diffusion-based models (Dprior, Dprior+Cost, MPD) obtain success rates that match or surpass the baselines while keeping a high diversity in generated trajectories.
When comparing CVAE and Dprior, their success rate and validity fraction are close, but the diversity score is higher for the diffusion-based models, in particular for higher-dimensional environments (using RobotPanda).
For instance, in the EnvWarehouse-RobotPanda, both CVAE and Dprior achieve $97\%$ \reviewerten{mean} success rate, but Dprior shows more \reviewerten{mean} variability ($50.9$ vs. $60.3$).
Moreover, Dprior achieves smaller \reviewerten{mean} errors in the desired end-effector goal position and orientation ($7.4$cm vs. $5.2$cm, $10.5^\circ$ vs. $6.7^\circ$), which due to model approximations are not $0$.
These errors are improved during cost optimization.
\reviewerten{
Both diffusion-based models, Dprior+Cost and MPD, achieve similar \reviewerten{mean} position and orientation errors, e.g., in this task, $2.00$cm and $1.1^\circ$.
}

To answer Q2, we look at the results when using newer obstacles in the scene under \textit{additional objects} columns.
The success rate and validity fraction drop for all methods, in comparison to the training environment, which is expected, in particular for the prior methods, CVAE and Dprior, which do not have knowledge of the new obstacles.
For instance, in the EnvNarrowPassageDense2D-RobotPointMass2D task, Dprior's \reviewerten{mean} success rate dropped from $95\%$ to $8\%$.
This is a task where an informed prior is particularly useful since traversing from the top to the bottom of the environment needs to be done through a very small passage.
We focus now on the higher-dimensional task EnvWarehouse-RobotPanda.
In the presence of new obstacles, all methods that perform cost optimization improve the success rate and validity fraction in comparison to using only the prior.
For example, MPD increases the \reviewerten{mean} validity fraction of Dprior from $18.4$ to $73.5\%$, which is the largest among all methods for this task.
Moreover, MPD also shows more \reviewerten{mean} diversity in the valid trajectories in comparison to baselines that first sample from the prior and then optimize the cost ($66.0$ for Dprior+Cost vs. $74.8$ for MPD).
Note that these results are more pronounced for the \textit{additional objects} tasks, and a discussion for why this happens is found in \cref{sec:mpd_vs_diffusion_prior_then_guide}.
When comparing MPD with an uninformed prior GPprior+Cost, we observe a \reviewerten{mean} success rate increase from $67.8\%$ to $96.0\%$, which supports the claim that the diffusion prior is useful to guide the trajectories through collision-free regions.
In terms of diversity, we expect the GPprior to provide the largest value since this is an uninformed prior.
MPD's diversity is the closest to the GPprior one or slightly larger for the EnvWarehouse-RobotPanda task.
These results show that MPD produces a higher percentage of collision-free trajectories (and inside joint limits) while being diverse.
It is also important to note the effect of optimizing the task cost.
Methods that optimize this cost produce trajectories resulting in end-effector pose errors much smaller than the ones obtained just sampling from the priors.
E.g., in the EnvWarehouse-RobotPanda task, the end-effector \reviewerten{mean} error in orientation drops from $7.1$ to $1.2$ between Dprior and MPD.

\begin{figure}[!t]
    \centering
    \includegraphics[width=0.99\columnwidth]{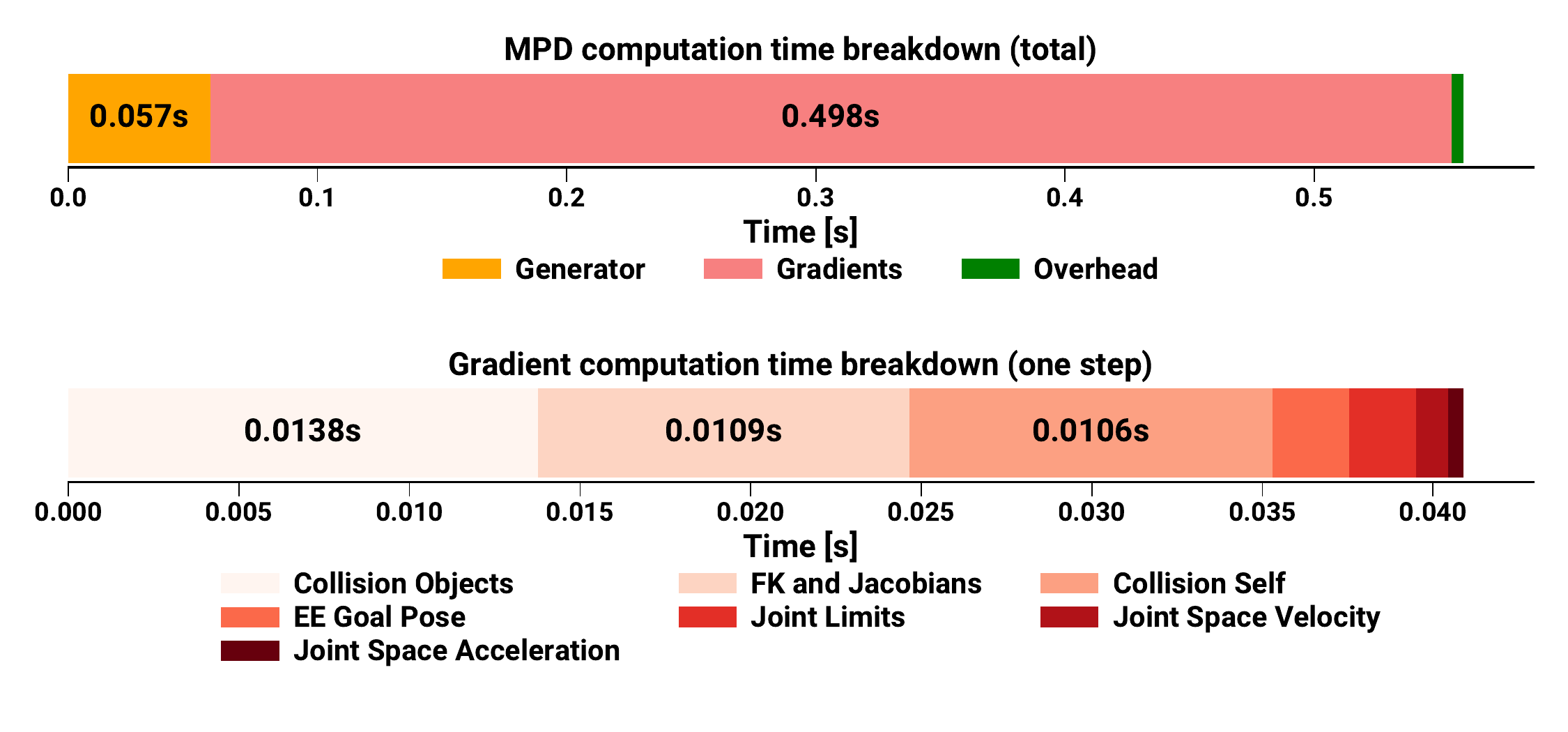}
    \vspace{-1.0cm}
    \caption[Computation time breakdown for sampling with MPD.]{
        Computation times breakdown for sampling $100$ trajectories with MPD in the EnvSpheres3D-RobotPanda environment.
        (Top)
        MPD total inference time and breakdown per generator and gradient computations.
        (Bottom)
        Breakdown for the computation of one gradient step.
    }
  \label{fig:mpd_timing}
\end{figure}

\begin{figure}[!t]
    \centering
    \includegraphics[width=0.49\columnwidth]{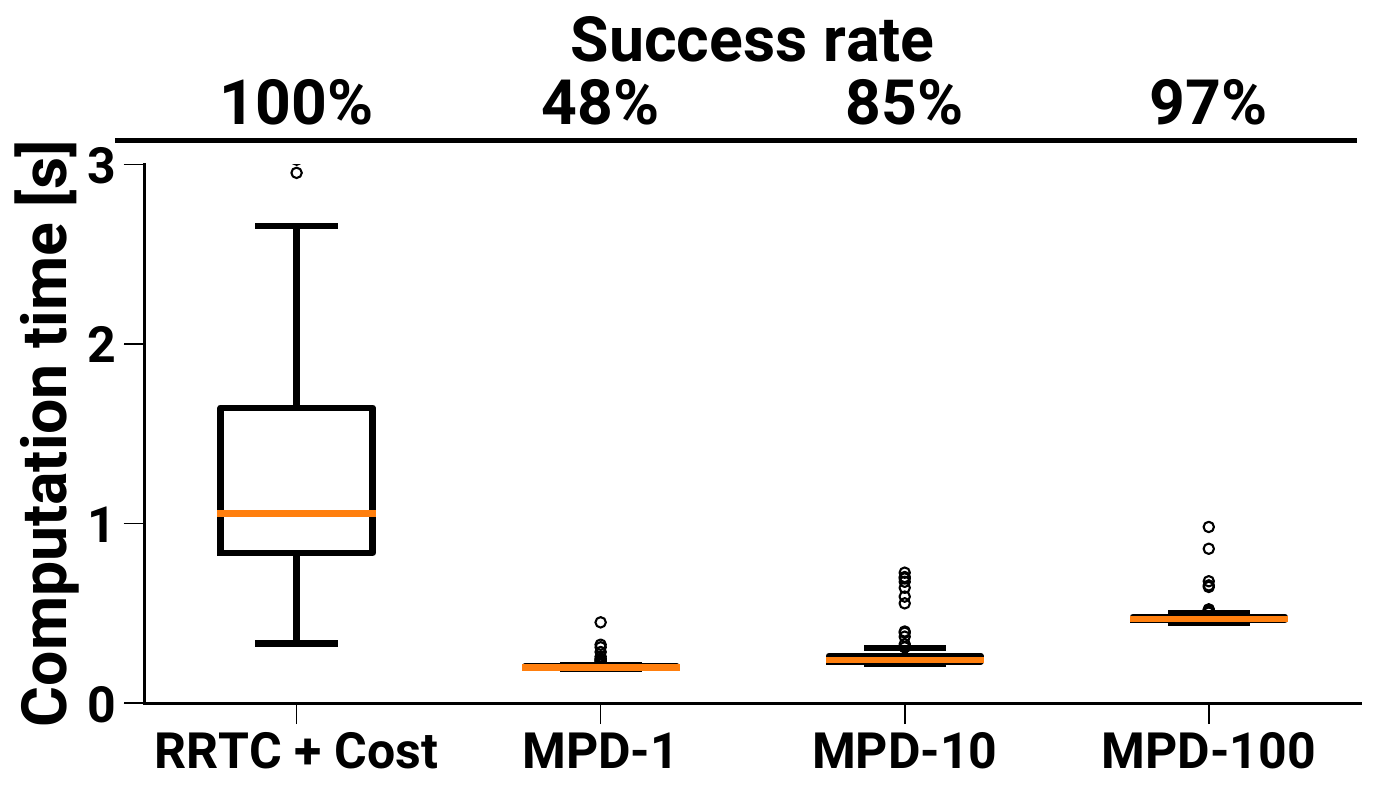}
    \includegraphics[width=0.49\columnwidth]{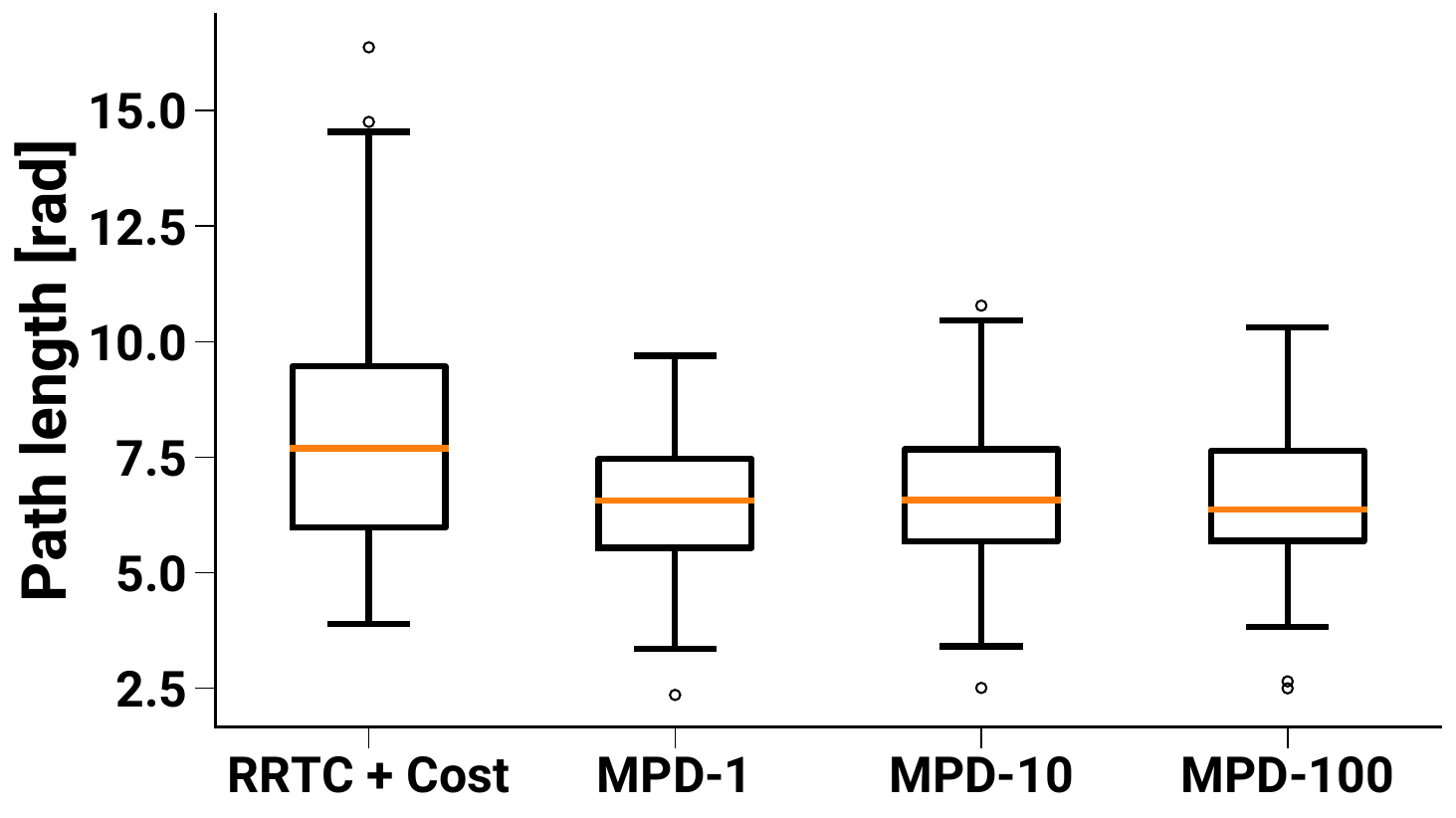}
    \vspace{-0.3cm}
    \caption[RRTConnect+Cost vs. MPD]{
        \reviewerten{
        (Left) computation time and success rate, and
         (right) shortest path lengths results
         in the EnvSpheres3D-RobotPanda task (additional objects) across $100$ contexts.
         We compare one RRT Connect sample with cost optimization (RRTC + Cost), and MPD with increasing batches ($1$, $10$, $100$).
        }
    }
  \label{fig:rrtconnect_then_guide_vs_mpd}
  \vspace{-0.6cm}
\end{figure}

The optimization time is highly implementation-dependent (and on CPU/GPU hardware versions).
We parallelize all cost computations using vectorized operations in PyTorch, but computing the costs is still done sequentially in Python.
Typically, a disadvantage of diffusion models is their slow sampling speed, but by using DDIM we reduce the number of sampling steps, thus achieving faster inference without a significant performance drop.
In \cref{fig:mpd_timing}, we show the computation time breakdown for MPD in the most complex environment (EnvSpheres3D-RobotPanda), using MPD with $15$ DDIM steps and $4$ intermediate gradient steps, which in total takes approximately $0.56$s.
The diffusion sampling alone takes $0.057$s ($\approx 3.8$ms per denoising network pass).
The costly part is the gradient computation, which is similar for all methods.
Therefore, the only difference between the baselines, is diffusion taking $0.057$s more.

\reviewerten{
\Cref{fig:motivation_planar_2_link} presented a motivation for initializations in optimization-based planning methods.
In our previous work~\cite{DBLP:conf/iros/Carvalho0BK023}, the experiments showed that (as expected) a sampling-based prior achieves $100\%$ success rates, but with more computation time (hence, we do not include it here in the main baselines).
In this work, we consider sampling \textit{one} trajectory using RRTConnect, since parallel sampling incurs too much computation time as it is not as easy to parallelize.
We use the same implementation as described in \cref{sec:tasks_datasets}.
The question we aim to answer is whether MPD has a benefit over first running a sampling-based planner and then solving the optimization problem from \cref{eq:trajectory_optimization} (we used the same number of steps as for MPD).
For this experiment, we chose the EnvSpheres3D-RobotPanda task with additional objects because it is the hardest task we considered (due to narrow passages).
\Cref{fig:rrtconnect_then_guide_vs_mpd} shows the results obtained when planning $100$ contexts.
We observe that sampling \textit{one} trajectory with RRTConnect uses more computation time than sampling $100$ with a learned diffusion model.
MPD's success rate increases with the batch size, while computation time remains almost constant.
The boxplot on the right side also shows that MPD produces trajectories with shorter lengths and less variance.
This experiment shows that offloading sampling-based planner computations and compressing them into a trajectory 
diffusion model leads to faster and better planning results.
}

\begin{figure}[t]
    \centering

    \includegraphics[height=0.115\textheight]{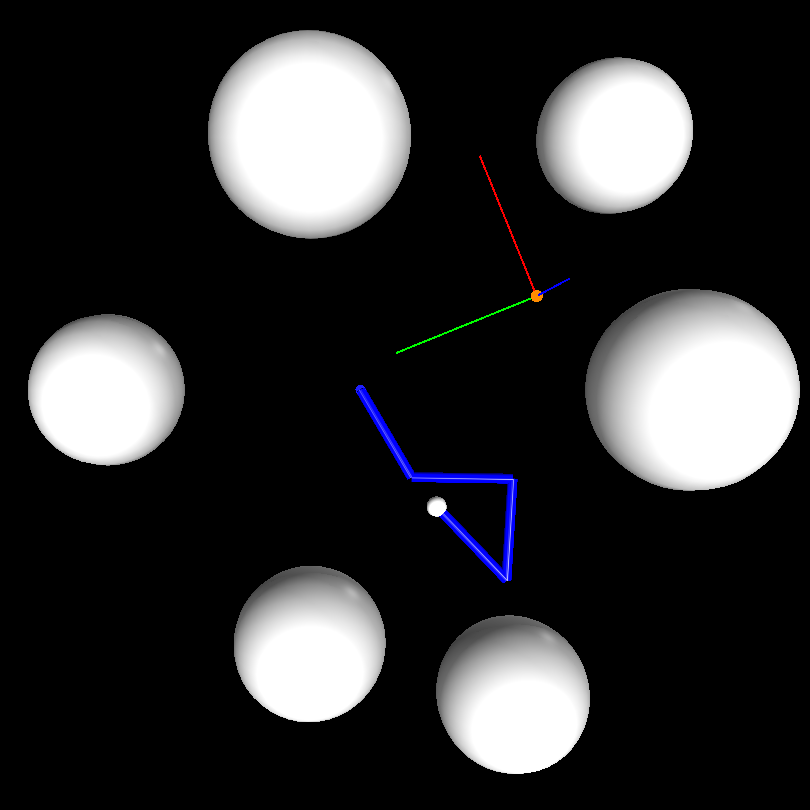}
    \includegraphics[height=0.115\textheight]{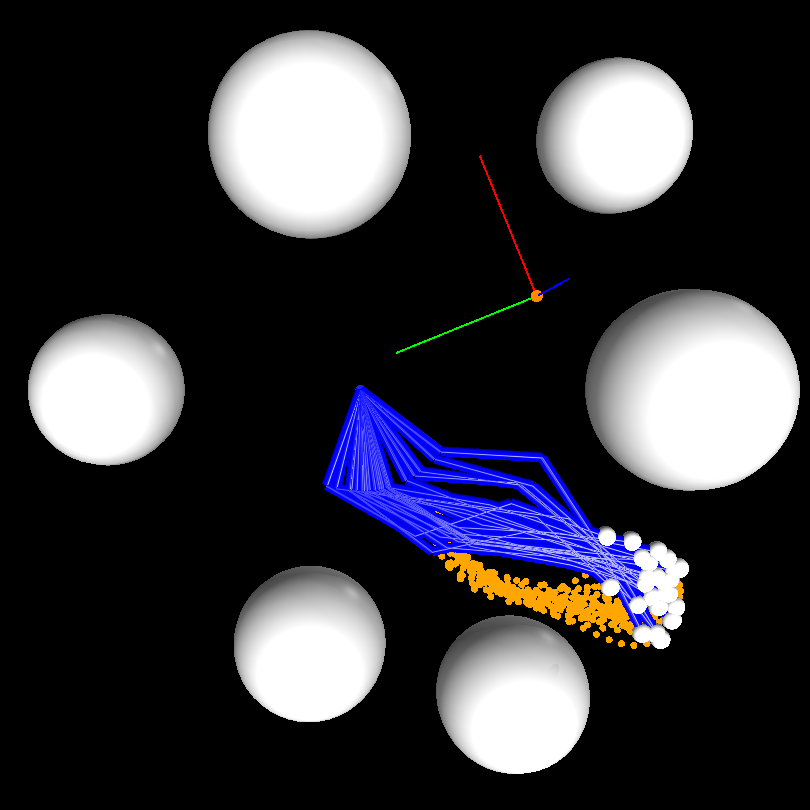}
    \includegraphics[height=0.115\textheight]{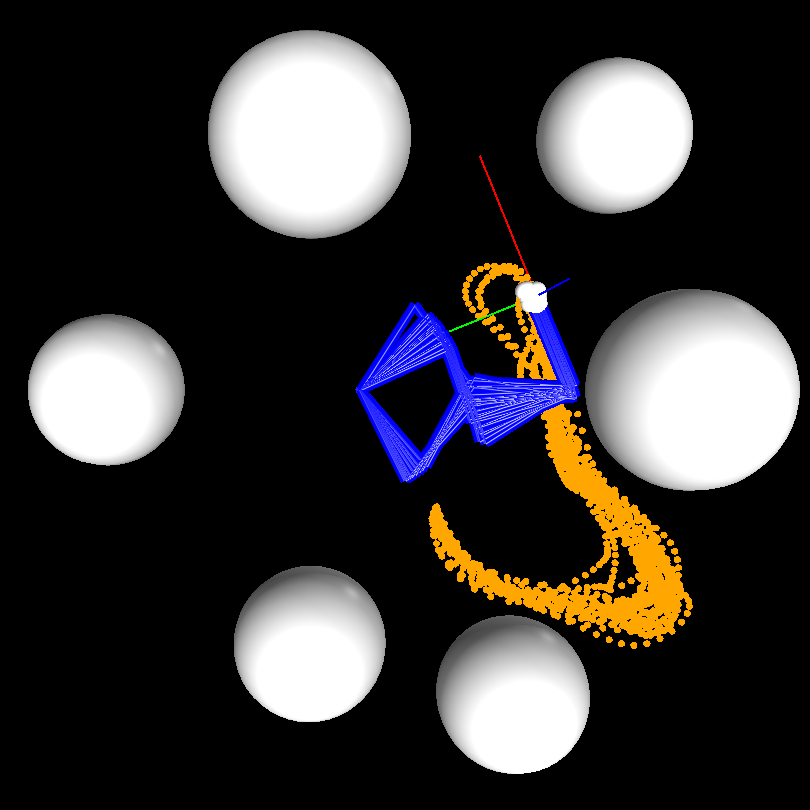}
    \\
    \includegraphics[width=0.99\columnwidth]{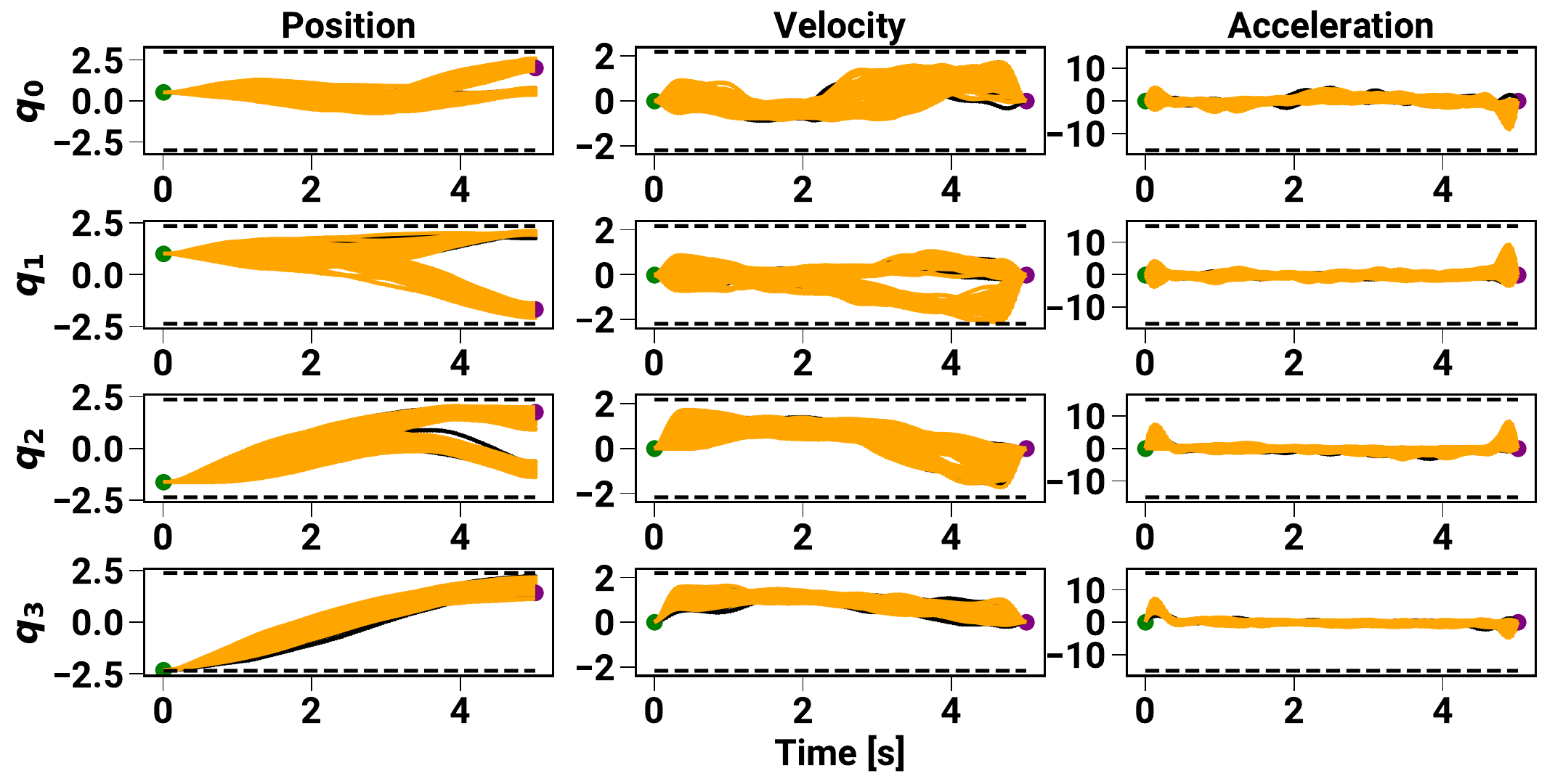}

    \vspace{-0.3cm}
    \caption[Trajectories generated with MPD in the EnvPlanar4Link-RobotPlanar4Link task.]{
        (Top)
        Task-space trajectories generated by MPD in the EnvPlanar4Link-RobotPlanar4Link task.
        The reference frame shown is the desired end-effector pose.
        The orange line shows the end-effector trajectories from the start (left) to the goal (right).
        Notice how the model generates multimodal goal joint positions for the same desired end-effector goal pose.
        (Bottom)
        Joint trajectories in time, where orange are collision-free and within joint limits (depicted by the dashed lines), and others are in black.
    }
    \label{fig:planar4link_results}
    \vspace{-0.6cm}
\end{figure}

\Cref{fig:planar4link_results} shows an example of a planning task in the EnvPlanar4Link-RobotPlanar4Link environment, using a trajectory duration of $5$ seconds.
Given a goal end-effector pose, the model generates multimodal trajectories in joint space, which are collision-free and within joint limits.
The rightmost figure on the top row shows that final joint configurations achieve the same end-effector pose.
We note that the diffusion model was trained using one trajectory per context, but when tested in a new context not seen during training, the model generates multimodal trajectories, which we attribute to the capacity of the denoising network to generalize to new contexts and model multimodal distributions quite well.

\begin{figure}[!t]
  \centering
  \includegraphics[width=0.8\columnwidth]{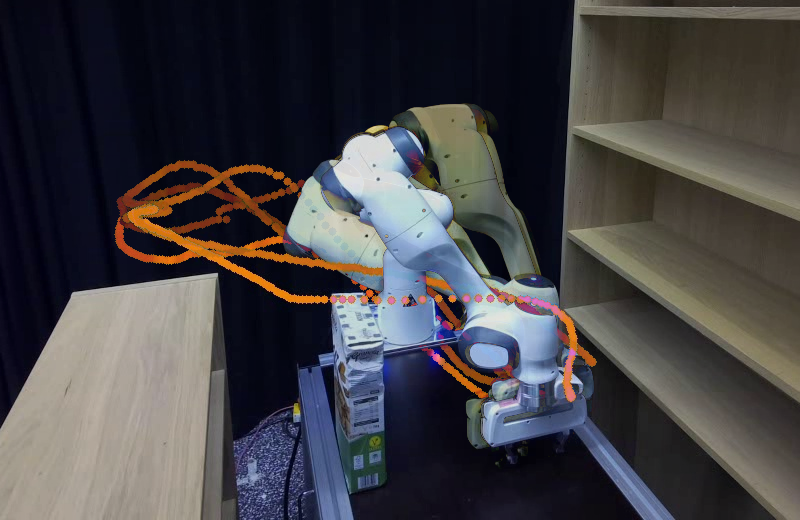}
  \vspace{-0.2cm}
  \caption[Example of multimodal trajectories executed in the warehouse environment using MPD.]{
    Example of multimodal trajectories executed in the warehouse environment using MPD.
    Starting from the same joint configuration, the task is to reach an end-effector pose on the table while avoiding collisions with the environment.
    The orange lines depict the end-effector trajectories, and the overlayed images show the last frame on the trajectory.
    Our model generates multiple goal robot configurations for similar desired end-effector poses.
  }
  \label{fig:EnvWarehouse-RobotPanda-mpd}
\end{figure}

To show that our model is transferable from the digital twin to a real-world scenario, we execute MPD by replicating the EnvWarehouse-RobotPanda environment and adding additional objects to the scene, whose pose we obtain through a marker-based system.
\Cref{fig:EnvWarehouse-RobotPanda-mpd} shows an example of how MPD generates multimodal trajectories with different joint goal configurations for similar end-effector poses.

\subsection{MPD vs. Diffusion Prior and Cost Optimization}
\label{sec:mpd_vs_diffusion_prior_then_guide}

\begin{figure}[!t]
    \centering

    \hfill
    \subfloat[Demonstrations]{%
        \includegraphics[width=0.30\columnwidth]{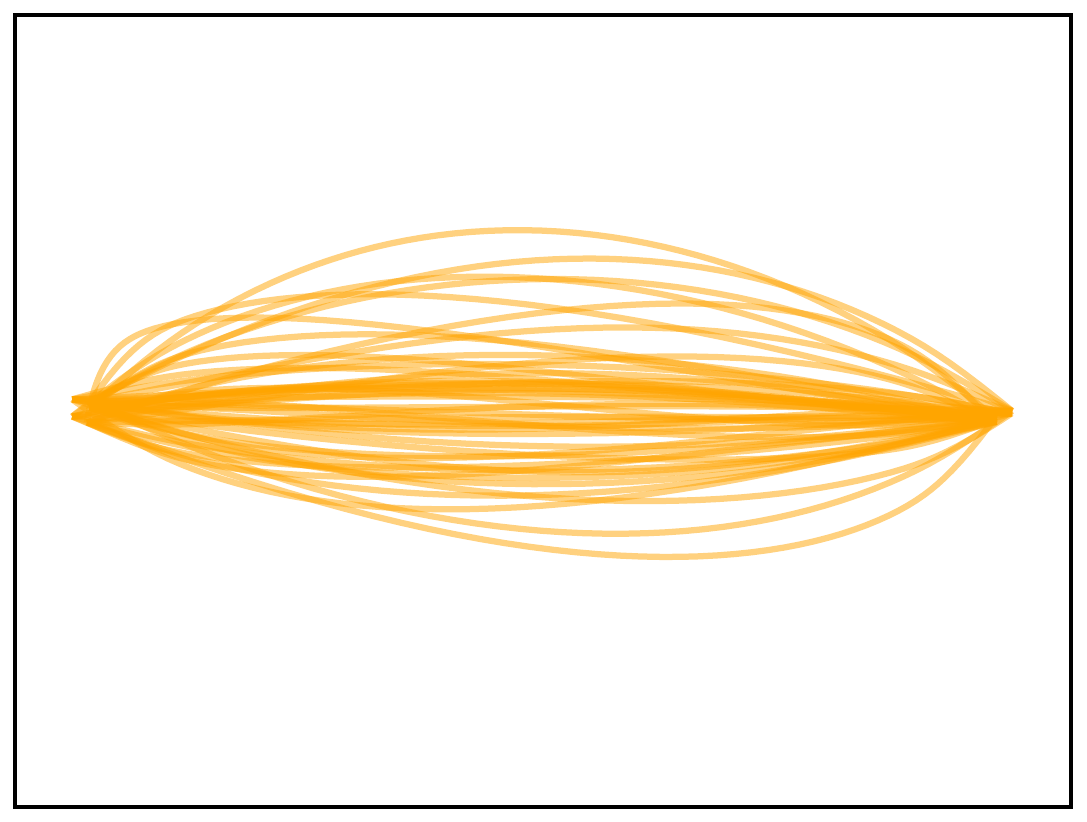}
        \label{fig:EnvEmpty2D_demonstrations}
    }
    \hfill
    \subfloat[Dprior]{%
        \includegraphics[width=0.30\columnwidth]{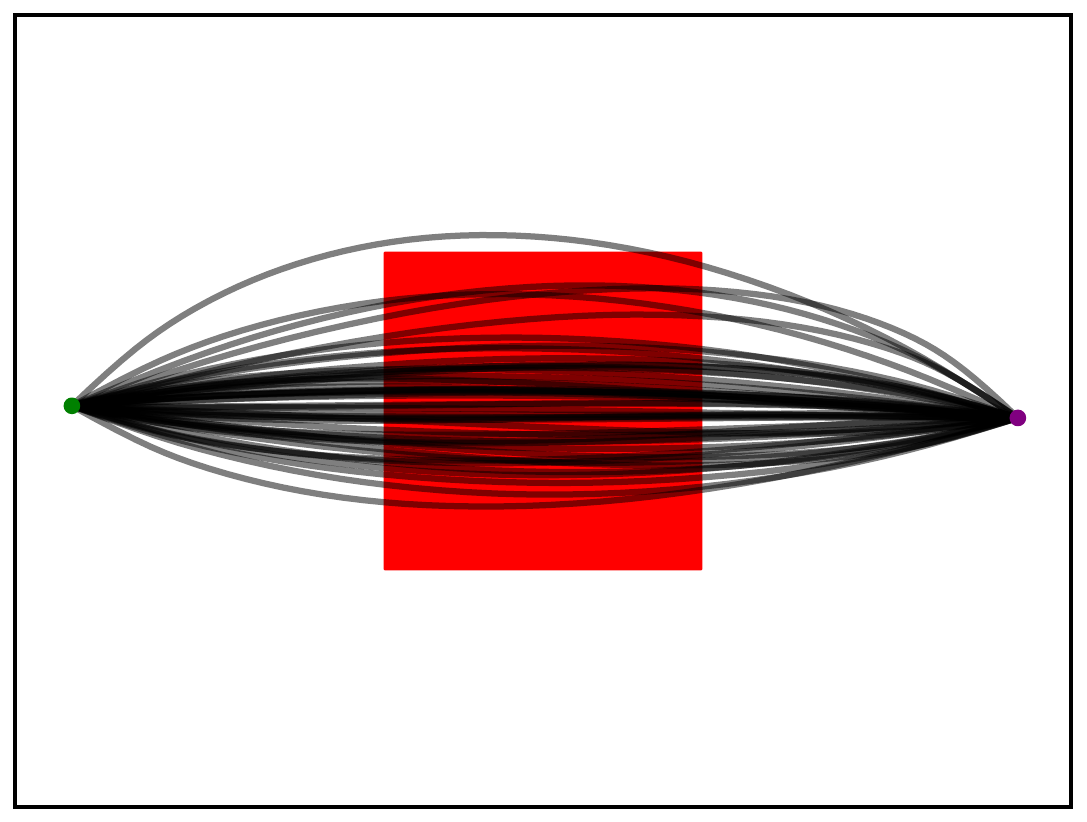}
        \label{fig:EnvEmpty2D_diffusion_prior}
    }
    \hfill
    \subfloat[Dprior + Cost]{%
        \includegraphics[width=0.30\columnwidth]{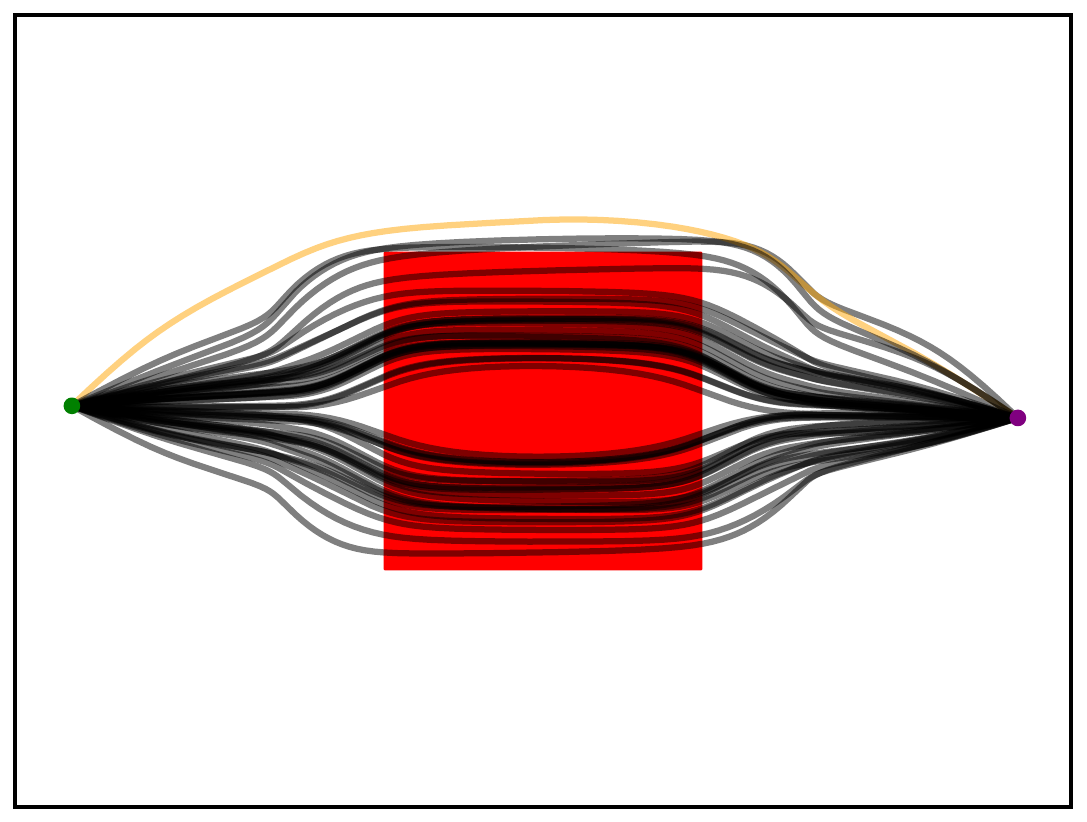}
        \label{fig:EnvEmpty2D_diffusion_prior_then_guide}
    }
    \hfill
    \\

    \subfloat[MPD]{%
        \label{fig:EnvEmpty2D_mpd}

        \hfill
        \includegraphics[width=0.23\columnwidth]{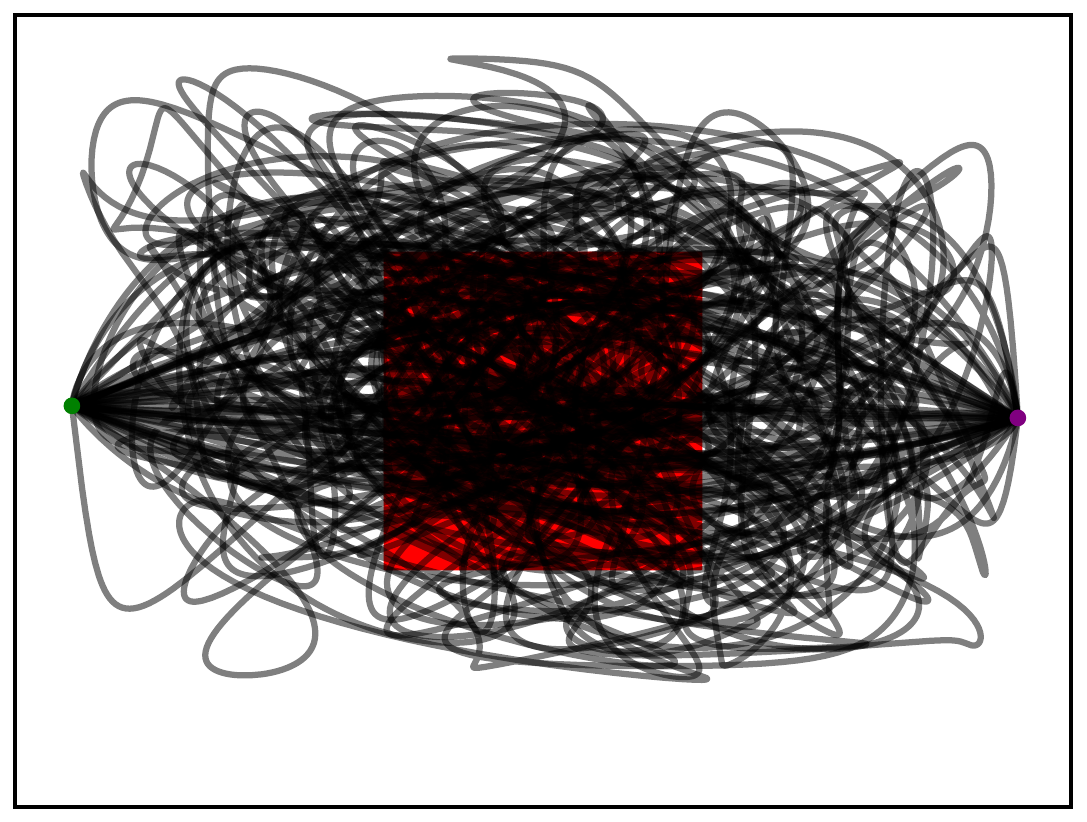}
        \hfill
        \includegraphics[width=0.23\columnwidth]{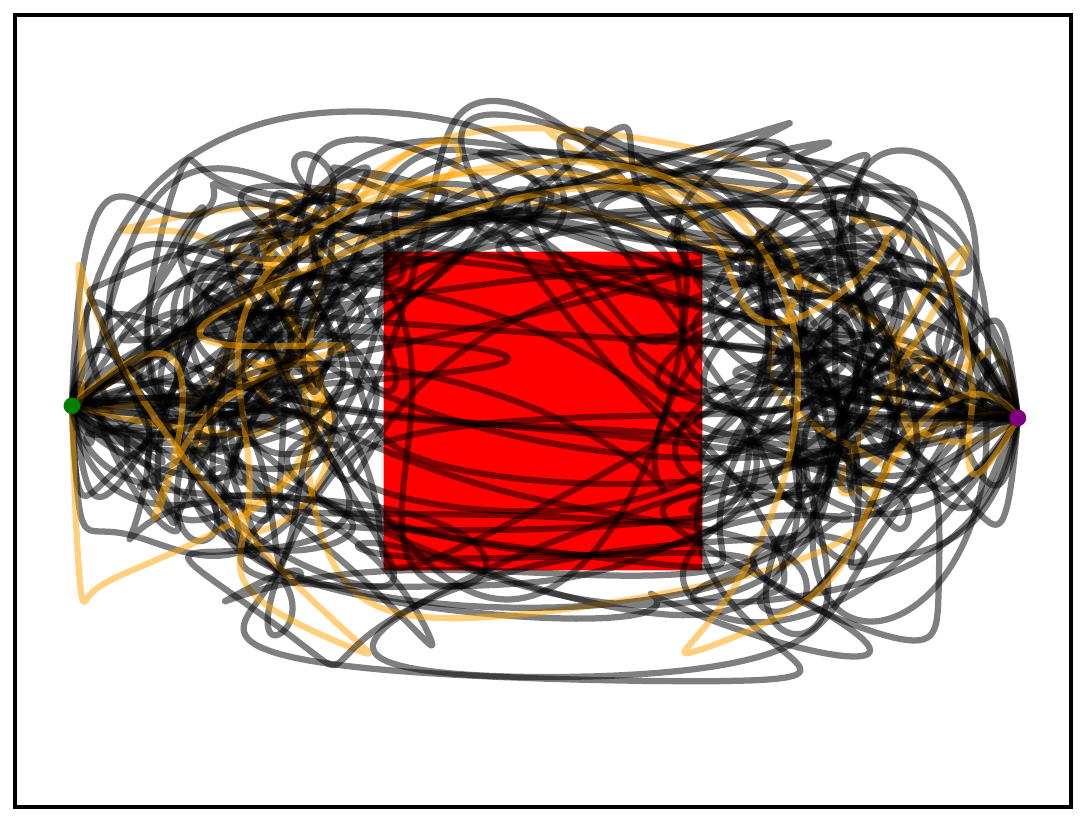}
        \hfill
        \includegraphics[width=0.23\columnwidth]{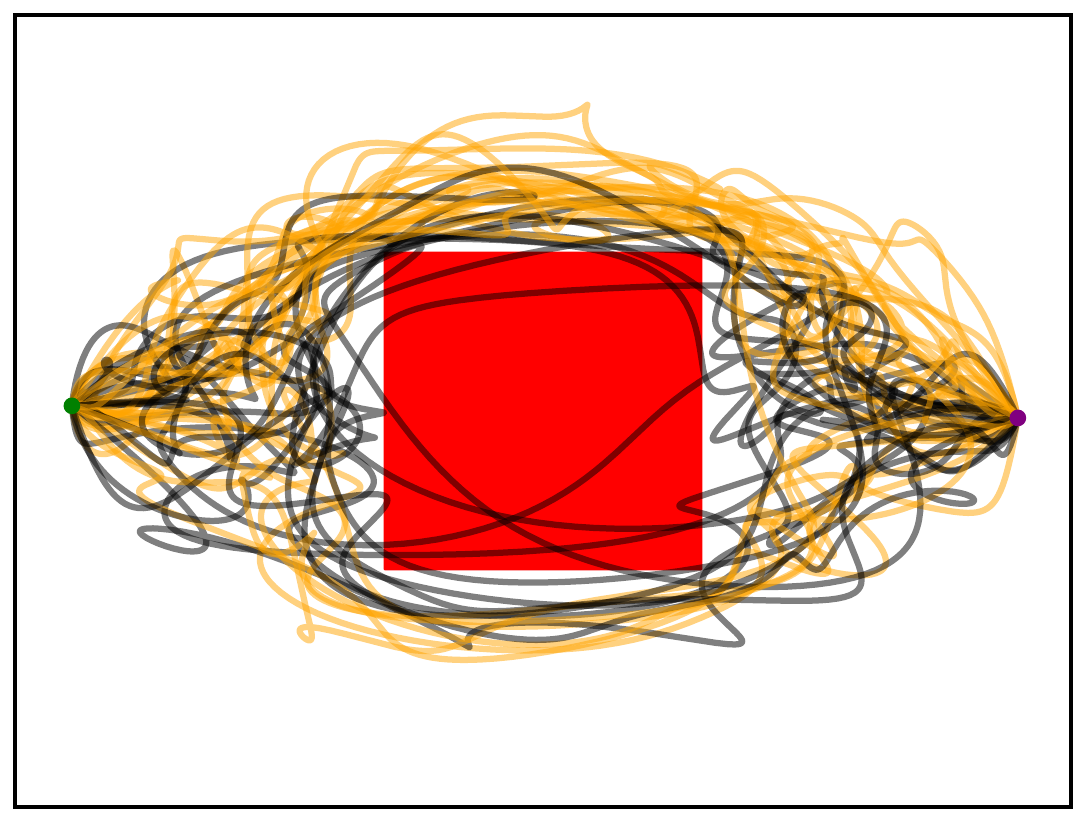}
        \hfill
        \includegraphics[width=0.23\columnwidth]{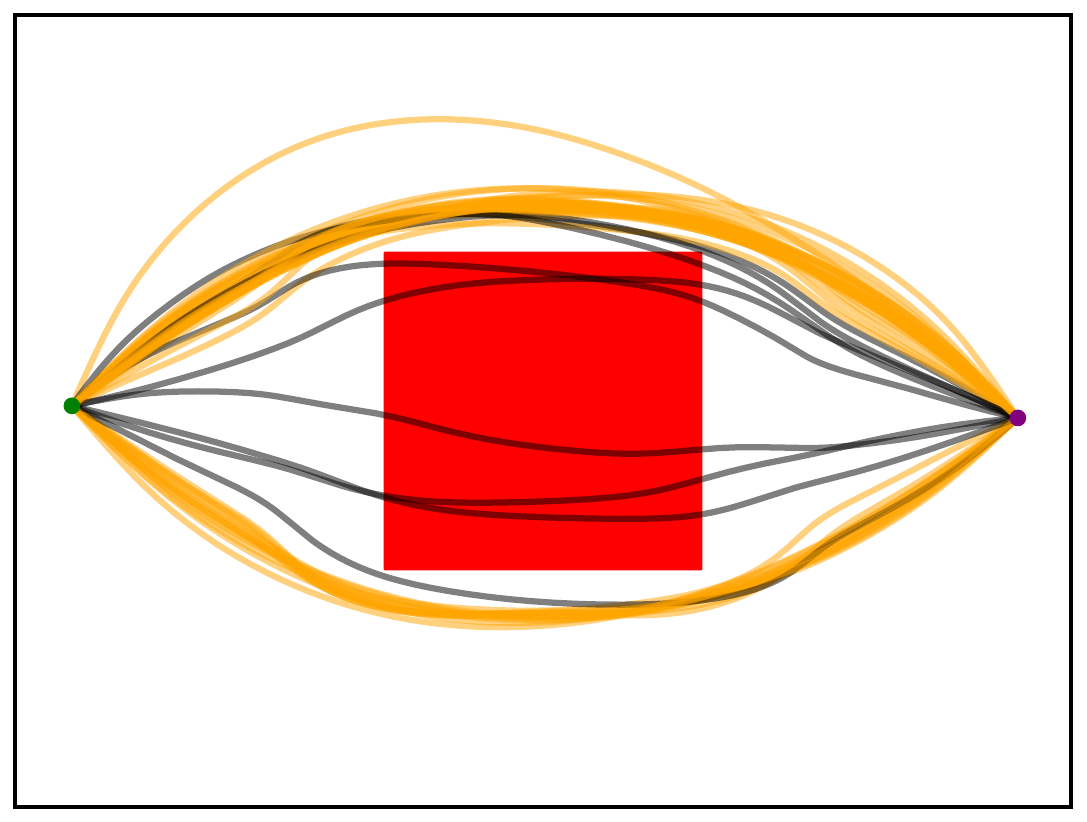}
        \hfill
    }
    \vspace{0.25cm}
    \noindent\begin{tikzpicture}
    \draw[->, line width=2pt] (0,0) -- (\columnwidth-30pt,0)
        node[below=0.1cm, align=left, at start] {$i=N$}
        node[below=0.1cm, align=center, midway] {MPD denoising process}
        node[below=0.1cm, align=right, at end] {$i=0$};
    \end{tikzpicture}

    \caption[Illustration of the benefits of sampling with MPD in a $2$-dimensional task.]{
        The figures illustrate the benefits of sampling with MPD instead of first sampling from the prior and then optimizing the cost function.
        \protect\subref{fig:EnvEmpty2D_demonstrations}
        The robot is a $2$D point mass in an environment without any obstacles, and demonstration trajectories are obtained by sampling from a Gaussian Process prior.
        At inference, an obstacle (in red) is inserted into the environment.
        \protect\subref{fig:EnvEmpty2D_diffusion_prior} Samples from the learned diffusion prior.
        \protect\subref{fig:EnvEmpty2D_diffusion_prior_then_guide} Results of first sampling from the prior and then optimizing the collision cost function.
        \protect\subref{fig:EnvEmpty2D_mpd} Samples from the posterior distribution generated with MPD.
        The second row illustrates how MPD's denoising process moves from Gaussian noise to clean trajectories while avoiding the red obstacle.
        Trajectories that are in collision with the red obstacles are shown in black, and collision-free trajectories in orange.
    }
    \label{fig:EnvEmpty2D_diffusion_prior_guide_mpd}

    \vspace{-0.25cm}    
\end{figure}

In this section, we answer Q3.
One important aspect when using cost-guided diffusion in the context of motion planning is to understand in what situations can sampling from the denoising posterior distribution \cref{eq:ddpm_posterior} be beneficial, compared to first sampling from the diffusion prior and then optimizing the cost function starting from the proposed samples.
In \cref{fig:EnvEmpty2D_diffusion_prior_guide_mpd} we show a simple example to motivate this difference.
We consider a $2$D point mass in an environment without any obstacle (\cref{fig:EnvEmpty2D_demonstrations}), generate a small set of demonstrations with a Gaussian Process trajectory, and learn a trajectory diffusion prior.
At inference we add a new obstacle, the red square, and do optimization using the object collision cost.
The figures show two aspects of MPD.
First, while sampling from the diffusion prior and then optimizing the cost fails to provide collision-free trajectories (\cref{fig:EnvEmpty2D_diffusion_prior_then_guide}), MPD can obtain a larger percentage of those (\cref{fig:EnvEmpty2D_mpd}).
This phenomenon happens because once a large portion of the trajectory is in collision with the obstacle, it becomes more difficult to remove it from collision.
Second, in \cref{fig:EnvEmpty2D_mpd}, we notice how trajectory samples from MPD evolve during the denoising process.
The evolution shows that sampling from the posterior distribution moves the samples towards regions of the configuration space where the posterior has higher probability density, thus generating samples that are collision-free but close to the prior distribution.
Therefore, the benefits of sampling from the posterior are more diverse and collision-free trajectories, which can be observed in \cref{fig:results_simulation_swarmplots} by comparing Dprior+Cost and MPD results, especially in the environments with additional objects.

\subsection{Impact of the B-spline Parametrization}
\label{sec:experiments_bspline_vs_waypoints}

\begin{figure}[t]
    \centering

    \includegraphics[width=0.90\columnwidth]{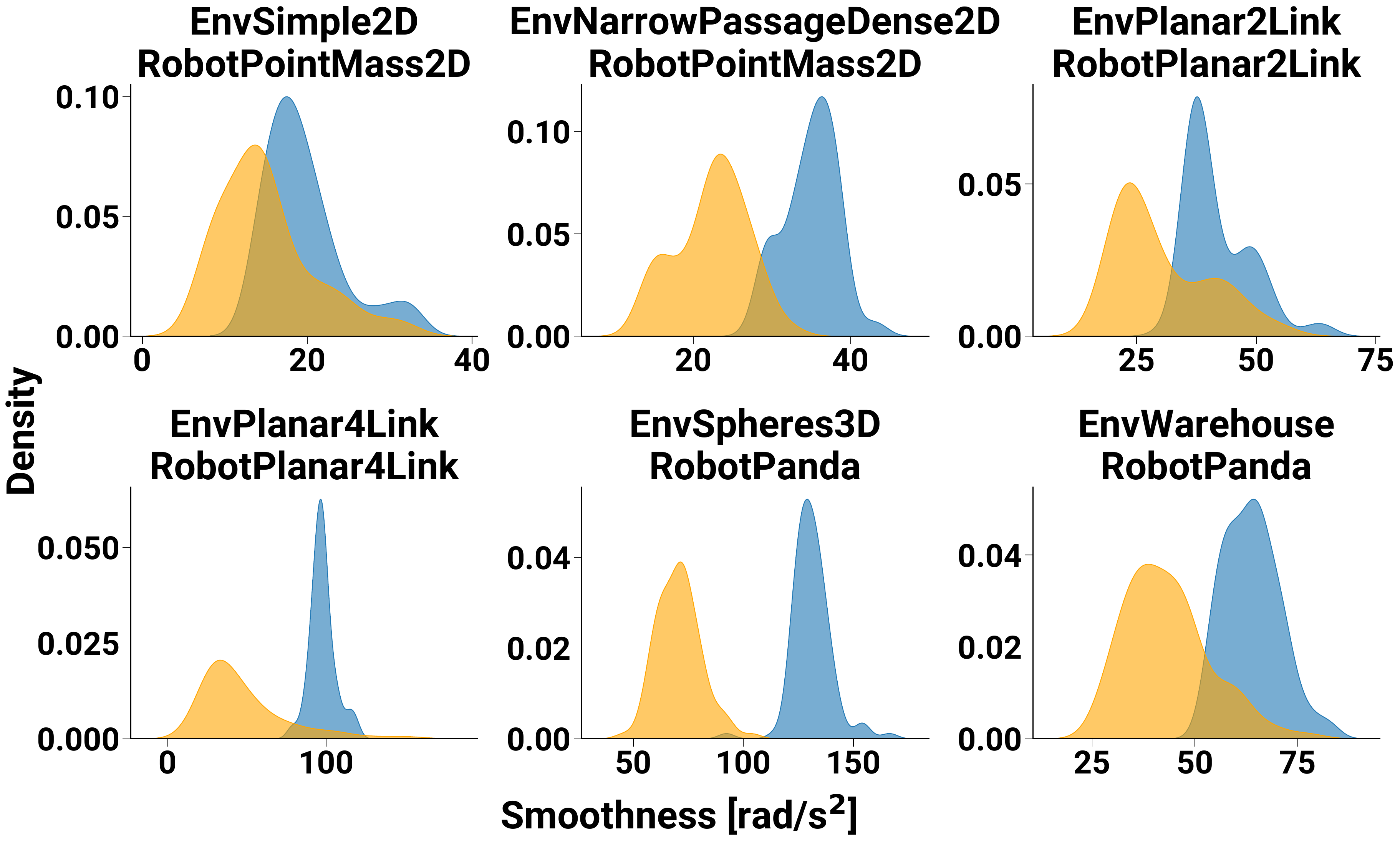}
    \\
    \includegraphics[width=0.5\columnwidth]{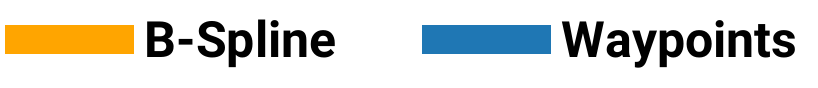}

    \vspace{-0.35cm}    
    
    \caption[KDE plots smoothness.]{
        KDE plots of the smoothness distribution of all contexts in all tasks with additional objects (lower values mean smoother trajectories).
    }
    \label{fig:bspline_vs_waypoints_kde}

    \vspace{-0.5cm}    
\end{figure}

\begin{figure}[!t]
    \captionsetup[subfigure]{labelformat=empty}
    \centering
    \begin{minipage}[t]{0.99\columnwidth}
        \centering
        \subfloat[(a) B-spline]{\raisebox{-\height}{\includegraphics[width=0.20\textwidth]{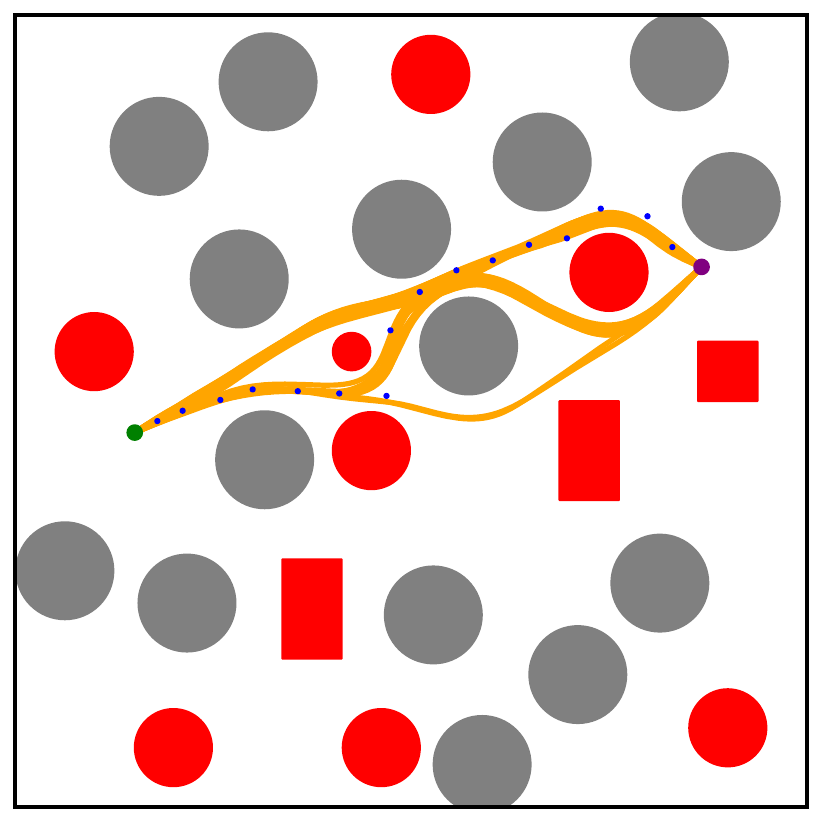}}}%
        \subfloat{\raisebox{-\height}{\includegraphics[width=0.79\textwidth]{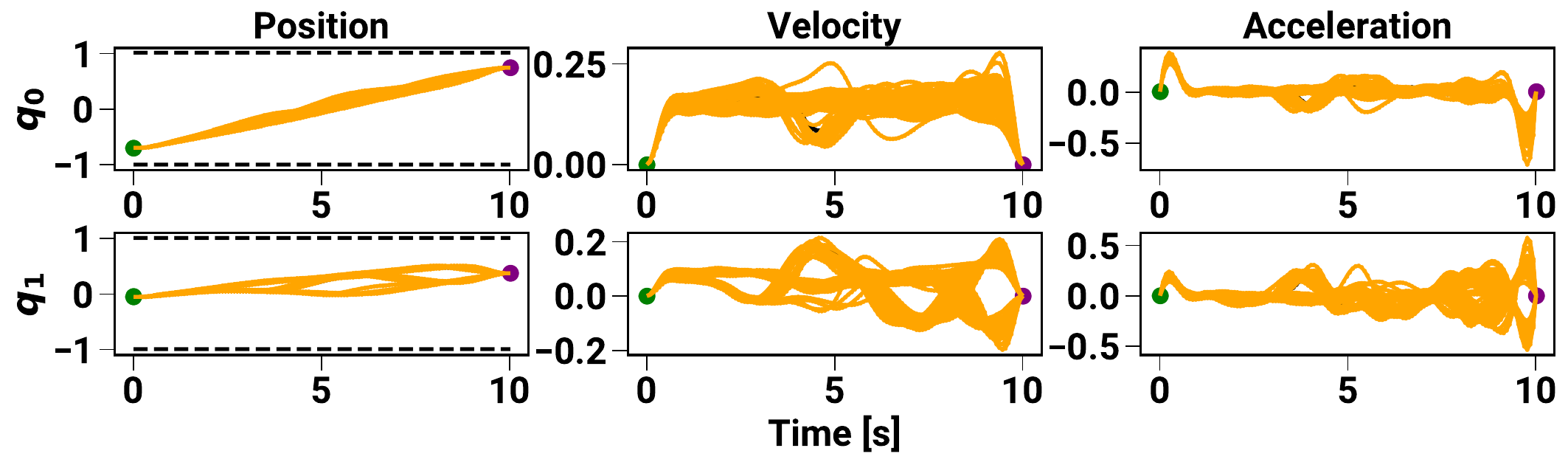}}}%

        \vspace{-0.25cm}
        \subfloat[(b) Waypoints]{\raisebox{-\height}{\includegraphics[width=0.20\textwidth]{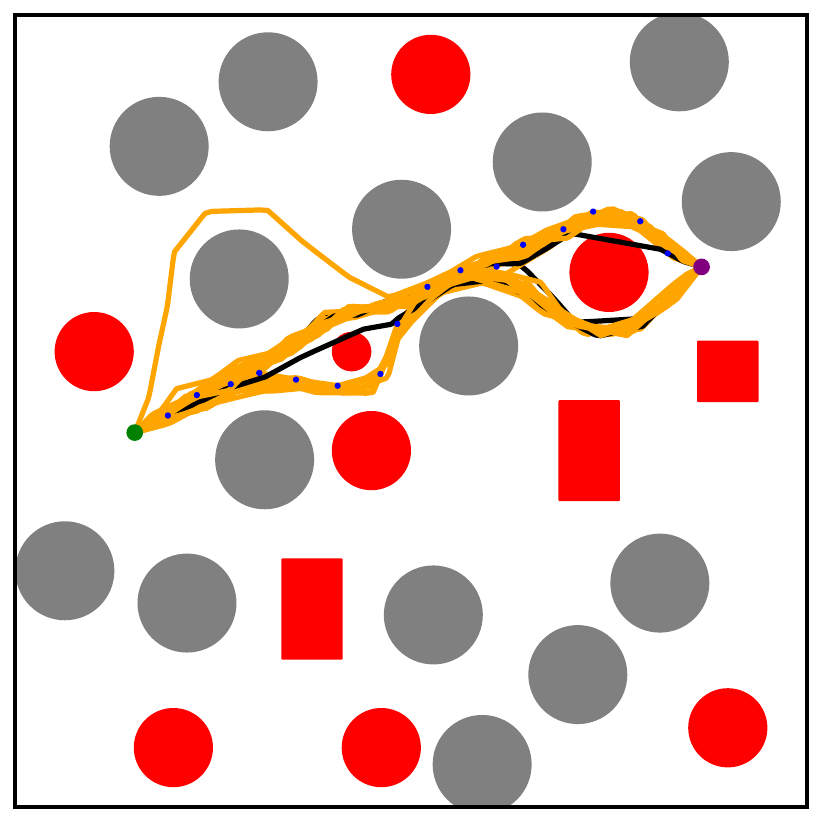}}}%
        \subfloat{\raisebox{-\height}{\includegraphics[width=0.79\textwidth]{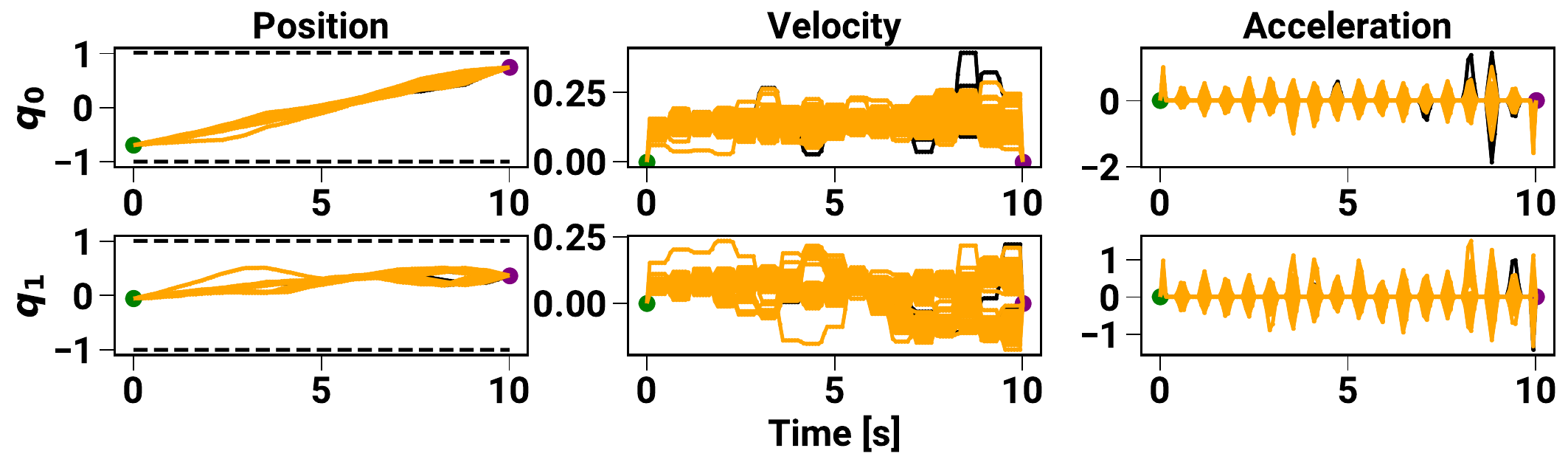}}}%
    \end{minipage}%
    \hfill

    \caption[Illustration of MPD generated trajectories in the EnvSimple2D-RobotPointMass2D task with B-spline and waypoint trajectory representations.]{
        MPD results in the EnvSimple2D-RobotPointMass2D task using different trajectory parameterizations.
        Trajectories in orange are collision-free, and black in collision.
        (a) The first row shows the generated trajectories using the B-spline parametrization, while (b) the second one is generated with waypoints.
        By inspecting the trajectories' evolution, we observe jumps in the acceleration profile due to using a few waypoints.
        At the same time, the B-spline parametrization ensures smoothness in velocity and acceleration profiles.
        The blue dots in the environment plots show the control points or waypoints of one trajectory.
    }
    \label{fig:bspline_vs_waypoints}
    
    \vspace{-0.5cm}
\end{figure}

One of this work's contributions is performing diffusion in the space of B-spline coefficients instead of waypoints.
Hence, to answer Q4, we compare both representations.
The benefits have been detailed in \cref{sec:trajectory_parametrization}, but perhaps the main one is generating smooth trajectories.
As the velocity and acceleration of a linear waypoint trajectory representation are computed with central finite differences, they are constant between waypoints and lead to large jumps.
There are two ways to get smoother trajectories with this representation.
First, we can increase the number of waypoints, as commonly done in optimization-based algorithms.
Second, after running the denoising process, we can do extra optimization steps to smoothen and optimize the trajectory to remain collision-free and within joint limits while minimizing the task cost.
Both of these options have drawbacks in terms of additional computations, e.g., in the first case, the denoising network needs to process larger inputs.
We performed an ablation experiment where we replaced the B-spline with waypoints.
We used the same number of waypoints as B-spline coefficients, which guarantees that the denoising network size and computations 
are similar.
As expected, there are no major differences across all metrics except for smoothness values~\footnote{Smoothness is measured as the sum of accelerations along the trajectory.}, which are better (lower) for the B-spline compared to the sparse waypoint representation.
To better grasp the difference visually, \cref{fig:bspline_vs_waypoints_kde} shows the distribution of smoothness values obtained with MPD across all contexts and tasks.
The waypoint representation consistently leads to worse smoothness.
\Cref{fig:bspline_vs_waypoints} shows a visual comparison in the EnvSimple2D-RobotPointMass2D tasks.
\begin{wrapfigure}{r}{0.45\columnwidth}
  \vspace{-0.3cm}
  \centering
  \includegraphics[width=0.99\linewidth]{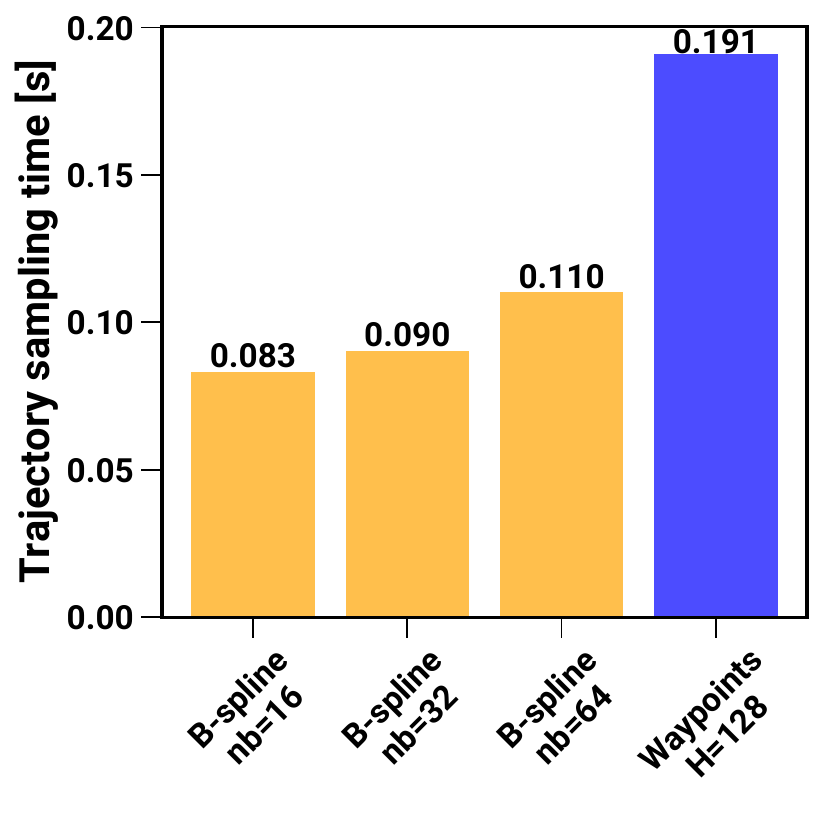}
  \vspace{-1cm}
  \caption[
    Bspline vs waypoints computation time.
    ]{
    Computation times for denoising a batch of $1000$ trajectories with B-splines and dense Waypoints using $H=128$ and $d=7$ with $15$ steps of DDIM.
  }
  \label{fig:bspline_vs_waypoints_computation_time}
  \vspace{-0.35cm}
\end{wrapfigure}
The figures show that even though the waypoint representation can generate valid trajectories, they are not smooth, as can be seen in the straighter path in the environment plot and in the velocity and acceleration time profiles.
These types of trajectories are harder for a robot controller to follow and can lead to jerky movements.

\reviewersix{
A benefit of B-splines in comparison to using a dense trajectory is the faster generation time, as the denoising function processes inputs with lower dimensions.
To generate trajectories with $H$ points, we can sample from a model that outputs this trajectory size, or we can sample $n_b < H$ control points of a B-spline, and then interpolate the trajectory to $H$ using \cref{eq:bspline}.
\Cref{fig:bspline_vs_waypoints_computation_time} shows the computation times when generating a batch of trajectories with B-splines for an increasing number of control points, resulting from sampling the diffusion model and interpolating with \cref{eq:bspline}, and sampling waypoint trajectories with $H=128$.
We consider only denoising times and interpolation (whose time is minimal), since the computation of costs/gradients only depends on the number of dense points $H$, and is the same for B-splines or dense waypoint representations.
The results show the computational benefit of performing diffusion with lower-dimensional representations, instead of dense trajectories.
}

\subsection{Learning and Adapting from Human Demonstrations}

\begin{figure}[!t]
    \centering
    \includegraphics[width=0.72\columnwidth]{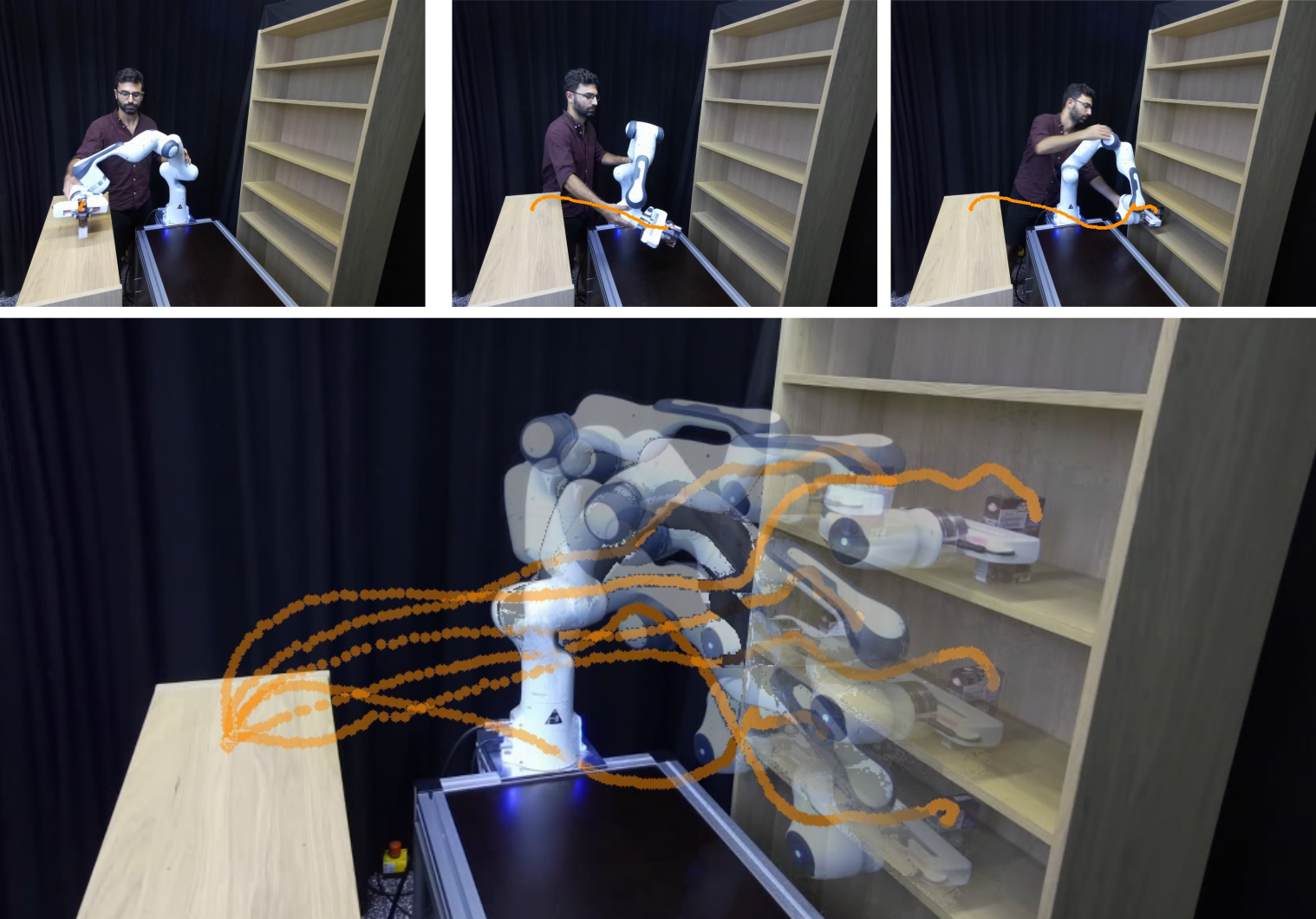}
    \caption[Human demonstrations via kinesthetic teaching for a place task in the EnvWarehouse-RobotPanda environment.]{
        (Top)
        Human demonstrations via kinesthetic teaching for a pick-and-place task in the EnvWarehouse-RobotPanda environment.
        The end-effector trajectories are depicted in orange.
        (Bottom)
        Overlay of several demonstrations. The robot configurations are shown at the end of the demonstration.
    }
    \label{fig:human_demonstrations}
    \vspace{-0.3cm}
\end{figure}

Learning the prior distribution on trajectories is agnostic to the expert demonstrations.
Previously, we assumed that trajectories would be generated by a sampling-based path planner.
However, in real-world tasks, it is sometimes desirable and easier for a human to provide demonstrations via kinesthetic teaching of what movement they would like the robot to execute.
Therefore, to answer Q5, in this experiment, we learn the trajectory distribution directly from human demonstrations, showing that MPD can be used in this scenario as well.
Using the EnvWarehouse-RobotPanda environment, we consider the task of moving one item (a tea box) from several locations in the small to the big shelf and letting a user perform multiple demonstrations from random start and goal poses by moving the robot in gravity compensation with kinesthetic teaching from regions on the small shelf to the big shelf, generating approximately $100$ trajectories.
These are augmented using small noise values to perform a total of approximately $4$k trajectories.
Examples of human-demonstrated trajectories are shown in \cref{fig:human_demonstrations}.
The diffusion prior model is then trained on joint space trajectories similarly and with the same hyperparameters as the simulation experiments.

\begin{figure}[!t]
    \centering
    \subfloat[Diffusion prior]{
        \includegraphics[width=0.30\columnwidth]{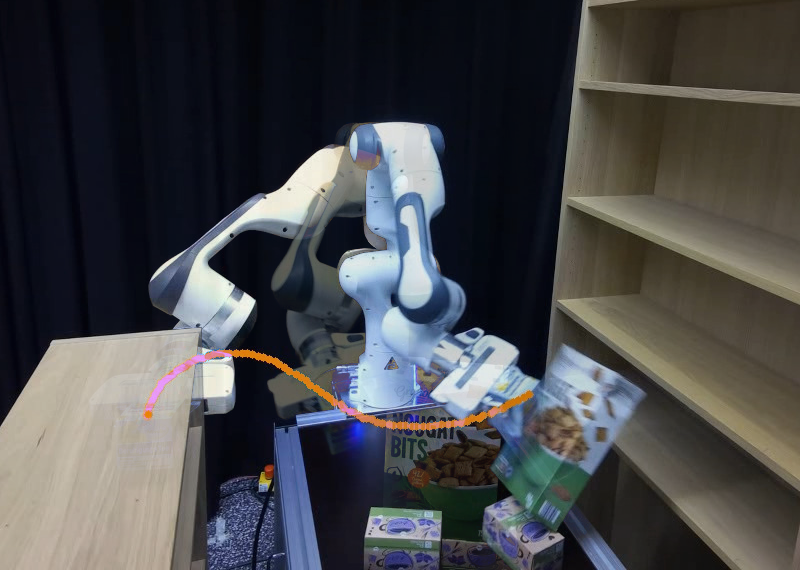}
        \label{fig:real_world_diffusion_prior_fail}
    }
    \hfill
    \subfloat[MPD]{
        \includegraphics[width=0.30\columnwidth]{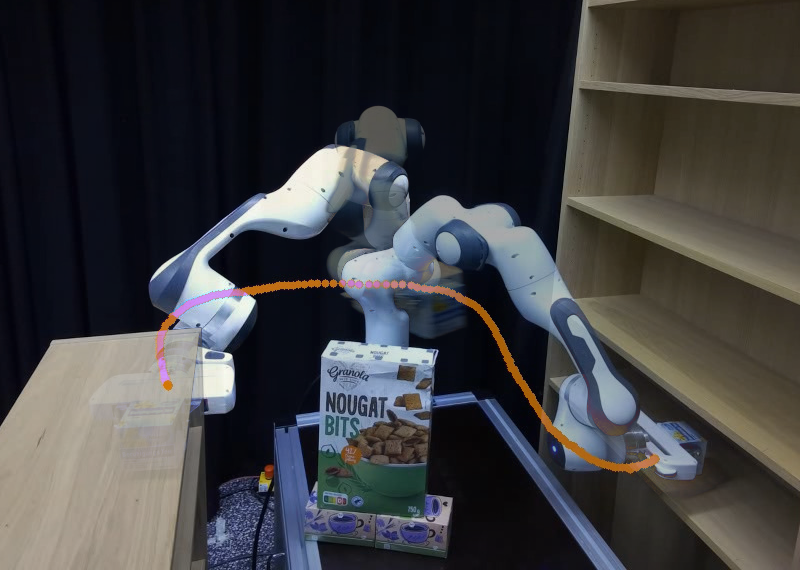}
        \includegraphics[width=0.30\columnwidth]{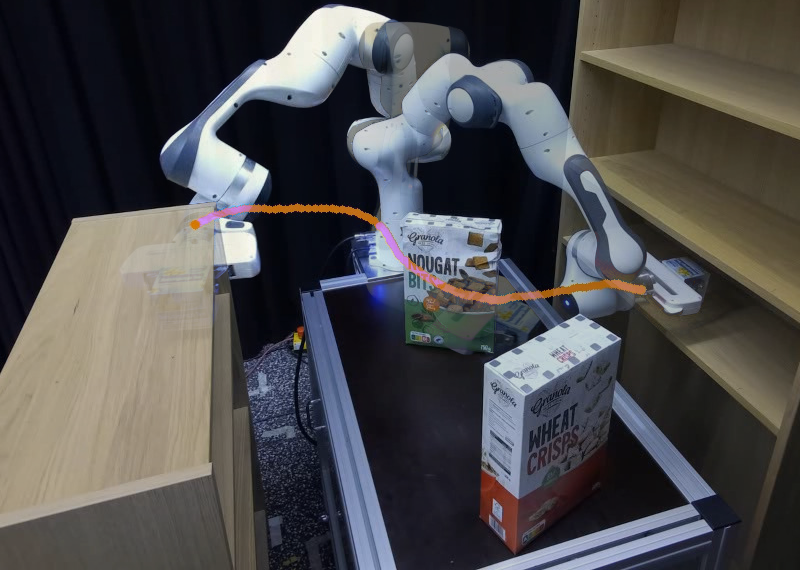}
        \label{fig:real_world_mpd_success}
    }
    \caption[Illustration of trajectories generated with the diffusion prior and MPD.]{
        (a)
        As the demonstrations did not include the objects placed on the table, the diffusion prior failed to obtain a collision-free trajectory and crashed against the obstacle.
        (b)
        By using the collision-avoidance cost during posterior sampling, MPD solutions can stay close to the human demonstrations (see \cref{fig:human_demonstrations}) while avoiding collisions with the obstacles and reaching the desired end-effector pose.
    }
    
    \vspace{-0.5cm}
\end{figure}

For these experiments, we only consider diffusion-based methods.
We tested MPD by placing additional objects in the scene (cereal and tea boxes) (see \cref{fig:real_world_mpd_success}), and computed the environment's signed distance function using the updated object poses, estimated with a marker-based system.
For inference, we use the same hyperparameters as the simulation experiments to generate trajectories from the current robot's joint position to a desired end-effector goal pose, which we assume to be determined by a higher-level task planner.

\begin{table}[!t]
    \centering
    \small
    \caption[Performance metrics for a real-world task using MPD.]{
         Performance metrics for a real-world motion planning task in the EnvWarehouse-RobotPanda environment with additional objects using diffusion-based models.
    }
    \label{tab:real_world_results}

    \resizebox{\columnwidth}{!}{%

    \begin{tabular}{l|rrr}
        \textbf{Algorithms} & \textbf{Success rate} [\%] & \textbf{Fraction valid} [\%] & \textbf{Diversity} \\
        \hline
        Dprior & $0.0$ & $-$ & $-$ \\
        Dprior+Cost & $100.0$ & $37.0$ & $27.95$ \\
        MPD & $100.0$ & $78.0$ & $59.50$ \\
        \midrule
    \end{tabular}

    }

    \vspace{-0.3cm}
    
\end{table}

The relevant metrics and results across different contexts are found in \cref{tab:real_world_results}.
The diffusion prior fails to obtain any collision-free trajectory since the prior trajectory distribution passes through the objects, as shown in an example in \cref{fig:real_world_diffusion_prior_fail}.
Both Dprior+Cost and MPD achieved a $100\%$ success rate while finding different modes for solving the task.
\Cref{fig:real_world_mpd_success} shows two valid trajectories obtained with MPD after placing the additional objects in different locations.
These trajectories are collision-free and close to the demonstrations from \cref{fig:human_demonstrations}.

\section{Limitations and Future Work}
\label{sec:limitations}

In this section, we present some limitations of MPD and leave proposals for future work.

To incorporate first-order gradients while sampling from the posterior distribution, we use the approach from classifier-guided diffusion.
Instead, we can use a projection step at each denoising step, such that the sample lies on some manifold (for instance, ensuring safety constraints)~\cite{christopher2024constrained}.
However, the projection can be computationally expensive.

In this work, we kept the trajectory duration fixed, since optimizing it during the denoising process is not trivial, as the phase-time derivative would constantly change. 
Thus, the joint velocity and acceleration limits are not constant in phase space.
A research direction would be to optimize the trajectory duration, possibly using a separate B-spline curve to represent the phase-time derivative from \cref{eq:bspline_rs}~\cite{DBLP:journals/trob/KickiLTBWSP24}.

The current inference speed highly depends on the current Python implementation.
One common appointed disadvantage of diffusion models is that they are slow to sample due to the number of denoising steps.
However, in our problem, much computation time is used for the cost function gradients.
Even though single-cost computation is not slow (done at $\approx 100$Hz) and parallelizable, the costs must be computed sequentially due to the Python GIL.
Using other programming frameworks, such as JAX~\cite{jax2018github}, which uses JIT compilation, might be an alternative to improve speed further.

We considered environments that do not undergo major structural changes.
These are common across several scenarios, such as warehouses, where shelves are not constantly moved, but new obstacles can appear.
This work can be extended with an additional context variable that encodes the current environment.
However, for proper generalization, this would require a large amount of data to be collected to cover a large possibility of arrangements across many environments~\cite{fishman2022mpinets}.
Instead of generalizing, our work focuses on specializing in a single environment (e.g., as in~\cite{bency2019neuralpathplanning}) and showing the properties of diffusion for motion planning.

\section{Conclusion}
\label{sec:conclusion}

In this article, we proposed Motion Planning Diffusion (MPD), an algorithm that uses diffusion for learning-based motion planning tasks.
A novelty in our approach is to parametrize the trajectory using B-spline coefficients and model the denoising network in this weight space.
Compared to a waypoint representation, using the same number of coefficients, this parametrization ensures smoothness and processing of smaller inputs.
During training, the diffusion model learns a distribution over control points obtained by fitting a B-spline to paths obtained from a sampling-based planner or human demonstrations.
At inference time, given a motion planning cost function to optimize, which includes avoiding object collisions, promoting joint limits, and reaching an end-effector goal pose, we propose using planning-as-inference to obtain a distribution of trajectories that balances the prior distribution and the cost likelihood.
We do this by sampling the denoising posterior distribution using cost-guided diffusion.
Therefore, the generated samples are strongly biased toward prior solutions, which are more useful than a uniformed prior, such as a straight-line trajectory in configuration space.
To show generalization, we add additional objects to the environments used during training, leading to adaptation to new environments.

Our experimental results show that diffusion models are desirable priors for encoding trajectory distributions because they model multimodality quite well, and blending sampling and optimization is beneficial in generating collision-free trajectories.
We empirically show and provide intuition on why sampling from the posterior is beneficial compared to sampling from the prior and then optimize the cost function using those samples as initialization.
Along with learning from paths generated in simulation, we also learn a real-world pick-and-place task from human trajectory demonstrations via kinesthetic teaching.
Across all simulated and real-world tasks, MPD either matches or surpasses the evaluated baselines, showing it can generate a higher percentage of valid trajectories while keeping a higher diversity in the generated samples.

\section*{Acknowledgments}

This work was funded by the German Federal Ministry of Education and Research projects IKIDA (01IS20045)
and 
Software Campus project ROBOSTRUCT (01S23067),
and 
by the German Research Foundation project METRIC4IMITATION (PE 2315/11-1)
and supported by the Foundation for Polish Science (FNP).

\begin{appendices}

\section{Hyperparameters Training}
\label{sec:hyperparameters_training}

\begin{table*}[ht]
    \scriptsize
    \centering
    \caption[Hyperparameters for training MPD.]{Hyperparameters for training MPD.}

    \resizebox{\textwidth}{!}{

        \begin{tabular}{lccccccc}
        \hline
        \textbf{Task} & \textbf{Dataset} & \textbf{B-spline} & \textbf{UNet} & \textbf{UNet dim} & \textbf{Context} & \textbf{Batch} & \textbf{Optimization} \\
             & \textbf{size} & \textbf{control points} & \textbf{input dim} & \textbf{multiplier} & \textbf{out dim} & \textbf{size} & \textbf{steps} \\
        \hline
        EnvSimple2D-RobotPointMass2D & 10k & 22 & 32 & (1, 2, 4) & 32 & 128 & 2M \\
        EnvNarrowPassageDense2D-RobotPointMass2D & 10k & 30 & 32 & (1, 2, 4) & 32 & 128 & 2M \\
        EnvPlanar2Link-RobotPlanar2Link & 10k & 22 & 32 & (1, 2, 4) & 32 & 128 & 2M \\
        EnvPlanar4Link-RobotPlanar4Link & 100k & 22 & 32 & (1, 2, 4, 8) & 128 & 512 & 3M \\
        EnvSpheres3D-RobotPanda & 1M & 30 & 32 & (1, 2, 4, 8) & 128 & 512 & 3M \\
        EnvWarehouse-RobotPanda & 500k & 22 & 32 & (1, 2, 4, 8) & 128 & 512 & 3M \\
        \hline
        \end{tabular}
    }

    \label{tab:hyperparameters_training}

    \vspace{-0.5cm}
\end{table*}

\Cref{tab:hyperparameters_training} shows the hyperparameters used for training the diffusion model in each task.
We generate one trajectory per context (start and goal configuration), so the dataset size equals the total number of contexts and trajectories.
The number of control points is task-dependent and chosen such that after fitting a $5$th order B-spline, we obtain over $99\%$ of collision-free paths with a minimal collision rate.
The \textit{\mbox{U-Net} input dim} is the number of dimensions the control points are projected at the network input.
The number of entries in the \textit{U-Net dim multiplier} is the U-Net depth (layers) ($3$ or $4$, depending on the task), while the numbers represent the channels in each layer of the U-Net, which are multiplied by the input dimension.
\textit{Context out dim} is context network output size.
We use a fixed number of $N=100$ diffusion steps for all tasks.
The context and the denoising networks' parameters are optimized using mini-batch gradient descent with the Adam optimizer~\cite{DBLP:journals/corr/KingmaB14} and a learning rate ${3 \times 10^{-4}}$.
The $2$D models took approximately $12$ hours to train, and the more complex ones $24$ hours (these can be improved by early stopping).
Large batch sizes are beneficial when training diffusion because if the dataset has $D$ points, the model is trained for $O$ optimization steps with a batch size $B$, then each data point should be seen roughly $O B/D$ times (approximately the number of epochs).
The diffusion loss function \cref{eq:diffusion_loss} includes an expectation over the $N$ diffusion time steps.
Therefore, the pair of data (control points, diffusion step) is roughly seen $O B / (D N)$ times, which motivates using a larger batch size and a larger number of optimization steps than other generative models.
The CVAE baseline model is trained with the same hyperparameters.

\section{Hyperparameters Inference}
\label{sec:hyperparameters_inference}

\begin{table}[ht]
    \centering
    \caption[Motion planning cost weights.]{Motion planning cost weights (\cref{eq:trajectory_optimization_unconstrained}).}
    \resizebox{\columnwidth}{!}{
        \begin{tabular}{lccccc}
            \textbf{Costs} & Collision & Joint limits & Task & Velocity & Acceleration \\
            \hline
            \textbf{Weights} & 0.9 & 0.5 & 0.5 & 0.2 & 0.2 \\
            \hline
        \end{tabular}
        \label{tab:hyperparameters_inference_cost_weights}    
    }
\end{table}

We use DDIM for sampling from the diffusion model with $15$ steps and a quadratic noise schedule to focus more time steps in lower noise regions.
Cost guidance is activated at the last $i_{\text{cost}} = 3$ steps of denoising, the number of intermediate gradient steps is set to $M=4$, and $\lambda_{\text{prior}} = 0.25$ (see \cref{alg:mpd_algorithm_planning}), amounting to $12$ cost gradient steps.
The gradient weight is $\gamma = 1.0$ and $\delta = 0.15$ (note that the update is done in the unit-normalized control points space).
We found that good values for $i_{\text{cost}}$ to be approximately $1/3$ (or less) of the total denoising steps, and $\lambda_{\text{prior}}$ in $[0.25, 0.5]$.
$M$ and $i_{\text{cost}}$ are a tradeoff between using more cost guidance vs. more denoising steps from the prior.
If used for sampling within the training environments (without additional objects), $M$ can be lowered.
The cost weights reflect the emphasis put on each cost and are task-dependent, but not algorithm-dependent.
These are shown in \cref{tab:hyperparameters_inference_cost_weights}.
The trajectory is interpolated to $128$ points using a B-spline or waypoint parametrization to evaluate the costs and compute gradients on the dense trajectory representation.
We did not vary the task duration and left its optimization for future work.
Therefore, in \cref{sec:general_results_simulation} we use $10$s (to report collision-free results), and $5$s for the experiment in \cref{fig:planar4link_results} (to show the effect of joint limits costs).
In the real-world tasks, we kept the duration at $6$s to safely operate the robot.

\end{appendices}

\printbibliography

\section*{Biographies}

\begin{IEEEbiography}[{\includegraphics[width=1in,height=1in,clip,keepaspectratio]{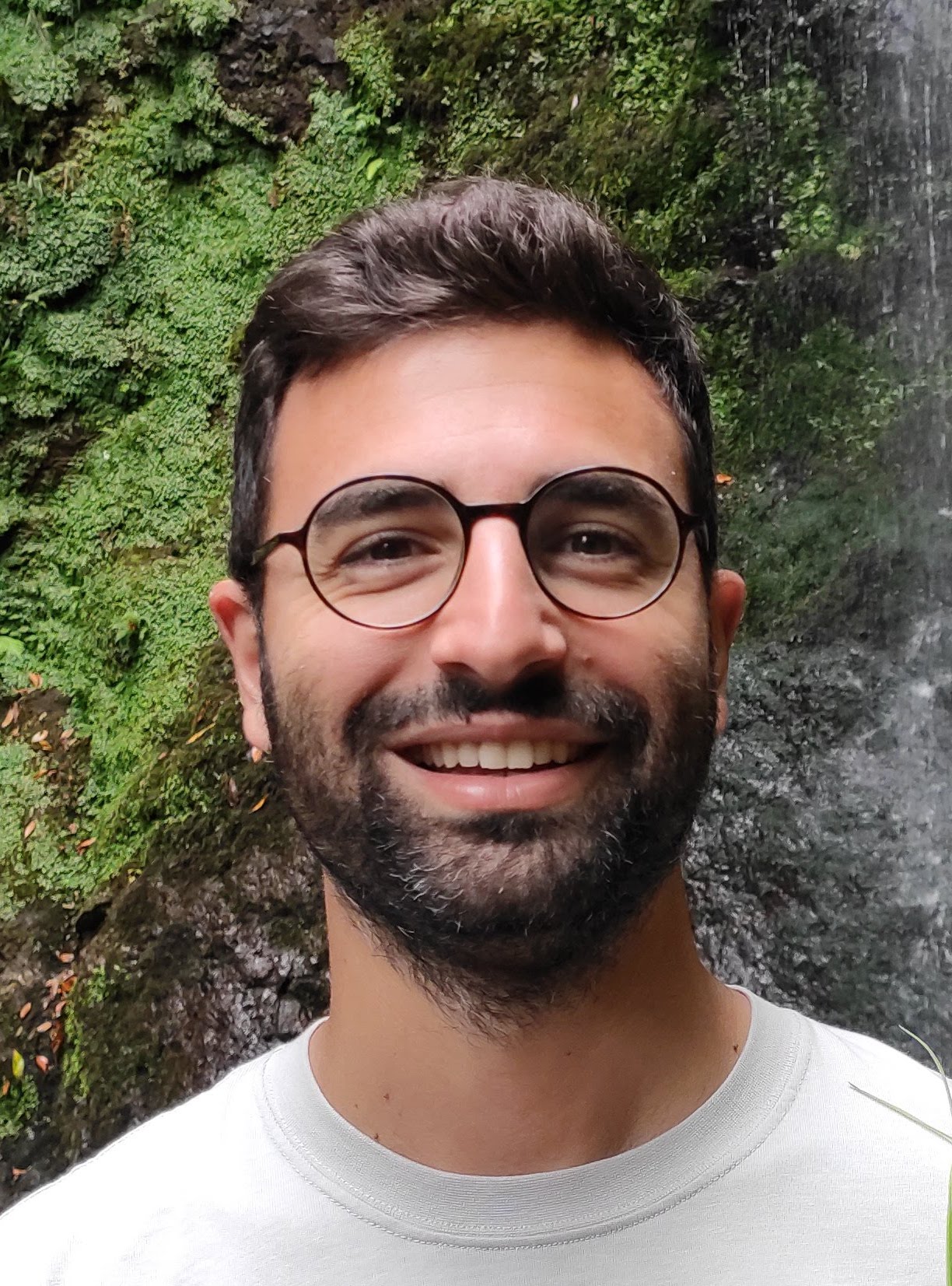}}]{Jo\~{a}o Carvalho}
is a Postdoctoral Researcher at the Intelligent Autonomous Systems Lab of the Technical University of Darmstadt,
where he obtained his Ph.D. in January 2025.
He previously obtained his M.Sc. degree in Computer Science from the University of Freiburg.
His research interests are centered around learning approaches for robot manipulation, from generative models for robot motion planning, imitation learning, and grasping. 
\end{IEEEbiography}

\vspace{-1.0cm}

\begin{IEEEbiography}[{\includegraphics[width=1in,height=1in,clip,keepaspectratio]{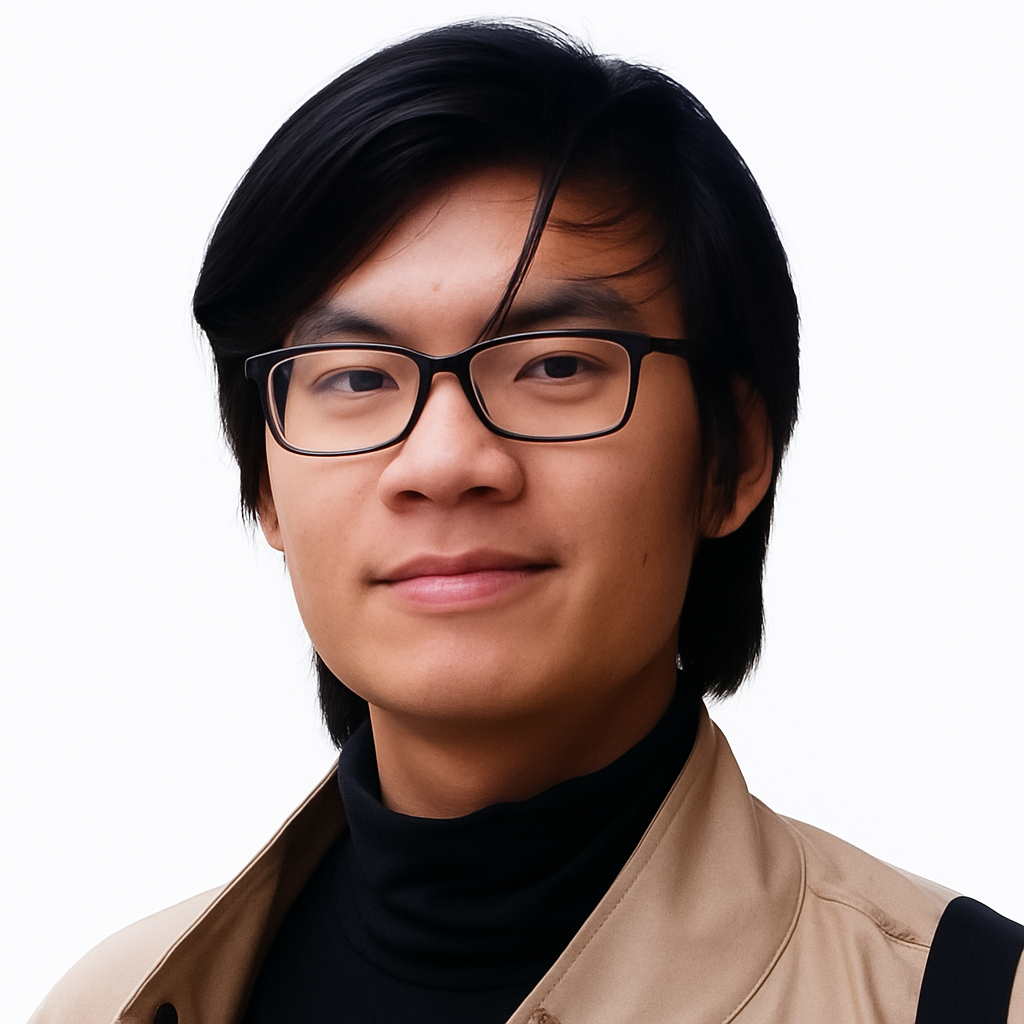}}]{An Thai Le}
is a Ph.D. candidate at the Intelligent Autonomous Systems Lab of the Technical University of Darmstadt.
He obtained his M.Sc. degree from the University of Stuttgart (Germany).
His research interests are related to scaling planning methods to long-horizon, high-dimensional state-space, number of plans, and number of agents, using batch planning methods.
\end{IEEEbiography}

\vspace{-1.0cm}

\begin{IEEEbiography}[{\includegraphics[width=1in,height=1in,clip,keepaspectratio]{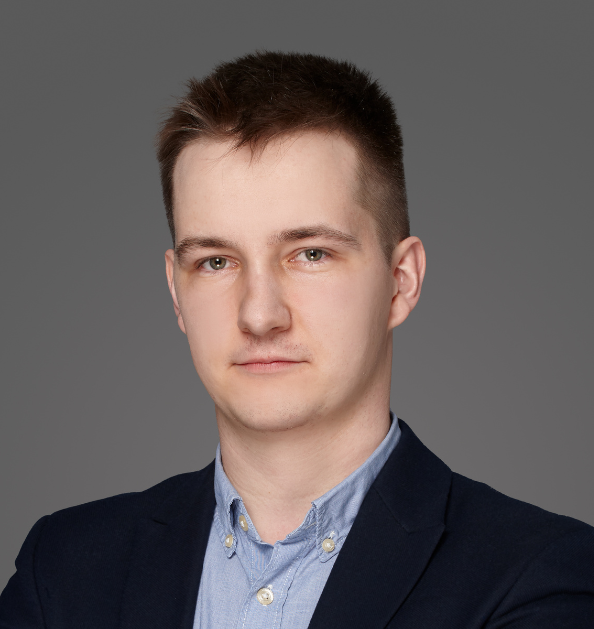}}]{Piotr Kicki}
is a post-doc in the robotics team at IDEAS and an assistant professor at the Institute of Robotics and Machine Intelligence at Poznan University of Technology.
He received his B.Eng. and M.Sc. degrees in automatic control and robotics from Poznan University of Technology, Poland in 2018 and 2019, respectively. He completed his Ph.D. from the same university in 2024.
His primary research interests center on the application of machine learning to improve robot motion planning and control.
\end{IEEEbiography}

\vspace{-1.0cm}

\begin{IEEEbiography}[{\includegraphics[width=1in,height=1in,clip,keepaspectratio]{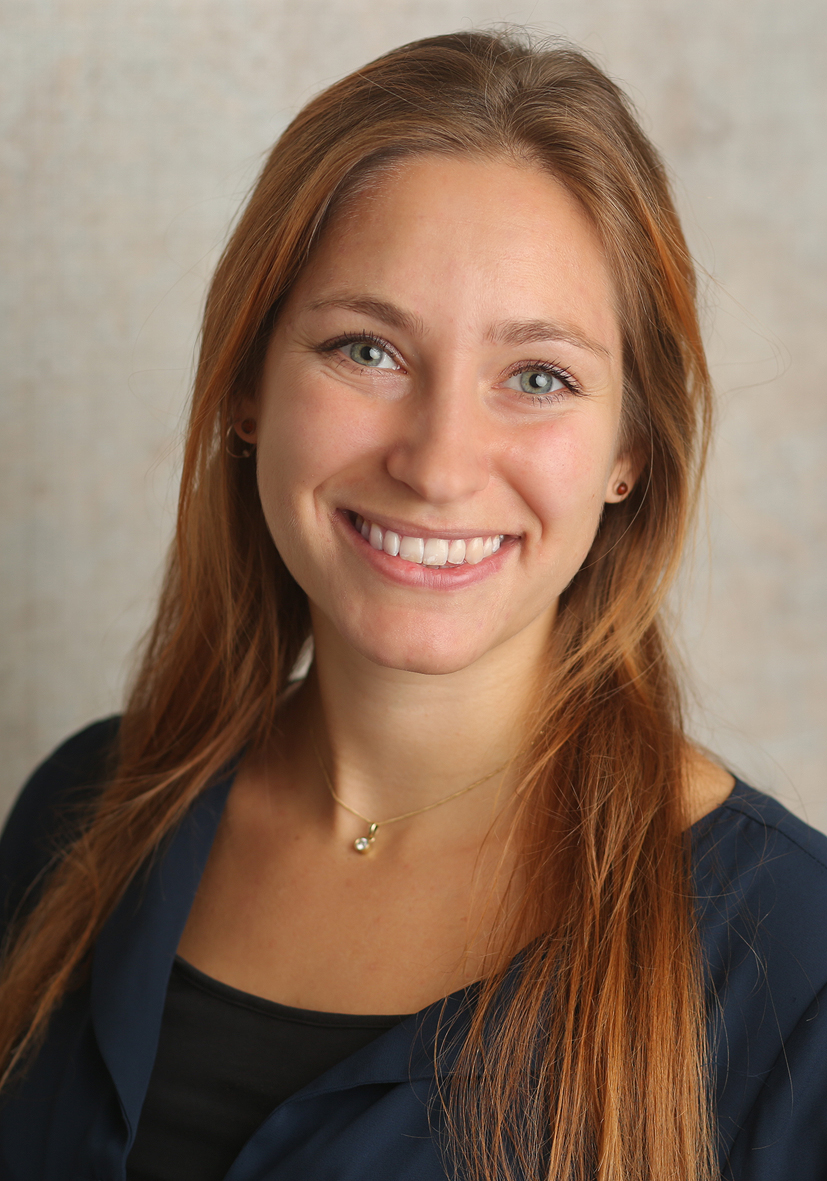}}]{Dorothea Koert}
is an Independent Research Group Leader of the interdisciplinary BMBF junior research group IKIDA, which started in October 2020.
She completed her Ph.D. from the Technical University of Darmstadt in February 2020.
In 2019 she was awarded the AI-Newcomer award by the German Society for Computer Science (GI).
During her Ph.D. she has worked on imitation learning and interactive reinforcement learning, for autonomous and semi-autonomous acquisition of motion skill libraries in human-robot collaboration.
\end{IEEEbiography}

\vspace{-1.0cm}

\begin{IEEEbiography}[{\includegraphics[width=1in,height=1in,clip,keepaspectratio]{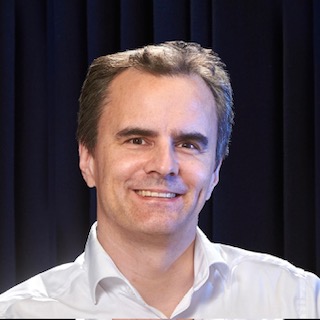}}]{Jan Peters}
is a full professor (W3) for Intelligent Autonomous Systems at the Computer Science Department of the Technical University of Darmstadt, department head of the research department on Systems AI for Robot Learning (SAIROL) at the German Research Center for Artificial Intelligence,
and a founding research faculty
member of The Hessian Center for Artificial Intelligence.
He has received the Dick Volz Best 2007 US Ph.D. Thesis Runner-Up Award, 
RSS - Early Career Spotlight, INNS Young Investigator Award, and IEEE Robotics \& Automation
Society's Early Career Award, as well as numerous best paper awards.
He received an ERC Starting Grant and was appointed an IEEE fellow, AIAA fellow, and ELLIS fellow.
\end{IEEEbiography}

\vfill

\end{document}